\documentclass[sn-mathphys-num]{sn-jnl}

\usepackage{amsmath} 
\usepackage{amssymb} 
\usepackage[normalem]{ulem} 
\usepackage{algorithm} 
\usepackage{algpseudocode} 
\usepackage{comment} 
\usepackage{multirow} 
\usepackage{color} 
\usepackage{tikz} 
\usetikzlibrary{bayesnet} 
\usepackage{subcaption} 
\usepackage{diagbox} 
\usepackage{float}
\usepackage{placeins}
\usepackage{booktabs} 
\usepackage{enumitem}

\newcommand{\argmax}{\mathop{\rm arg~max}\limits} 

\begin{document}

\title{Composite Gaussian Processes Flows \\ for Learning Discontinuous Multimodal Policies}
\author*[1]{\fnm{Shu-yuan} \sur{Wang}}\email{wang.shuyuan.wm8@is.naist.jp}
\author[1]{\fnm{Hikaru} \sur{Sasaki}}\email{sasaki.hikaru@is.naist.jp}
\author[1]{\fnm{Takamitsu} \sur{Matsubara}}\email{takam-m@is.naist.jp}
\affil[1]{\orgdiv{Graduate School of Science and Technology}, \orgname{Nara Institute of Science and Technology}, \orgaddress{\state{Nara}, \country{Japan}}}

\abstract{
Learning control policies for real-world robotic tasks often involve challenges such as multimodality, local discontinuities, and the need for computational efficiency.
These challenges arise from the complexity of robotic environments, where multiple solutions may coexist.
To address these issues, we propose Composite Gaussian Processes Flows (CGP-Flows), a novel semi-parametric model for robotic policy.
CGP-Flows integrate Overlapping Mixtures of Gaussian Processes (OMGPs) with the Continuous Normalizing Flows (CNFs), enabling them to model complex policies addressing multimodality and local discontinuities.
This hybrid approach retains the computational efficiency of OMGPs while incorporating the flexibility of CNFs.
Experiments conducted in both simulated and real-world robotic tasks demonstrate that CGP-flows significantly improve performance in modeling control policies.
In a simulation task, we confirmed that CGP-Flows had a higher success rate compared to the baseline method, and the success rate of GCP-Flow was significantly different from the success rate of other baselines in chi-square tests.
}

\maketitle

\section{Introduction}
Gaussian Processes (GPs) have demonstrated significant potential in the domain of robotic policy learning~\cite{deisenroth2013gaussian, schreiter2015sparse, sasaki2022gaussian}. Their effectiveness is due to many basic robotic control tasks require continuous function modeling of control policies~\cite{tsurumine2019deep, shen2020deep}.
GPs excel in such scenarios due to their use of kernel methods, which employ kernel functions for measuring the similarity between data points.
The kernel methods allow GPs to use training data to model smooth policies~\cite{schulz2018tutorial}.

\begin{figure}[!t]
    \centering
    \begin{minipage}[b]{0.4\hsize}
        \centering
        \includegraphics[width=1.0\hsize, trim=0 0 0 20, clip]{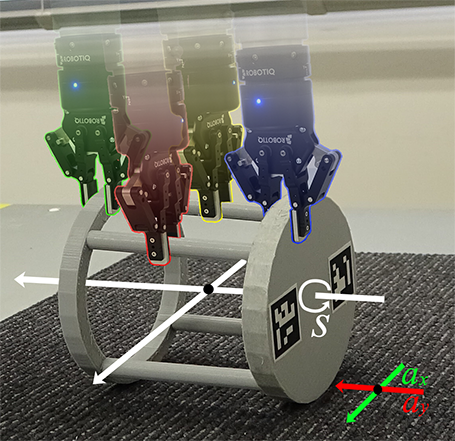}
        \subcaption{}
        \label{fig:demo:a}
    \end{minipage}
    \begin{minipage}[b]{0.4\hsize}
        \centering
        \includegraphics[width=1.0\hsize, trim=0 0 0 20, clip]{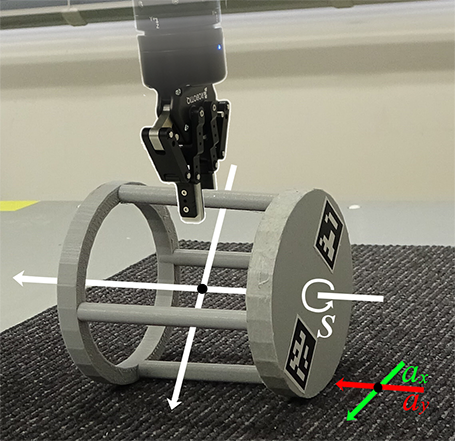}
        \subcaption{}
        \label{fig:demo:b}
    \end{minipage} \\ 
    \begin{minipage}[b]{0.8\hsize}
        \centering
        \includegraphics[width=1.0\hsize]{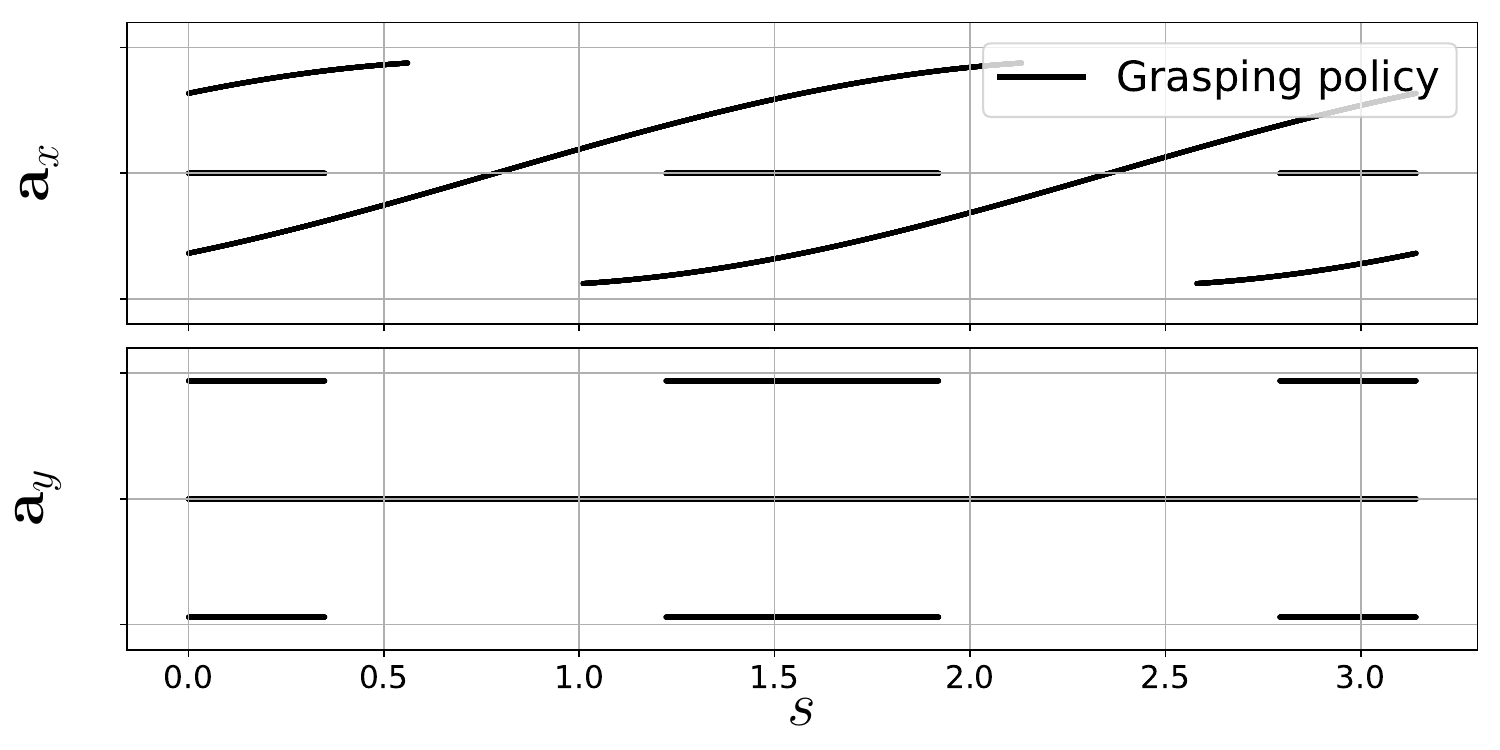}
        \subcaption{}
        \label{fig:demo:c}
    \end{minipage}
    \caption{Example of a real-world robot task in which a robot grasps an object such as a chair: Chair rotation $s$ is the state, and 2D coordinate of the gripper on X and Y axes $\mathbf{a}=\{\mathbf{a}_x, \mathbf{a}_y\}$ denote the action. (a) This specific pose has four graspable positions. (b) As chair rotates counterclockwise, number of graspable positions drops to one. (c) Expert policy for grasping task. Its modality is varied by state.}
    \label{fig:demo}
\end{figure}

However, real-world robotic tasks often require complex control policies that GP models struggle to learn due to their inherent limitations. In this research, we consider a common issue, which is multimodality in decision making. For example, as illustrated in Fig.\ref{fig:demo}, a robot is tasked with grasping a chair-like object, where the number of graspable positions is unknown and changeable based on the object's rotation. As the chair rotates from the pose in Fig.\ref{fig:demo:a} to that in Fig.\ref{fig:demo:b}, the number of graspable positions decreases from four to one, consequently altering the policy from quadmodal to unimodal. The modality of the expert policy changes based on the state (Fig.\ref{fig:demo:c}) and these changes result in local discontinuities in the policies.

\begin{figure}[t]
\centering
    \begin{minipage}[b]{0.4\hsize}
        \centering
        \includegraphics[width=1.0\hsize]{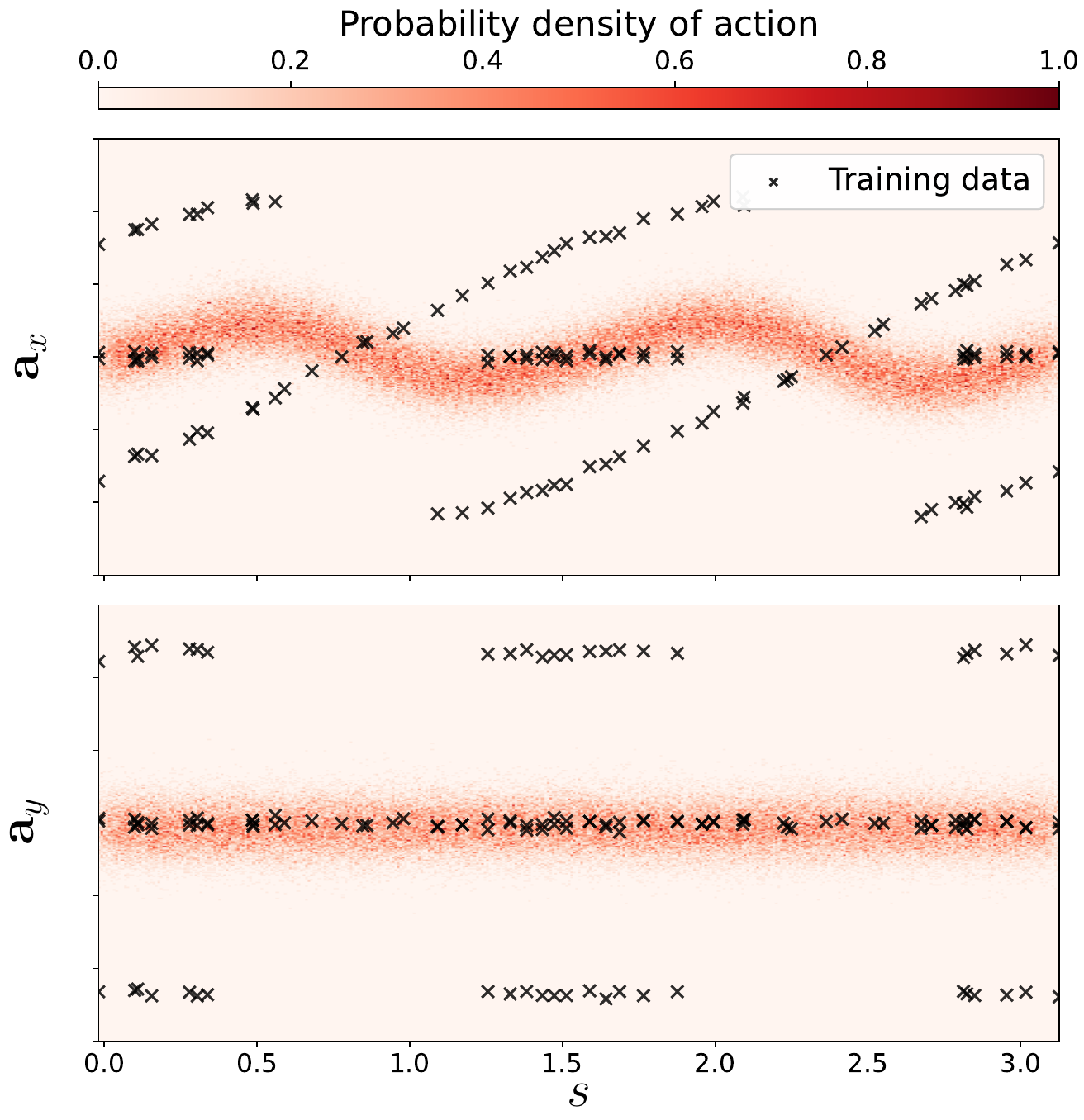}
        \subcaption{GP policy}
        \label{fig:four_policy:a}
    \end{minipage}
    \begin{minipage}[b]{0.4\hsize}
        \centering
        \includegraphics[width=1.0\hsize]{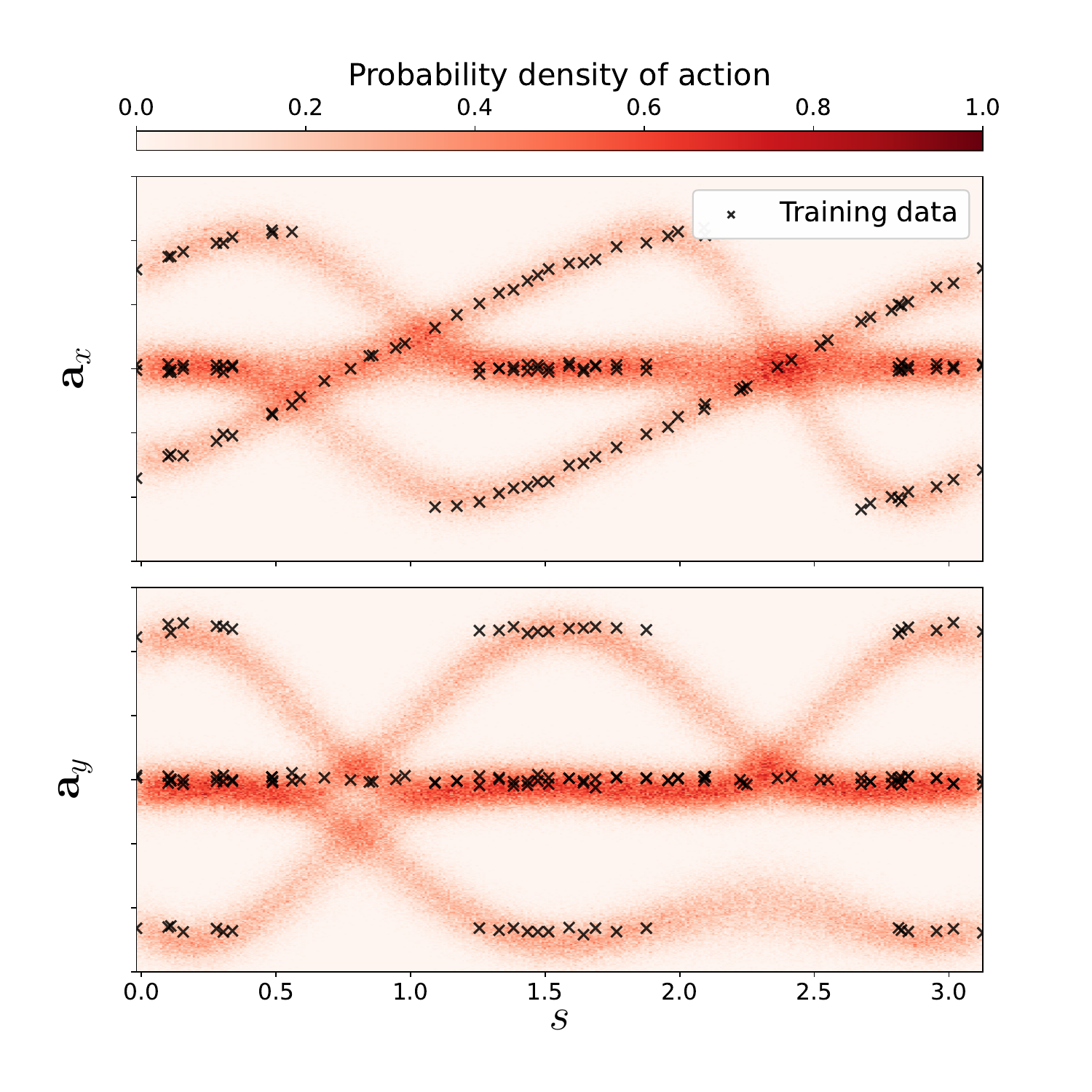}
        \subcaption{OMGP policy}
        \label{fig:four_policy:b}
    \end{minipage}
    \begin{minipage}[b]{0.4\hsize}
        \centering
        \includegraphics[width=1.0\hsize]{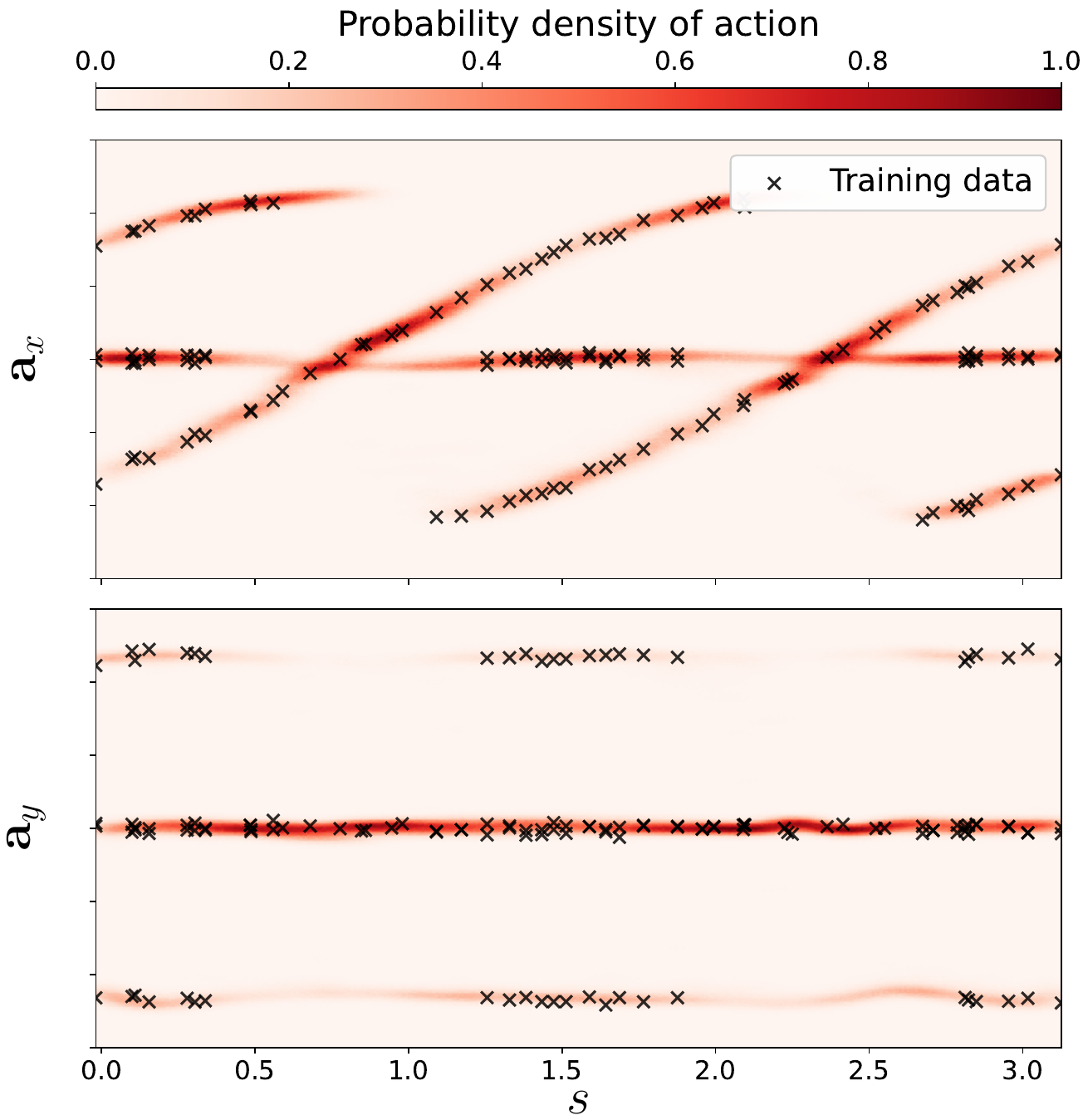}
        \subcaption{NGGP policy}
        \label{fig:four_policy:c}
    \end{minipage}
    \begin{minipage}[b]{0.4\hsize}
        \centering
        \includegraphics[width=1.0\hsize]{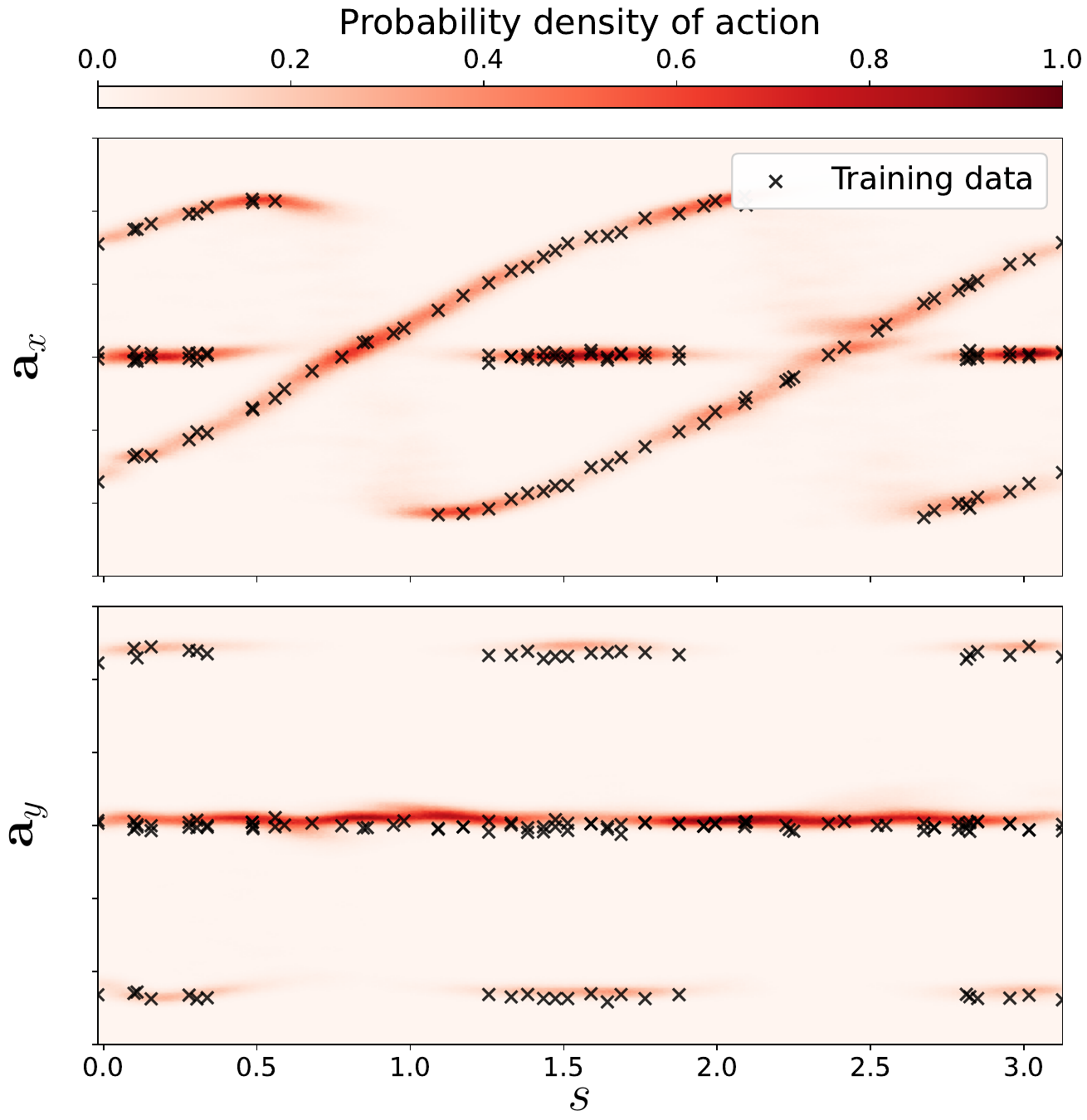}
        \subcaption{CGP-Flow policy}
        \label{fig:four_policy:d}
    \end{minipage} \\
    \caption{Learned policies by imitation learning framework for robot task in Fig.~\ref{fig:demo} with four probabilistic models: (a) GPs, (b) OMGPs, (c) NGGPs and (d) CGP-Flows. Black markers are learning data. Color represents probability density of action corresponding to each state.}
    \label{fig:four_policy}
\end{figure}

There are two common strategies to address the challenge of multimodal GP modeling.
The first uses a mixture of GPs to model multimodal policies.
An example is the Overlapping Mixtures of Gaussian Processes (OMGPs)~\cite{lazaro2012overlapping}, which model multimodal policies using overlapping GPs. The number of mixtures, $M$, is a hyperparameter that must be set to match the number of modalities in the expert policy.
However, the modality of expert policies is usually unknown in real-world robotics tasks.
If the mixture number is set smaller than the true number, the learned policy will fail to capture the multimodal nature of the expert policy.
Since OMGPs assume that multimodality is fixed across the entire space, they cannot accommodate the discontinuities caused by changes in multimodality (Fig.~\ref{fig:four_policy:b}).
Another strategy is to transform GP unimodal distributions into multimodal distributions.
A standout example is the Non-Gaussian Gaussian Processes (NGGPs)\cite{sendera2021non}, which integrate GP regression with conditional Continuous Normalizing Flows (cCNFs)\cite{papamakarios2021normalizing}. The cCNFs leverage Neural Ordinary Differential Equations (Neural-ODEs)~\cite{chen2018neural} to transform the GP model into the target distribution.
This idea of using GPs for cCNFs allows NGGPs to capture modalities and local discontinuity in policies.
However, the training time of cCNF's Neural-ODE is proportional to the modalities and the discontinuities of the target policies, and the transformation of GP base distributions may compromise precision in high-accuracy policy modeling (Fig.~\ref{fig:four_policy:c}).

\begin{figure*}[!t]
    \centering
    \includegraphics[width=0.99\hsize]{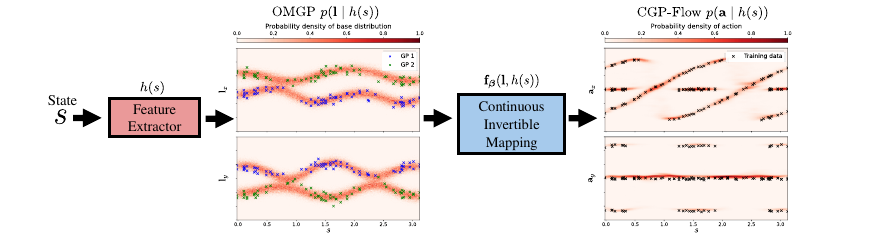}
    \caption{Fundamental framework of our approach: State is first compressed into a lower-dimensional condition by $h(\cdot)$, which is then used as components of OMGP's multimodal base distributions. Subsequently, Neural-ODE $\mathbf{f}_{\boldsymbol{\beta}}(\cdot)$ transforms base distributions into the target distributions.}
\label{fig:demo:transformation}
\end{figure*}

Previous research argues that smaller disparities between the base and target distributions can reduce computation time and improve the accuracy of the Neural-ODE transformations~\cite{grathwohl2018ffjord}.
Based on this idea, we hypothesize that highly expressive base distributions, such as multimodal distributions, improve cCNF's modeling performance.
We explore this hypothesis by proposing a novel method: Composite Gaussian Processes Flows (CGP-Flows). 
CGP-Flows integrate OMGPs as composite GPs to address multimodal base distributions, resulting in reduced complexity of the Neural-ODE transformation and achieving an optimal balance among computational efficiency, accuracy, and flexibility.
A straightforward Neural-ODE transformation enhances the expressive power of CGP-Flow, leading to superior accuracy over NGGPs when modeling local discontinuities (Fig.~\ref{fig:four_policy:d}).
The proposed framework of CGP-Flows is illustrated in Fig.~\ref{fig:demo:transformation}.
We evaluate and verify the performance of OMGPs, NGGPs, and our proposed method through simulations and a real-world robotics task using the Behavior Cloning (BC) \cite{osa2018} of expert policies.
The complexity of the policy modeling arises from the multimodal and locally discontinuous nature of the designed expert policies. Comparisons among the OMGPs, NGGPs, and CGP-Flows demonstrate that our method exhibits a significant advantage in successfully completing these tasks.

In summary, this paper makes the following contributions:

\begin{enumerate}[label=(\arabic*)]
    \item Novel hybrid model: We propose CGP-Flows, an innovative semi-parametric approach that combines sparse OMGPs with Neural-ODE. This design reduces the complexity of distribution transformations, balancing computational efficiency and modeling accuracy.
    \item Computational effective policy modeling: CGP-Flows address challenges in robotic policy learning, including multimodality, local discontinuities, and efficiency. We demonstrate superior adaptability and performance by comparing CGP-Flows with OMGP and NGGP in both simulation and real-world tasks.
    \item Extensive experiments: We conducted experiments with three tasks to validate the proposed CGP-Flows model under diverse settings and conditions. These include multiple simulation tasks with varying levels of complexity and real-world robotic BC experiments. The results consistently demonstrate the robustness and superiority of CGP-Flows in modeling policies for complex, multimodal, and discontinuous environments.
\end{enumerate}

The remainder of this paper is organized as follows:
Section 2 reviews related works on policy modeling.
Section 3 introduces the preliminaries, including GPs, OMGPs, NGGPs, and BC.
Section 4 presents the CGP-Flows model, detailing its structure and theoretical basis.
Section 5 describes two simulations that compare the performance of CGP-Flows with baseline models of OMGPs and NGGPs.
Section 6 demonstrates a real-world robot experiment, evaluating the accuracy and efficiency of CGP-Flows against baselines and analyzing its policy behavior.
Finally, in sections 7 and 8, we discuss and conclude the proposed method and outline future research directions.

\section{Related Works}\label{section 2}
\subsection{Normalizing Flows in Robotics}

Normalizing flows have demonstrated their versatility in modeling complex distributions in robotics.
Chang et al. introduced IL-flOw~\cite{chang2021ilflow}, they applied normalizing flows to imitation learning from observations, addressig the challenges of trajectory reconstruction in multimodal settings by modeling expert state transitions directly.
This approach provided robust performance while smoothing reward signals and avoiding instability in adversarial training.
On the other hand, Chisari et al. introduced Point Flow Match (PFM)~\cite{chisari2024learning}, a method that applies conditional Flow Matching (cFM) to robotic manipulation policy learning. This approach utilizes conditional flows to model policies directly from point cloud data, enabling the generation of long-horizon, multi-step trajectories adaptable to various task scenarios.
Moreover, Fadel et al. proposed a Contextual Movement Model (CMM)~\cite{fadel2023contextual} using conditional Normalizing Flows (cNFs) to analyze spatiotemporal data, integrating contextual variables such as speed and positional relationships for efficient, real-time predictions. These advancements collectively highlight the potential of normalizing flows in handling multimodality, nonlinear dynamics, and local discontinuities in robotic policies.

While IL-flOw, PFM, and CMM leverage normalizing flows for specific tasks, they lack the efficiency required for general policy modeling. IL-flOw decouples reward learning but struggles with scalability to complex policies. PFM excels in manipulation tasks but is constrained by reliance on point clouds. The CMM is efficient for spatiotemporal data but fails to generalize to varying, multimodal policies.

\subsection{Gaussian Processes in Robotic Policy}
GPs have been widely utilized in robotic control due to their ability to probabilistically perform regression on functions that describe relationships within data while avoiding overfitting, even with limited data.
The Probabilistic Inference for Learning Control (PILCO)~\cite{deisenroth2013gaussian} was introduced using a GP model to enhance data efficiency in autonomous learning systems, particularly in robotics and control.
This method significantly reduces the extensive interaction typically required in model-based reinforcement learning, making it practical for real-time applications and accelerating learning.
Another application~\cite{schreiter2015sparse} presents the use of sparse GPs, which select a representative subset of the available training data as a pseudo dataset to provide more efficient modeling. This paper demonstrates the capability of the sparse GP policy model in real-time robot control tasks. 
Furthermore, another work described a multimodal policy search algorithm based on a policy of a sparse OMGP~\cite{sasaki2021variational}. Since policy search tasks often require high computational efficiency, this method applies a sparse technique to OMGPs to reduce their computational cost. The sparse OMGP policies allow the policy search algorithm to learn multimodal policies, although it still has limitations in modeling local discontinuities and requires a known mixture number of $M$.

These works collectively focus on data efficiency or reducing computation time, both of which are crucial for advancing real-time robotic systems.
However, since their algorithms are typically suited for simpler policies and environments, these methods are incapable of learning complex multimodal policies or policies with local discontinuities caused by the environment.

\subsection{Deep Generative Models in Robotic Policy}
The development of models for robotic policy learning has seen significant advancements with the integration of deep learning techniques. Recent research has focused on multimodal imitation learning frameworks. For instance, Imitation Generative Adversarial Nets (Imitation GAN)~\cite{hausman2017multi} is a framework based on Generative Adversarial Nets (GAN)~\cite{creswell2018generative}. Imitation GAN uses a latent intention distributed by a categorical or uniform distribution to select a specific mode of the multimodal policy.
Simulations in a previous work~\cite{hausman2017multi} show that the ability to capture latent intention of an Imitation GAN outperforms GAN, but it cannot learn all the modes for complex multimodal tasks, such as a four-modal reaching task.
Another more advanced approach is Implicit Behavioral Cloning (IBC)~\cite{florence2022implicit}, which runs optimization on an implicit energy function instead of directly on the multimodal policy model. 
Simulations and experiments show that IBC has great performance on contact rich tasks and can learn bimodal policies. 
However, its training process can take over 15 hours even for simple bimodal tasks.
Another work~\cite{chi2023diffusion} proves that Diffusion Policy has a better success rate than IBC on complex multimodal tasks; unfortunately, it needs over 12 hours to learn each task.

The above studies emphasize the advancements and challenges in deep learning for robotic policy learning. The high computational cost of these models can lead to substantial time and resource expenditure when adapting them to different tasks, limiting their practicality in dynamic environments. To address this issue, we propose an approach that integrates OMGPs with a deep generative model. This method strikes a balance between cost and performance, providing a practical solution for real-world robotics tasks.

\section{Preliminaries}\label{section 3}
In this section, we introduce the probabilistic model of GP-based models and BC as an imitation learning method.

\subsection{Gaussian Processes}
GPs are generally used to regress a function between input data $\mathbf{X}\triangleq\{\mathbf{x}_{n}\}_{n=1}^{N}$ and output data $\mathbf{Y}\triangleq\{\mathbf{y}_{n}\}_{n=1}^{N}$ in training data $\mathcal{D}\triangleq \{\mathbf{X}, \mathbf{Y}\}$.
For the training data, GPs assume the following existence of function $f$ between input $\mathbf x_n$ and output data $\mathbf y_n$: $\mathbf{y}_n=f(\mathbf{x}_n)+\boldsymbol{\epsilon}_n$, where $\boldsymbol{\epsilon}_n \sim \mathcal{N}(\mathbf{0}, \sigma^2)$ is Gaussian noise.

\begin{figure}[t]
    \centering
    \resizebox{0.32\linewidth}{!}{
    \begin{tikzpicture}
        \node[latent, circle] (f)     at (1.5, 0) {${\mathbf{f}}$};
        \node[const]          (X)     at (0, 0) {$\mathbf{X}$};
        \node[obs, circle]    (y)     at (3, 0) {$\mathbf{Y}$};
        \node[const]          (theta) at (1.5, 1.0) {$\theta$};
        \node[const]          (sigma) at (3.0, 1.0) {$\sigma$};
        \draw[->] (X) -- (f);
        \draw[->] (f) -- (y);
        \draw[->] (theta) -- (f);
        \draw[->] (sigma) -- (y);
    \end{tikzpicture}}
    \centering
    \caption{Graphical model of GPs: $\mathbf{X}$ is input data; $\mathbf{Y}$ is output data. Function values are denoted by $\mathbf{f}$. $\theta$ is kernel function parameter. $\sigma$ is a parameter of likelihood.}
    \label{fig:graphical:GP}
\end{figure}
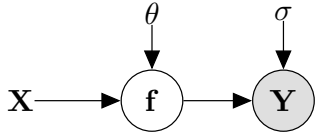

GPs provide a prior of function $f$ as $f \sim \mathcal{GP}(\mathbf{0}, k(\mathbf{x}, \mathbf{x}))$,

where $k(\cdot, \cdot)$ is a kernel function.
The function values are set to $\mathbf{f}=\{f(\mathbf x_n)\}_{n=1}^N$, and the prior of function values $\mathbf f$ is given: 
\begin{align}
    p(\mathbf{f}\mid \mathbf{X}) = \mathcal{N}(\mathbf{f}\mid \mathbf{0}, \mathbf{K_X}),
\end{align}
where $\mathbf{K_X}$ is a kernel gram matrix, and each element of $\mathbf K_{\mathbf X}$ is computed as $[\mathbf{K_x}]_{ij} = k(\mathbf{x}_i, \mathbf{x}_j)$.
Since $\boldsymbol{\epsilon}_n$ is assumed to be Gaussian, the likelihood can be modeled using Gaussian distribution as $p(\mathbf{Y} \mid \mathbf{f}) = \mathcal{N}(\mathbf{Y} \mid \mathbf{f}, \sigma^2 \mathbf{I})$.
Using this likelihood, the probability distribution of output data $\mathbf Y$ is modeled:
\begin{align}\label{eq:gp:model}
    p(\mathbf{Y}\mid \mathbf{X}) =  \int p(\mathbf{Y\mid f})p(\mathbf{f \mid X}) ~ \mathrm d \mathbf f.
\end{align}
The graphical model of GP is illustrated in Fig.~\ref{fig:graphical:GP}.

The predictive distribution for output $\mathbf{y}_{*}$ at $\mathbf{x}_{*}$ is analytically given:
\begin{align}\label{eq:gp:predictive_distribution}
    p(\mathbf{y}_* \mid \mathbf{x}_*, \mathbf{X}, \mathbf{Y}) =\mathcal{N}(\mathbf{y}_* \mid \boldsymbol{\mu}_*(\mathbf x_*),\boldsymbol{\sigma}_*(\mathbf x_*)),
\end{align}
where $\boldsymbol{\mu}_*(\cdot)$ and $\boldsymbol{\sigma}_*(\cdot)$ are the mean and the variance function of the predictive distribution.
The formulation details were previously described~\cite{RasmussenW06}.

\subsection{Overlapping Mixtures of Gaussian Processes}
OMGPs \cite{lazaro2012overlapping} were proposed to model data generated from multimodal functions using a mixture of GPs.

OMGPs assume a multimodal prior for function $f$ between input data $\mathbf X = \{\mathbf x_n\}_{n=1}^N$ and output data $\mathbf Y = \{\mathbf y_n\}_{n=1}^N$ as $f \sim \prod_{m=1}^M \mathcal{GP}^{(m)}(\mathbf{0}, k^{(m)}(\mathbf{x}, \mathbf{x}))$,

where the mixture number $M$ denotes the modality of the function, and $k^{(m)}(\cdot, \cdot)$ is a kernel function for the $m$-th GP prior.
This multimodal GP prior assumes that the function value takes on $M$ values for input data.
The prior for all $M$ function values $\{\mathbf{f}^{(m)}\}$ is modeled:
\begin{align}
    p(\{\mathbf{f}^{(m)}\} \mid \mathbf{X})=\prod_{m=1}^M \mathcal{N}(\mathbf{f}^{(m)} \mid \mathbf{0}, \mathbf{K}_{\mathbf{X}}^{(m)}),
\end{align}
where $\mathbf{f}^{(m)}$ is the function values for the $m$-th GP and $\mathbf{K}_{\mathbf{X}}^{(m)}$ is a kernel gram matrix for the $m$-th GP.

The multimodal GP prior assumes multiple outputs of function $f$.
However, actual output data $\mathbf y_n$ is related to only one of the $M$ functions.
OMGPs introduce binary indicator matrix $\mathbf{Z}$, where the element $\left[\mathbf{Z}\right]_{nm}$ denotes whether the $n$-th output data $\mathbf{y}_n$ is assigned to $m$-th GP.
Each row of $\mathbf{Z}$ contains only one non-zero entry, ensuring that each data point is assigned to only one GP.
The prior on binary indicator matrix $\mathbf{Z}$ is defined:
\begin{align}
    p(\mathbf{Z})=\prod_{m=1,n=1}^{M, N}[\mathbf{\Pi}]_{nm}^{[\mathbf{Z}]_{nm}}, \label{eq:prior:indicator:matrix}
\end{align}
where $\mathbf{\Pi}$ is the $N \times M$ probability matrix.

Using the indicator matrix, the likelihood of the OMGPs is given:
\begin{align}
    p({\mathbf{Y}}\mid \{\mathbf{f}^{(m)}\}, \mathbf{Z}) = \prod_{m=1,n=1}^{M, N} \mathcal{N}(\mathbf{y}_n\mid \mathbf{f}^{(m)}_n, \sigma^2)^{[\mathbf{Z}]_{nm}}. \label{eq:likelihood:omgp}
\end{align}

\begin{figure}[t]
    \centering
    \resizebox{0.42\linewidth}{!}{
    \begin{tikzpicture}
        \node[latent, circle] (f)     at (1.5, 0)    {${\mathbf{f}}^{(m)}$};
        \node[const]          (X)     at (0, 0)      {$\mathbf{X}$};
        \node[obs, circle]    (y)     at (3.5, 0)    {$\mathbf{Y}$};
        \node[const]          (theta) at (1.5, 1.2) {$\theta^{(m)}$};
        \node[const]          (sigma) at (2.7, 1.2) {$\sigma$};
        \node[latent, circle] (z)     at (4.2, 1.2) {$\mathbf{Z}$};
       \plate[inner sep=0.4cm, yshift=0.cm] {plateN} {(f) (theta)} {};
       \node[const] (M) at (1., -0.7) {$M$};

        \draw[->] (X) -- (f);
        \draw[->] (f) -- (y);
        \draw[->] (theta) -- (f);
        \draw[->] (z) -- (y);
        \draw[->] (sigma) -- (y);
    \end{tikzpicture}}
    \caption{Graphical model of OMGPs: Variable $\mathbf{X}$ denotes input data, and $\mathbf{Y}$ represents output data. OMGP model has $M$ function variables $\{\mathbf{f}^{(m)}\}$, and each function value $\mathbf f^{(m)}$ has a GP prior. $\theta^{(m)}$ is a kernel function parameter for $m$-th GP prior. Variable $\mathbf Z$ is a binary indicator matrix to assign output data $\mathbf y_n$ to the $M$-th GP. $\sigma$ is a likelihood parameter.}
    \label{fig:graphical:OM-GPs}
\end{figure}
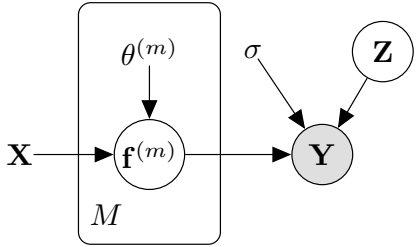

Using the above likelihood and priors, the probabilistic model of OMGPs is expressed:
\begin{align}\label{eq:OM-GPs:model}
     p({\mathbf{Y}}\mid {\mathbf X})
     = & \int p({\mathbf{Y}}\mid \{\mathbf{f}^{(m)}\}, {\mathbf{Z}})  p(\{\mathbf{f}^{(m)}\}\mid\mathbf X)p({\mathbf{Z}}) ~ \mathrm d \{\mathbf f^{(m)}\}\mathrm d \mathbf Z. 
\end{align}
A graphical model of OMGPs is illustrated in Fig.~\ref{fig:graphical:OM-GPs}.

To calculate the predictive distribution of OMGPs, the posterior distribution of function values $\{\mathbf f^{(m)}\}$ and indicator matrix $\mathbf Z$ are needed.
However, posterior distributions cannot be solved analytically.
Therefore, OMGPs utilize variational inference to analytically obtain an approximation of posterior distribution.
Using an approximation of posterior distributions $q(\mathbf f^{(m)})$ and $q(\mathbf Z)$, the predictive distribution of OMGPs is given:
\begin{align}\label{eq:OM-GPs:prediction}
    p(\mathbf{y}_* \mid \mathbf{x}_*, \mathbf{X}, \mathbf{Y}) &\approx 
    \prod^{M}_{m=1} \left[\boldsymbol{\Pi}\right]_{*m} \int p(\mathbf{y}_* \mid \mathbf{f}^{(m)}, \mathbf{x}_*, \mathbf{X}) q(\mathbf{f}^{(m)}) ~ \mathrm{d} \mathbf{f}^{(m)} \nonumber \\
    &= \prod^{M}_{m=1} \left[\boldsymbol{\Pi}\right]_{*m} \mathcal{N}(\mathbf{y}_* \mid \boldsymbol{\mu}_*^{(m)}(\mathbf x_*), \boldsymbol{\sigma}_*^{(m)}(\mathbf x_*)),
\end{align}
where $\boldsymbol{\mu}_*^{(m)}(\cdot)$ and $\boldsymbol{\sigma}_*^{(m)}(\cdot)$ are the mean and variance function of the predictive distribution for the $m$-th modality.

\subsection{Non-Gaussian Gaussian Processes}
NGGPs \cite{sendera2021non} are a probabilistic model that can capture non-Gaussian distribution using GP's latent function regression and Neural-ODEs' distribution transformation.
NGGPs introduce latent variable $\mathbf L\triangleq\{\mathbf{l}_n\}_{n=1}^{N}$ between output data $\mathbf Y$ and input data $\mathbf X$.
NGGPs also assume a function relation between variable $\mathbf L$ and input data $\mathbf X$ as $\mathbf{l}_n = f(\mathbf{x}_n)+\boldsymbol{\epsilon}_n$.
Probabilistic distribution $p(\mathbf L\mid \mathbf X)$ is modeled using a GP like Eq.~\ref{eq:gp:model}, and this distribution of latent variable $\mathbf L$ is called a base distribution.
Gaussian-form distribution $p(\mathbf L\mid \mathbf X)$ is transformed into non-Gaussian distribution $p(\mathbf Y\mid \mathbf X)$ using the Neural-ODE.

\begin{figure}[t]
    \centering
    \resizebox{0.45\linewidth}{!}{
        \begin{tikzpicture}
            \node[latent, circle] (F)     at (1.5, 0) {${\mathbf{f}}$};
            \node[const]          (X)     at (0, 0) {$\mathbf{X}$};
            \node[latent, circle] (l)     at (3, 0) {$\mathbf{L}$};
            \node[const]          (theta) at (1.5, 1.2) {$\theta$};
            \node[const]          (sigma) at (3.0, 1.2) {$\sigma$};
            \node[obs, circle]    (y)     at (4.5, 0) {$\mathbf{Y}$};
            \node[const]          (beta)  at (4.5, 1.2) {$\boldsymbol{\beta}$};
            \draw[->] (X) -- (f);
            \draw[->] (f) -- (l);
            \draw[->] (l) -- (y);
            \draw[->] (theta) -- (f);
            \draw[->] (sigma) -- (l);
            \draw[->] (beta) -- (y);
        \end{tikzpicture}
    }
    \caption{Graphical model of NGGPs: Variables $\mathbf{X}$ and $\mathbf Y$ denote input and output data. $\mathbf{L}$ is a latent variable and $p(\mathbf L)$ is modeled by GPs. Both variable $\mathbf L$ and probabilistic distribution $p(\mathbf L)$ are transformed to variable $\mathbf Y$ and distribution $p(\mathbf Y)$ by Neural-ODE. $\theta$ and $\boldsymbol{\beta}$ are parameters of kernel function and Neural-ODE. $\sigma$ is a parameter of likelihood.}
    \label{fig:graphical:NGGP}
\end{figure}
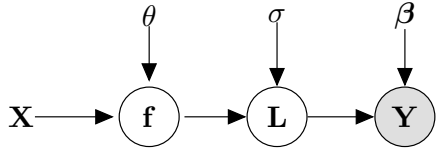

The transformations of Neural-ODE in the NGGP models provide an invertible transformation of both a variable and distribution.
The invertible transformation is expressed using continuous time $t$ and formulated using a Neural-ODE.
In this formulation, variable $\mathbf l_n$ is expressed as $\mathbf l_n=\mathbf l_n(t_0)$, and variable $\mathbf y_n$ is expressed as $\mathbf y_n=\mathbf l_n(t_1)$.
The variable's forward and backward transformation is formulated:
\begin{align}
    \mathbf{y}_n &=  \mathbf{l}_n + \int_{t_0}^{t_1} \mathbf{g}_{\boldsymbol{\beta}}(\mathbf{l}_n(t),t,h({\mathbf x_n})) ~\mathrm{d} t, \label{eq:cnf:forward} \\
    \mathbf{l}_n &=  \mathbf{y}_n - \int_{t_0}^{t_1} \mathbf{g}_{\boldsymbol{\beta}}(\mathbf{y}_n(t),t,h({\mathbf x_n})) ~\mathrm{d} t, \label{eq:cnf:backward} 
\end{align}
where ${\mathbf g}_{\boldsymbol{\beta}}(\cdot)$ is a continuous-time dynamic function:
\begin{align}
    {\mathbf g}_{\boldsymbol{\beta}}(\mathbf{l}_n(t), t, h({\mathbf x_n})) = \frac{\partial \mathbf{l}(t)}{\partial t},
\end{align} 
where $h(\cdot)$ is a condition extractor that converts the high-dimensional state to a low-dimensional condition for improved efficiency. 
Continuous-time dynamic function ${\mathbf g}_{\boldsymbol{\beta}}(\cdot)$ is modeled using a neural network and $\boldsymbol\beta$ is a parameter of the neural network.
The transformation of the probabilistic distribution can be formulated by a second order differential equation, called the \textit{instantaneous change of variables} \cite{chen2018neural}:
\begin{align}\label{eq:icv}
    \log p(\mathbf{Y}\mid \mathbf X) = 
    \log p(\mathbf{L}\mid \mathbf X) -  \sum_{n=1}^N\int_{t_0}^{t_1}\operatorname{Tr}\left(\frac{\partial {\mathbf g}_{\boldsymbol{\beta}}(\mathbf{l}_n(t),t,h({\mathbf x}_n))}{\partial \mathbf{l}_n(t)}\right) \mathrm{d} t.
\end{align}
The Neural-ODE transformation on variable and distribution can be solved by ODE solvers.
We obtain the optimal parameter of Neural-ODE $\boldsymbol{\beta}_*$ by maximizing log-likelihood $\log p(\mathbf{Y}\mid \mathbf X)$ and illustrate the graphical model of NGGPs in Fig.~\ref{fig:graphical:NGGP}.

The predictive distributions of NGGPs are obtained using Neural-ODE transformation and the GP's predictive distribution (Eq.~\ref{eq:gp:predictive_distribution}):
\begin{align}\label{eq:nggp:model}
    \log p(\mathbf{y}_*\mid \mathbf x_*, \mathbf X, \mathbf Y) =
    \log p(\mathbf{l}_*\mid \mathbf x_*, \mathbf X, \mathbf Y) -  \int_{t_0}^{t_1}\operatorname{Tr}\left(\frac{\partial {\mathbf g}_{\boldsymbol{\beta}_*}(\mathbf{l}_*(t),t,h({\mathbf x}_*))}{\partial \mathbf{l}_*(t)}\right) \mathrm{d} t.
\end{align}

\subsection{Behavior Cloning}
BC is a simple approach in imitation learning to learn robotic control policy~\cite{osa2018}.
In the BC framework, control policy $\pi$ is learned in a supervised learning manner using the demonstration data generated by expert $\pi^E$.
A dynamics model of a robot's environment is denoted as Markovian with state $\mathbf s_t$, action $\mathbf a_t$ , initial state probability $p(\mathbf{s}_0)$, and state transition distribution $p(\mathbf s_{t+1}\mid \mathbf s_t, \mathbf a_t)$.
When control policy $\pi(\mathbf a_t \mid \mathbf s_t)$ works on the dynamics model, trajectory $\boldsymbol{\tau}=\{\mathbf{s}_{1},\mathbf{a}_{1},\mathbf{s}_{2},\mathbf{a}_{2}, \cdots, \mathbf{a}_{T},\mathbf{s}_{T+1}\}$, which is a sequence of state-action pairs of $T$ steps, is obtained.
The trajectory distribution by expert policy $\pi^E$ is defined as $p(\boldsymbol{\tau}\mid \pi^{E}) = p(\mathbf{s}_{1})\prod_{t=1}^T \pi^{E}(\mathbf{a}_{t}\mid\mathbf{s}_{t})p(\mathbf{s}_{t+1}\mid \mathbf{s}_{t}, \mathbf{a}_{t})$.

Training data $\mathcal{D}$ with $N$ pairs of state and action are obtained from $E$ episodes of trajectories as $\mathcal{D}\triangleq \{\mathbf{s}_n, \mathbf{a}_n\}_{n=1}^{N}$, where $N=E\times T$.
When a control policy is modeled using a probabilistic mode, optimal policy $\pi^{*}$ can be derived by solving the following optimization problem using training data $\mathcal D$:
\begin{align}\label{eq:bc:policy}
    \pi^{*} = \argmax_\pi  \sum_{n=1}^{N} \log \pi(\mathbf{a}_n \mid \mathbf{s}_n).
\end{align}

\section{Composite Gaussian Processes Flows}
We propose CGP-Flows, a novel approach designed to learn control policies for robotic tasks.
Real-world robotic tasks often require complex control policies with smoothness, multimodality, and local discontinuities between states and actions.
The CGP-Flow model can learn such complex control policies with reasonable computational cost by integrating mixtures of GP models and distribution transformation by Neural-ODEs.

 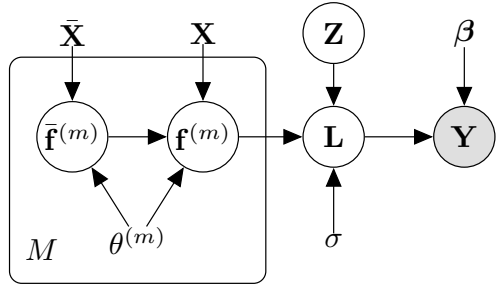
\begin{figure}[t]
    \centering
    \resizebox{0.5\linewidth}{!}{
    \begin{tikzpicture}
        \node[latent, circle] (f_bar)    at (0, 0) {$\bar{\mathbf{f}}^{(m)}$};
        \node[const]          (X_bar)    at (0, 1.2) {$\bar{\mathbf{X}}$};
        \node[latent, circle] (f)     at (1.5, 0) {${\mathbf{f}}^{(m)}$};
        \node[const]          (X)     at (1.5, 1.2) {$\mathbf{X}$};
        \node[latent, circle] (l)     at (3, 0) {$\mathbf{L}$};
        \node[const]          (theta) at (0.75, -1.2) {$\theta^{(m)}$};
        \node[const]          (sigma) at (3.0, -1.2) {$\sigma$};
        \node[obs, circle]    (y)     at (4.5, 0) {$\mathbf{Y}$};
        \node[const]          (beta)  at (4.5, 1.2) {$\boldsymbol{\beta}$};
        \node[latent, circle] (z)     at (3.0, 1.2) {$\mathbf{Z}$};
        \plate[inner sep=0.3cm, yshift=0.2cm] {plateN} {(f) (f_bar) (theta)} {};
        \node[const] (M) at (-0.35, -1.3) {$M$};
        \draw[->] (X_bar) -- (f_bar);
        \draw[->] (f_bar) -- (f);
        \draw[->] (X) -- (f);
        \draw[->] (f) -- (l);
        \draw[->] (sigma) -- (l);
        \draw[->] (l) -- (y);
        \draw[->] (theta) -- (f_bar);
        \draw[->] (theta) -- (f);
        \draw[->] (z) -- (l);
        \draw[->] (beta) -- (y);
    \end{tikzpicture}}
    \caption{Graphical model of CGP-Flows: Variables $\mathbf{X}$ and $\mathbf Y$ denote input and output data. $\mathbf{L}$ is a latent variable between variables $\mathbf f^{(m)}$ and $\mathbf Y$. Distribution $p(\mathbf L)$ is modeled by sparse OMGPs. $\bar{\mathbf X}$ and $\bar{\mathbf f}^{(m)}$ are sparse input and output. Both variable $\mathbf L$ and probabilistic distribution $p(\mathbf L)$ are transformed to variable $\mathbf Y$ and distribution $p(\mathbf Y)$ by the Neural-ODE. $\theta^{(m)}$ is a parameter of $m$-th sparse GP in sparse OMGP model. $\boldsymbol{\beta}$ are Neural-ODE parameters.}
    \label{fig:graphical:model}
\end{figure}

\subsection{CGP-Flow Model}
CGP-Flows assume latent variable $\mathbf L\triangleq\{\mathbf{l}_n\}_{n=1}^{N}$ between input data $\mathbf X$ and output data $\mathbf Y$, similar to NGGPs, and employ sparse OMGPs to model the probabilistic distribution of latent variable $\mathbf L$.
Sparse OMGP model utilizes mixtures of the $M$ sparse GP model to reduce the computational complexity by introducing sparse modeling to GPs.

The sparse OMGP assumes functional relationships between latent variable $\mathbf L$ and input data $\mathbf X$ as $\mathbf l_n = f(\mathbf x_n) + \boldsymbol{\epsilon}_n$ and places a multimodal prior for function $f$, $f \sim \prod_{m=1}^M \mathcal{GP}^{(m)}(\mathbf{0}, k^{(m)}(\mathbf{x}, \mathbf{x}))$.

Sparse OMGPs utilize pseudo-input $\bar{\mathbf X}\triangleq \{\bar{\mathbf x}_p\}_{p=1}^P$ and pseudo-function values $\bar{\mathbf f}^{(m)}\triangleq \{\bar{\mathbf f}^{(m)}_p\}_{p=1}^P$ that sparsely represent function values $\mathbf f^{(m)}$, where the size of pseudo-function values $P$ is less than the data $N$.
The probabilistic distributions of $M$ function values $\{\mathbf f^{(m)}\}$ and pseudo-function values $\{\bar{\mathbf f}^{(m)}\}$ are defined:
\begin{align}
    p(\{{\mathbf{f}}^{(m)}\}\mid\{\bar{\mathbf{f}}^{(m)}\},\bar{\mathbf{X}}, {\mathbf X}) \nonumber  &= \prod^{M}_{m=1} \mathcal{N}(\mathbf{f}^{(m)}\mid \mathbf{K}^{(m)}_{\mathbf{X}\bar{\mathbf{X}}} \mathbf{K}^{(m)^{-1}}_{\bar{\mathbf{X}}}\bar{\mathbf{f}}^{(m)}, \mathbf{K}_{\mathbf{X}}^{(m)}-\mathbf{K}^{(m)}_{\mathbf{X}\bar{\mathbf{X}}}\mathbf{K}^{(m)^{-1}}_{\bar{\mathbf{X}}}\mathbf{K}^{(m)}_{\bar{\mathbf{X}} \mathbf{X}}), \\
    p(\{\bar{\mathbf{f}}^{(m)}\}\mid \bar{\mathbf X}) &= \prod^M_{m=1} \mathcal{N}(\bar{\mathbf{f}}^{(m)}\mid \mathbf{0}, \mathbf{K}_{\bar{\mathbf{X}}}^{(m)}), 
\end{align}
where $\mathbf{K}^{(m)}_{\bar{\mathbf{X}}}$, $\mathbf{K}^{(m)}_{\mathbf{X}\bar{\mathbf{X}}}$, and $\mathbf{K}^{(m)}_{\bar{\mathbf{X}}\mathbf{X}}$ are kernel gram matrices for the $m$-th kernel function and each element is calculated as $\left[\mathbf{K}^{(m)}_{\bar{\mathbf{X}}}\right]_{ij} = k^{(m)}(\bar{\mathbf x}_i, \bar{\mathbf x}_j)$, $\left[\mathbf{K}^{(m)}_{\mathbf{X}\bar{\mathbf{X}}}\right]_{ij}= k^{(m)}(\mathbf x_i, \bar{\mathbf x}_j)$, and $\left[\mathbf{K}^{(m)}_{\bar{\mathbf{X}}\mathbf{X}}\right]_{ij}= k^{(m)}(\bar{\mathbf x}_i, \mathbf x_j)$.

Sparse OMGPs use binary indicator matrix $\mathbf Z$, and the prior of binary indicator matrix $\mathbf Z$ same as Eq. (\ref{eq:prior:indicator:matrix}) and the distribution of latent variable $\mathbf L$ are defined in the same way as  Eq. (\ref{eq:likelihood:omgp}).

Using the distributions, a sparse OMGP's base distribution is modeled and transformed to the non-Gaussian distribution of output data $\mathbf Y$ using Neural-ODE:
\begin{align}
    p({\mathbf{L}}\mid \bar{\mathbf X}, {\mathbf X}) & =\int p({\mathbf{L}}\mid \{{\mathbf{f}}^{(m)}\}, {\mathbf{Z}}) p(\{{\mathbf{f}}^{(m)}\}\mid\{\bar{\mathbf{f}}^{(m)}\},\bar{\mathbf{X}}, {\mathbf X}) \nonumber \\ & ~~~~~~~ \cdot p(\{\bar{\mathbf{f}}^{(m)}\}\mid \bar{\mathbf X})p({\mathbf{Z}}) ~ \mathrm d \{{\mathbf{f}}^{(m)}\}\mathrm d \{\bar{\mathbf{f}}^{(m)}\}~\mathrm d \mathbf Z, \\
 \log p(\mathbf{Y}\mid \bar{\mathbf{X}}, \mathbf{X})  &= \log p({\mathbf{L}}\mid \bar{\mathbf X}, {\mathbf X}) 
    - \sum_{n=1}^{N} \int_{t_0}^{t_1} \operatorname{Tr} \frac{\partial {\mathbf g}_{\boldsymbol{\beta}}(\mathbf{l}_n(t),t,h({\mathbf x_n}))}{\partial \mathbf{l}_{n}(t)} \mathrm{d} t.
\end{align}
In this paper, we use \textit{dopri5} as the ODE solver, whose details can be found in Appendix~\ref{section:dopri5_nfe}.
A graphical model of CGP-Flows is shown in Fig.~\ref{fig:graphical:model}.

\subsection{Variational Inference}
We designed a CGP-Flow model in the previous section. However, the integral on indicator matrix $\mathbf{Z}$ causes CGP-Flow's posterior distribution is analytically intractable.
Thus, we infer approximate posterior distribution and optimize the model parameters by variational inference.
In variational inference, we derive lower bound $J_L$ for marginal likelihood $p(\mathbf Y\mid \bar{\mathbf X}, \mathbf X)$ using variational distributions $q(\bar{\mathbf{f}}^{(m)})$ and $q(\mathbf Z)$
and find optimal variational distributions $q(\bar{\mathbf{f}}^{(m)})$ and $q(\mathbf Z)$ that maximize lower bound $J_L$.
As a result, the optimal variational distributions approximate the posterior distribution.
The lower bound of the marginal likelihood of CGP-Flow is derived:
\begin{align}
    \log p(\mathbf{Y}\mid \bar{\mathbf{X}}, \mathbf{X}) &= \log p({\mathbf{L}}\mid \bar{\mathbf X}, {\mathbf X}) 
    - \sum_{n=1}^{N} \int_{t_0}^{t_1} \operatorname{Tr} \frac{\partial {\mathbf g}_{\boldsymbol{\beta}}(\mathbf{l}_n(t),t,h({\mathbf x_n}))}{\partial \mathbf{l}_{n}(t)} \mathrm{d} t \nonumber \\
    & \geq J_\mathrm{base}({\mathbf L}, q({\mathbf{Z}}), q(\{\bar{\mathbf{f}}^{(m)}\})) 
    - \sum_{n=1}^{N} \int_{t_0}^{t_1} \operatorname{Tr} \frac{\partial {\mathbf g}_{\boldsymbol{\beta}}(\mathbf{l}_n(t),t,h({\mathbf x_n}))}{\partial \mathbf{l}_{n}(t)} \mathrm{d} t \nonumber \\
    & \triangleq J_L({\mathbf Y}, q({\mathbf{Z}}), q(\{\bar{\mathbf{f}}^{(m)}\})),
    \label{eq:cgpflows:likelihood}
\end{align}
where $J_\mathrm{base}$ is the lower bound of sparse OMGP's marginal likelihood (Eq. \ref{eq:OM-GPs:model}) and is computed by applying Jensen's inequality:
\begin{align}
    \log p({\mathbf{L}}\mid \bar{\mathbf X}, {\mathbf X}) &\geq J_\mathrm{base}({\mathbf L}, q({\mathbf{Z}}), q(\{\bar{\mathbf{f}}^{(m)}\})) \nonumber \\
    &=  \int p(\{\mathbf{f}^{(m)}\} \mid\{\bar{\mathbf{f}}^{(m)}\},\bar{\mathbf{X}}, \mathbf{X}) q(\{\bar{\mathbf{f}}^{(m)}\}) q(\mathbf{Z}) \nonumber \\
    & ~~~~ \cdot \log \frac{p(\mathbf{L} \mid\{\mathbf{f}^{(m)}\}, \mathbf{Z}) p(\{\bar{\mathbf{f}}^{(m)}\} \mid \bar{\mathbf{X}}) p(\mathbf{Z})}{q(\{\bar{\mathbf{f}}^{(m)}\}) q(\mathbf{Z})} 
    \mathrm{d}\{\mathbf{f}^{(m)}\} \mathrm{d}\{\bar{\mathbf{f}}^{(m)}\} \mathrm{d} \mathbf{Z}.
\end{align}

\begin{algorithm}[t]
    \caption{CGP-Flows training process}
    \label{algorithm:CGP-Flows}
    \begin{algorithmic}[1]
        \Require Training data $\{\mathbf{X}, \mathbf{Y}\}$
            \State Initialize parameters $\bar{\mathbf{X}}$, $\{\theta^{(m)}\}$, $\boldsymbol{\beta}$
            \While{$J_L$ not converged}
                \State \# E-step
                \While{$J_{L}$ not converged}
                    \State Update $q(\{\bar{\mathbf{f}}^{(m)}\})$ with Eq.~\ref{eq:cgpflows:qf}
                    \State Update $q(\mathbf{Z})$ with Eq.~\ref{eq:cgpflows:qz}
                \EndWhile
                \State \# M-step
                \While{$J_L$ not converged}
                    \State Compute $J_{L}$ with Eq.~\ref{eq:cgpflows:likelihood}
                    \State \# Update parameters 
                    \State        $\{\theta^{(m)}\} \leftarrow \{\theta^{(m)}\}+\nabla_{\{\theta^{(m)}\}} J_{L}$
                    \State        $\boldsymbol{\beta} \leftarrow \boldsymbol{\beta}+\nabla_{\boldsymbol{\beta}} J_{L}$
                \EndWhile
            \EndWhile
    \end{algorithmic}
\end{algorithm}

\subsection{Predictive Distribution}
The predictive distribution of CGP-Flows is calculated using the predictive distribution of sparse OMGP and the transformation of Neural-ODE:
\begin{align}\label{eq:cgpflows:predition_distribution}
    \log p(\mathbf{y}_{*}  \mid {\mathbf{x}}_{*}, \bar{\mathbf{X}}, \mathbf X, \mathbf{Y}) 
    = \log p(\mathbf{l}_* \mid \mathbf{x}_*, \bar{\mathbf{X}}, \mathbf X, \mathbf{L}) 
    - \int_{t_0}^{t_1} \operatorname{Tr}\left(\frac{\partial {\mathbf g}_{\boldsymbol{\beta^{*}}}(\mathbf{l}_*(t), t, h(\mathbf x_*))}{\partial \mathbf{l}_*(t)}\right){\mathrm{d}} t,
\end{align}
where the variable $\mathbf L$ is obtained using an inverse transformation of Neural-ODE in Eq.~\ref{eq:cnf:backward}, and $p(\mathbf{l}_* \mid \mathbf{x}_*, \bar{\mathbf{X}}, \mathbf X, \mathbf{L})$ is the predictive distribution of a sparse OMGP used as the base distribution.
The analytical solutions of the predictive distribution of the sparse OMGP are described in Appendix~\ref{section:solutions}.

\subsection{Computational Complexity}
The computational complexity of CGP-Flows is primarily composed of two parts: the sparse OMGP and the Neural-ODE. Each contributes to the overall computational complexity in distinct ways.

\subsubsection{Computational Complexity of Sparse OMGPs}
The sparse OMGP component, which models the multimodal base distributions, involves two primary steps during the training process.

\begin{itemize}
    \item \textbf{E-step:} This step computes the variational distributions \( q(\{\bar{\mathbf{f}}^{(m)}\}) \) (as in Eq.~\ref{eq:analytical:qf}) and \( q(\mathbf{Z}) \) (as in Eq.~\ref{eq:analytical:qz}). Following the analysis in prior work~\cite{sasaki2021variational}, the computational complexity for each of these computations is \(\mathcal{O}(MNP^2)\), where \(N\) is the number of data points, \(M\) is the number of GP components, and \(P\) is the number of pseudo-inputs.

    \item \textbf{M-step:} In this step, the computational complexity of maximizing the lower bound of the log-likelihood \( J_\mathrm{base} \) is also \(\mathcal{O}(MNP^2)\).
\end{itemize}

\subsubsection{Computational Complexity of Neural-ODE}
The computational complexity of integrating the instantaneous change of variables in the Neural-ODE transformation (as in Eq.~\ref{eq:cgpflows:likelihood}) is \(\mathcal{O}((D+H)UE)\)~\cite{grathwohl2018ffjord}, where \(D\) is the dimension of 
 a variable $\mathbf y$, \(H\) is the dimension of condition extractor $h(\mathbf x)$, \(U\) is the largest number of units in the Neural-ODE's layers, and \(E\) represents the number of function evaluations (NFEs). The precise definition of NFEs is provided in Appendix~\ref{section:nfe}.

\begin{table}[t]
    \centering
    \caption{Computational complexity of each component in CGP-Flows. \(N\), \(M\), and \(P\) are the number of data points, GP components, and pseudo-inputs of sparse OMGP base distribution. \(D\), \(H\), \(U\), and \(E\) are dimension of variable $\mathbf y$, the dimension of condition extractor $h(\mathbf x)$, the largest number of units in layers, and represents the NFEs of Neural-ODE.}
    \label{tab:cgp_flows_complexity}
    \begin{tabular}{|c||cccc|}
        \hline
        & $q(\{\bar{\mathbf{f}}^{(m)}\})$ & $q(\mathbf{Z})$ & $\log J_\mathrm{base}$ & Transformation by Neural-ODE  \\
        \hline 
        Complexity & $\mathcal{O}\left(M N P^2\right)$ & $\mathcal{O}\left(M N P^2\right)$ & $\mathcal{O}\left(M N P^2\right)$ & \(\mathcal{O}((D+H)UE)\) \\
        \hline
    \end{tabular}
\end{table}

\subsubsection{Computational Complexity of CGP-Flows}
As summarized in Table~\ref{tab:cgp_flows_complexity}, the computational complexity of CGP-Flows comprises two main components. First, the E-step computes variational distributions \( q(\{\bar{\mathbf{f}}^{(m)}\}) \) and \( q(\mathbf{Z}) \), each requiring computational complexity of \(\mathcal{O}(MNP^2)\).
Second, the M-step computes the lower bound of sparse OMGP \(\log J_\mathrm{base}\) and transformation by Neural-ODE, which has a combined computational complexity of \(\mathcal{O}(MNP^2 + (D+H)UE)\).

Compared to NGGP, which has a computational complexity of \(\mathcal{O}(N^3 + (D+H)UE)\), CGP-Flows is more efficient for larger datasets due to the use of sparse approximations.
Furthermore, while both CGP-Flows and NGGP share the same computational complexity for the Neural-ODE component, CGP-Flows significantly reduces the NFEs by leveraging multimodal base distributions.
This reduction in NFEs further improves the overall training efficiency of CGP-Flows.

\section{Simulations}
We validated the effectiveness of CGP-Flows by using this idea as the policy model of behavior cloning for two simulated tasks: ball-shooting and robotic object-swiping.
In the former, we evaluated the performance of the CGP-Flow model in learning local discontinuous and multimodal policies and compared CGP-Flow with NGGP and OMGP.
For the object-swiping task, we investigated the effects of the CGP-Flow parameters, e.g., the number of GP mixtures, the tolerance, and the neural ODE structure, on the policy performance and training efficiency.

\begin{figure}[t]
    \centering
    \begin{minipage}[b]{0.48\hsize}
        \centering
        \includegraphics[width=1.0\hsize, trim=120 0 120 0, clip]{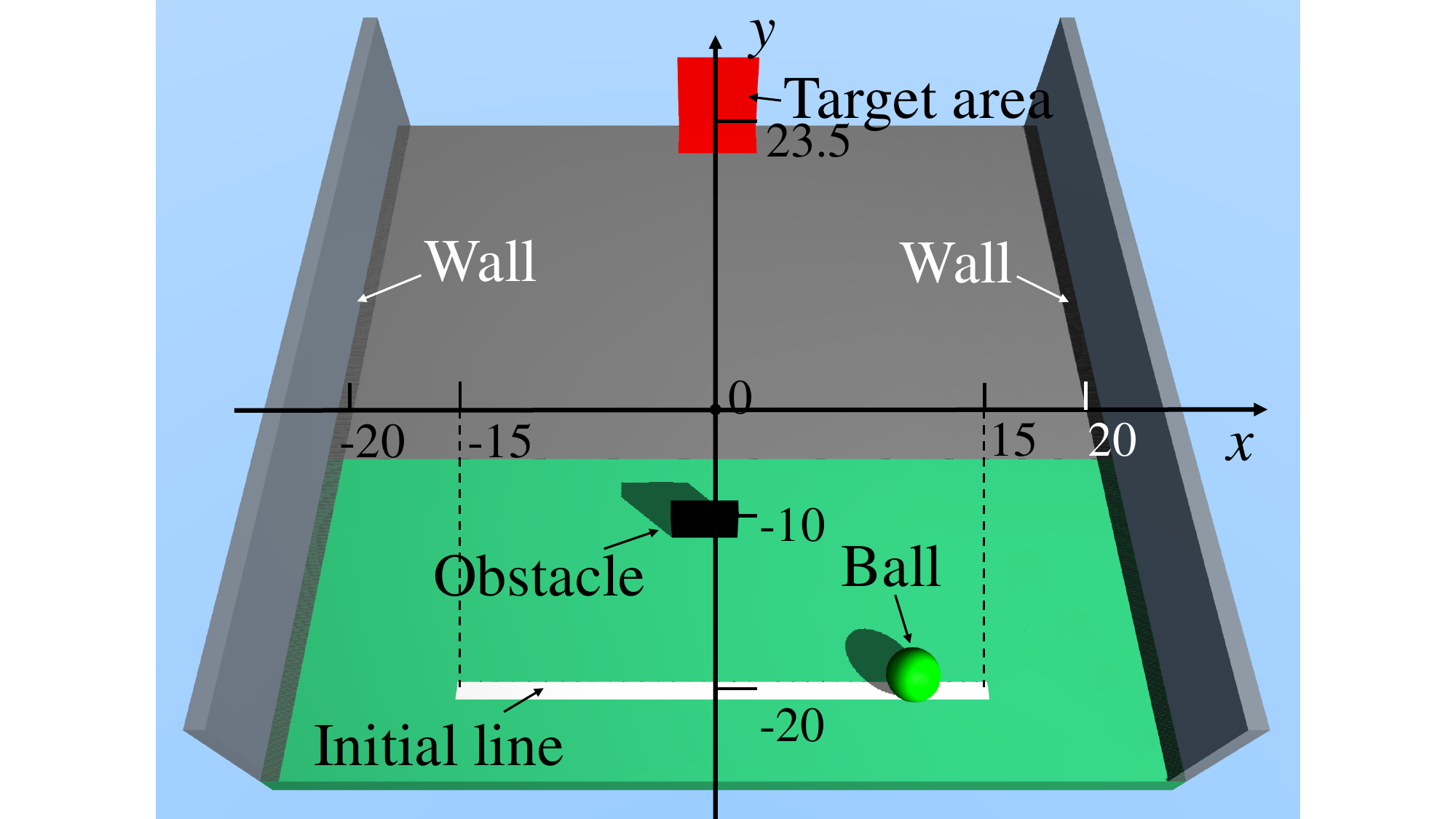}
        \subcaption{}
        \label{fig:shooting:environment:a}
    \end{minipage}
    \begin{minipage}[b]{0.48\hsize}
        \centering
        \includegraphics[width=1.0\hsize, trim=120 0 120 0, clip]{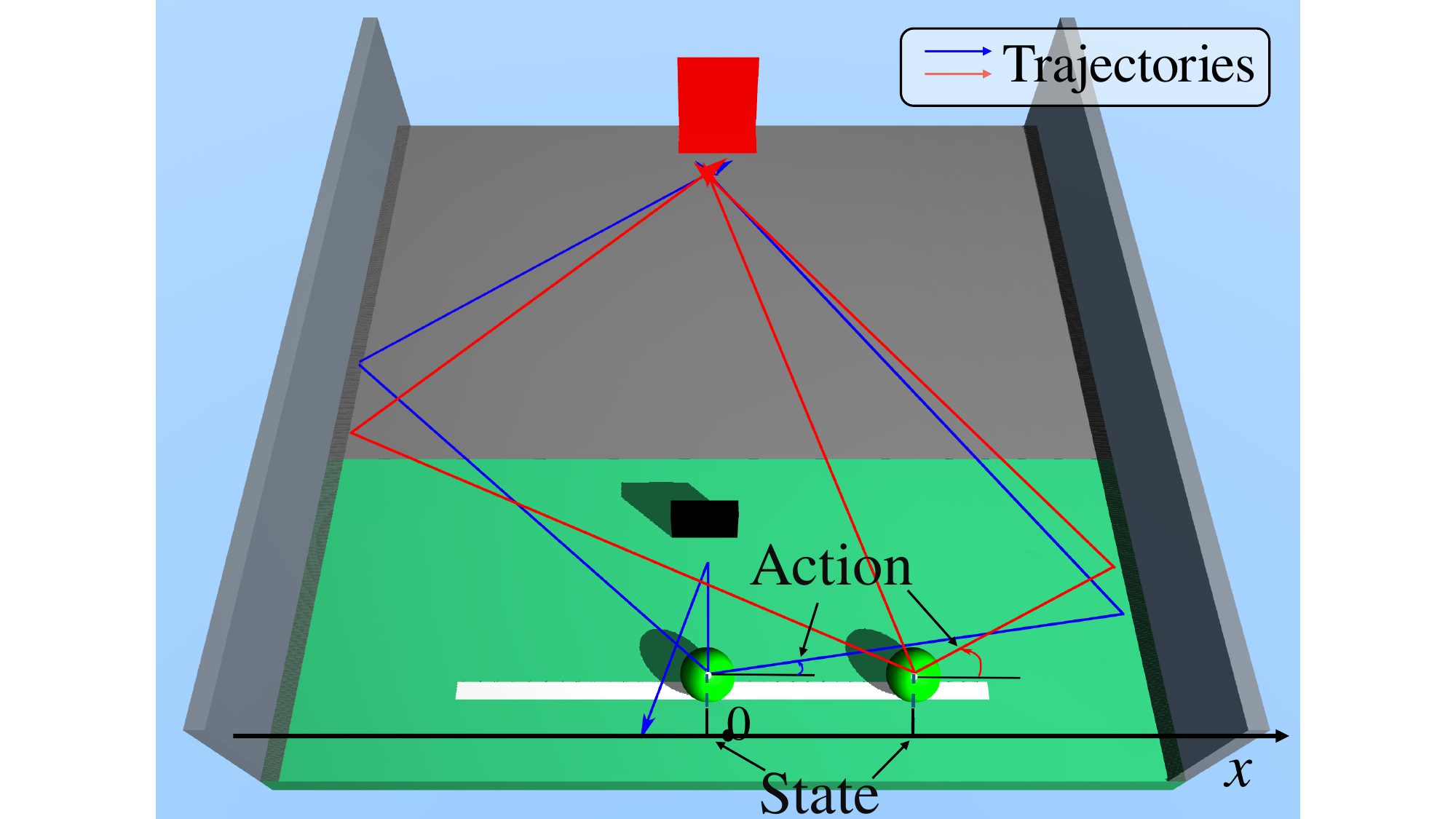}
        \subcaption{}
        \label{fig:shooting:environment:b}
    \end{minipage}
    \caption{Ball-shooting simulation environment: (a) Overview of environment. Ball is first placed randomly on initial line to reach the red target area. An obstacle is placed in the center of the environment. (b) Ball's pathway to target area. There are three paths for reaching the target area; however, at some initial positions, an obstacle directly limits path directly to target area.}
    \label{fig:shooting:environment}
\end{figure}

\begin{figure}[t]
    \centering
    \begin{minipage}[t]{0.49\linewidth}
        \centering
        \includegraphics[width=0.46\linewidth]{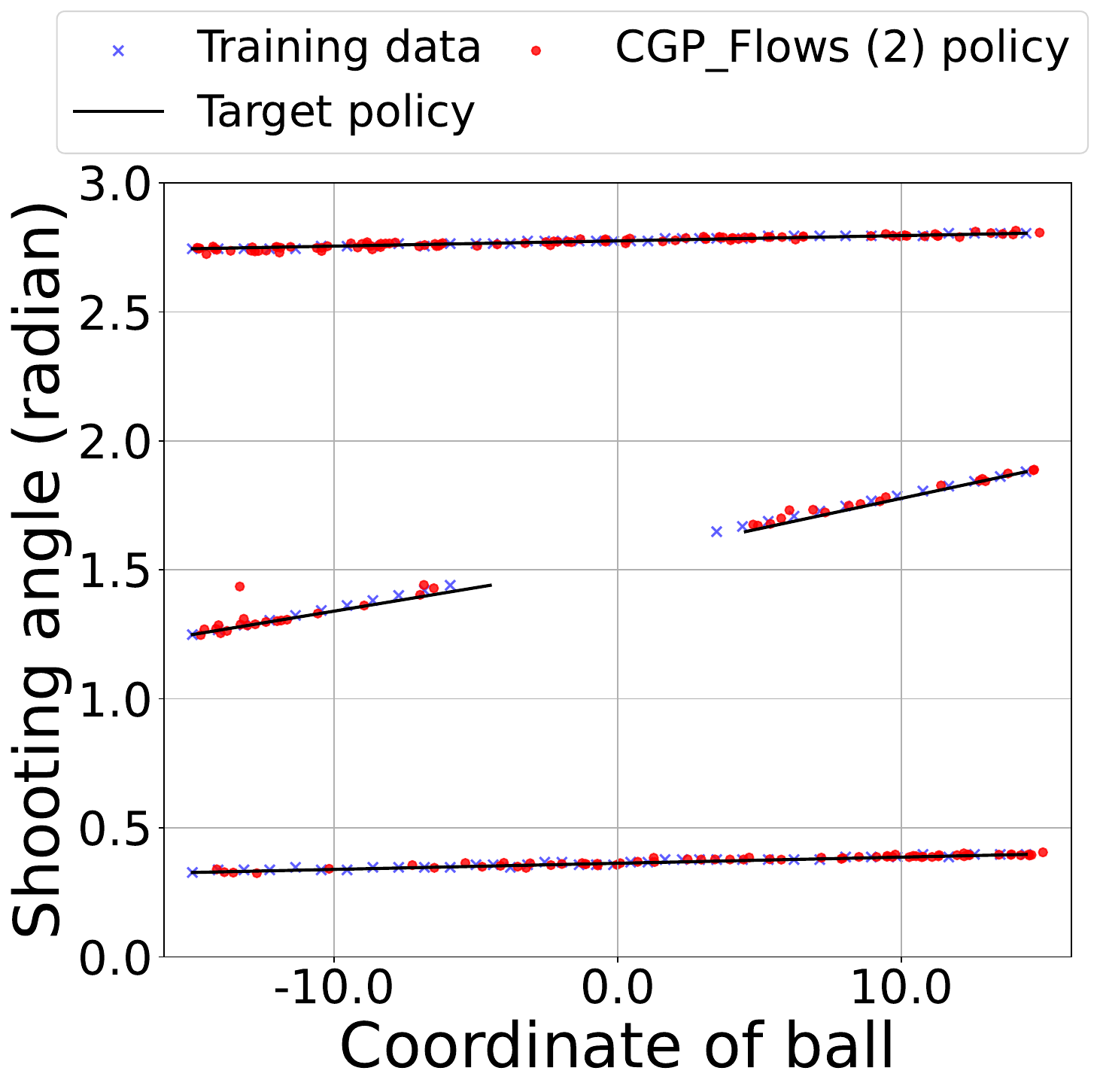}
        \includegraphics[width=0.52\linewidth]{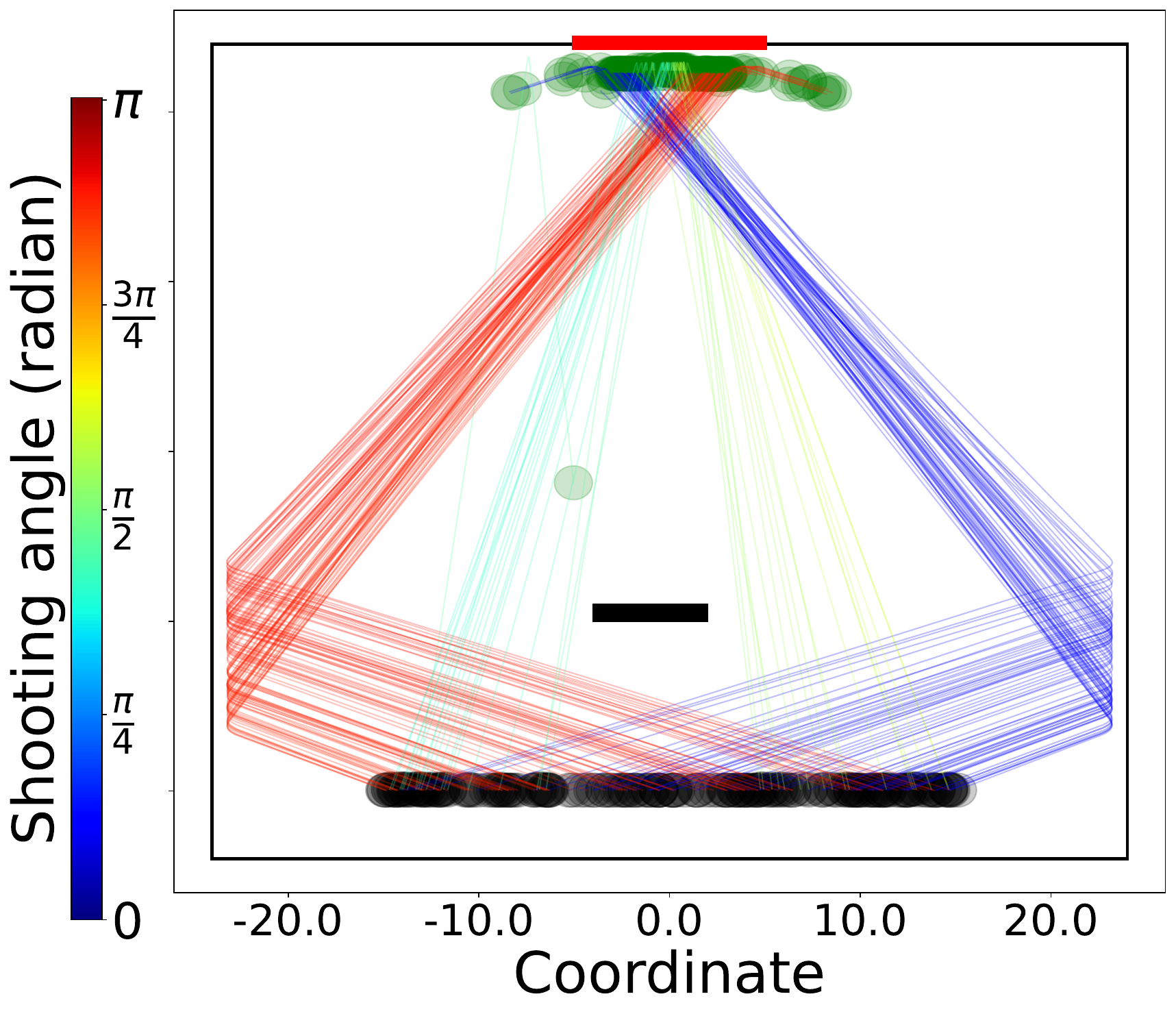}
        \subcaption{CGP-Flow policy with $M=2$}
        \label{fig:shooting:plot:a}
    \end{minipage} 
    \begin{minipage}[t]{0.49\linewidth}
        \centering
        \includegraphics[width=0.46\linewidth]{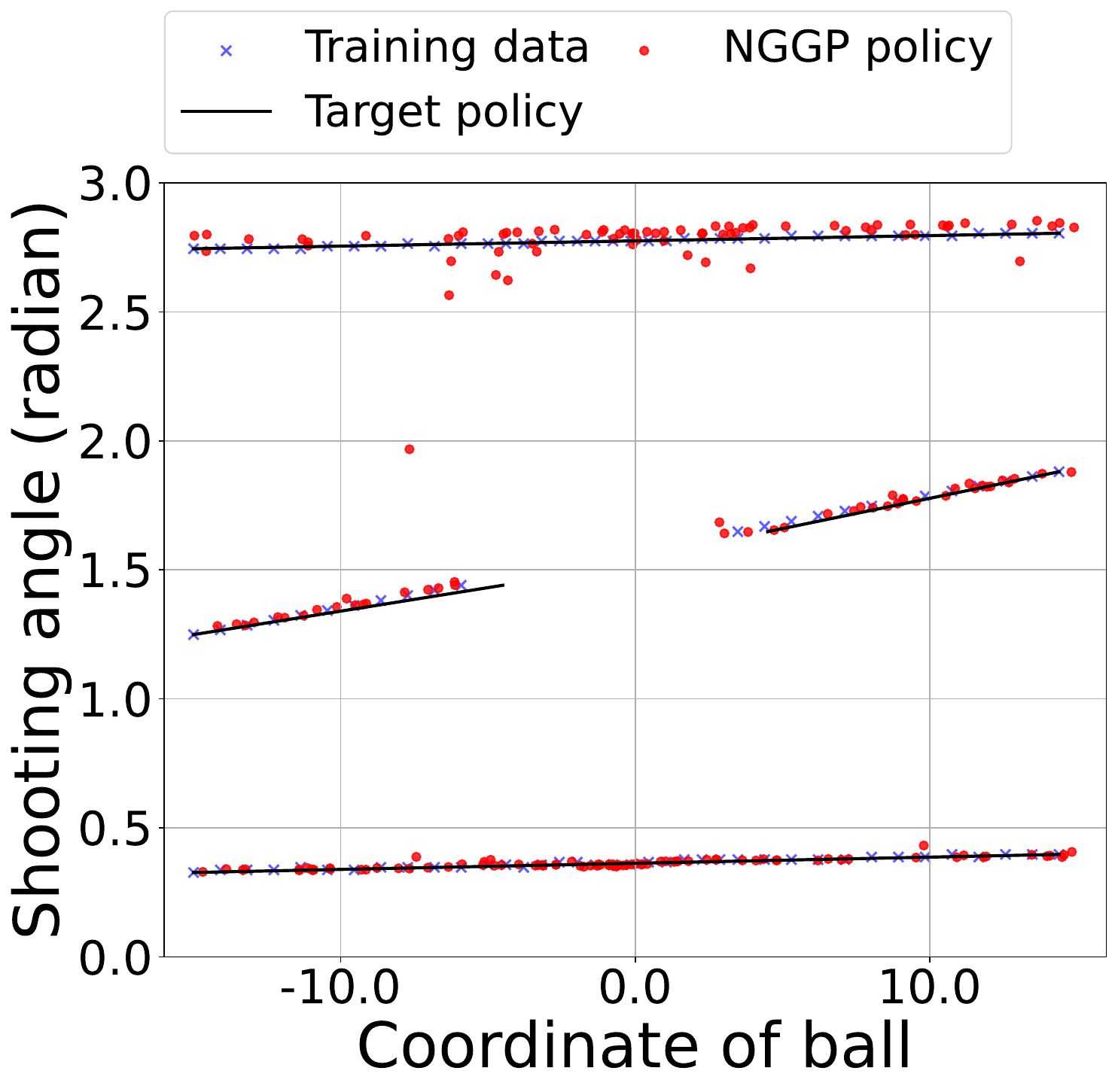}
        \includegraphics[width=0.52\linewidth]{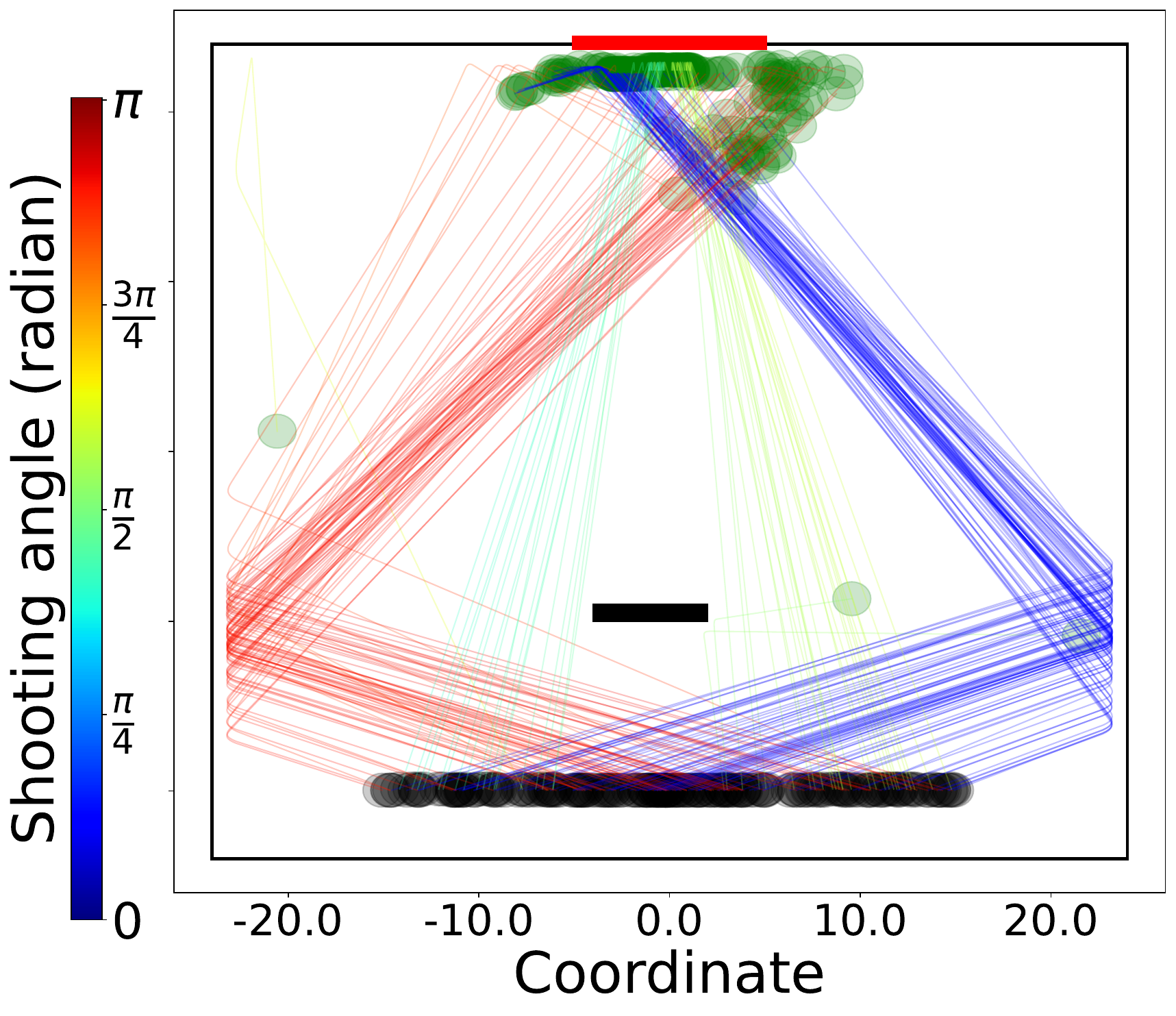}
        \subcaption{NGGP policy}
        \label{fig:shooting:plot:b}
    \end{minipage} \\
    \begin{minipage}[t]{0.49\linewidth}
        \centering
        \includegraphics[width=0.46\linewidth]{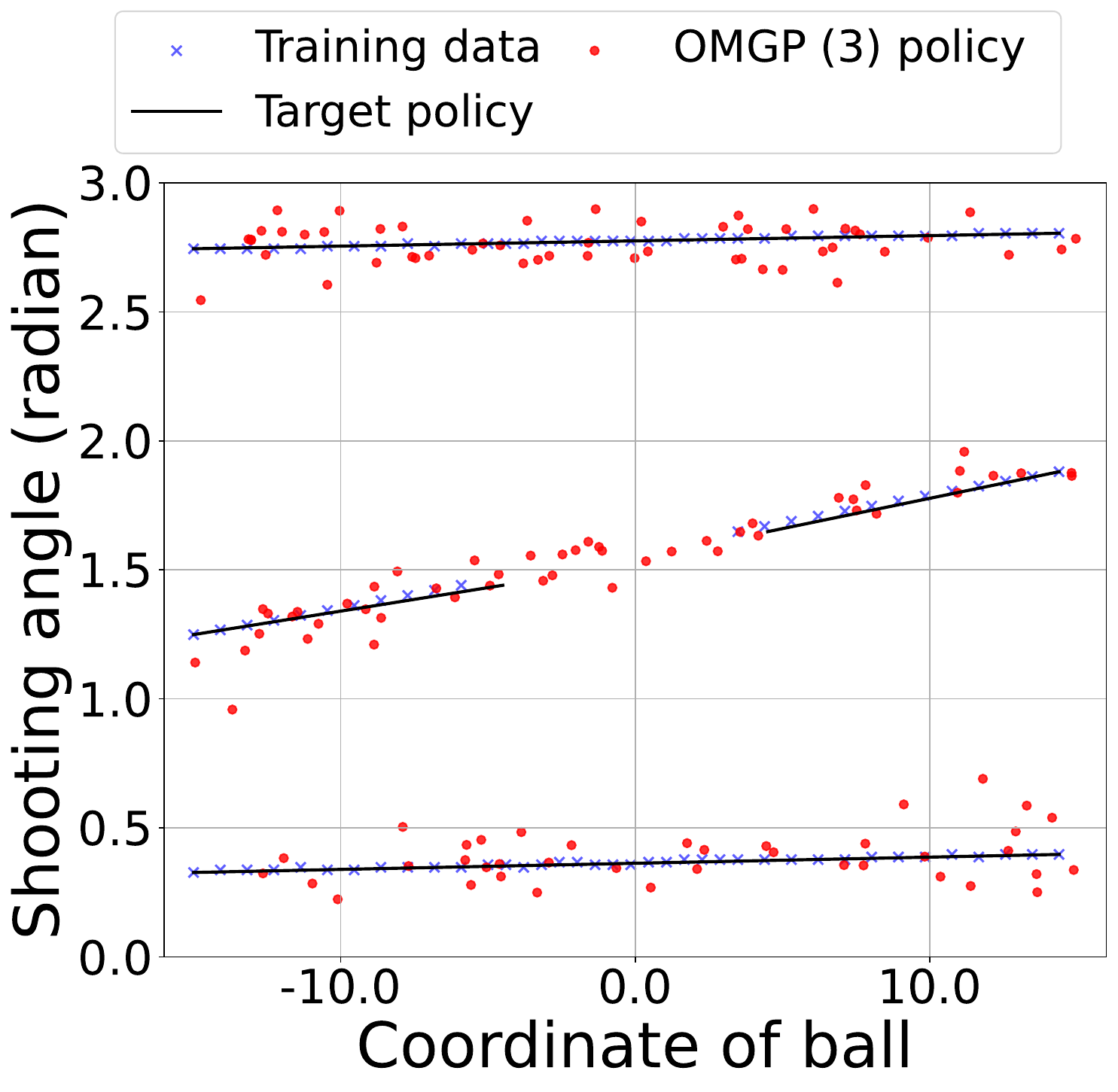}
        \includegraphics[width=0.52\linewidth]{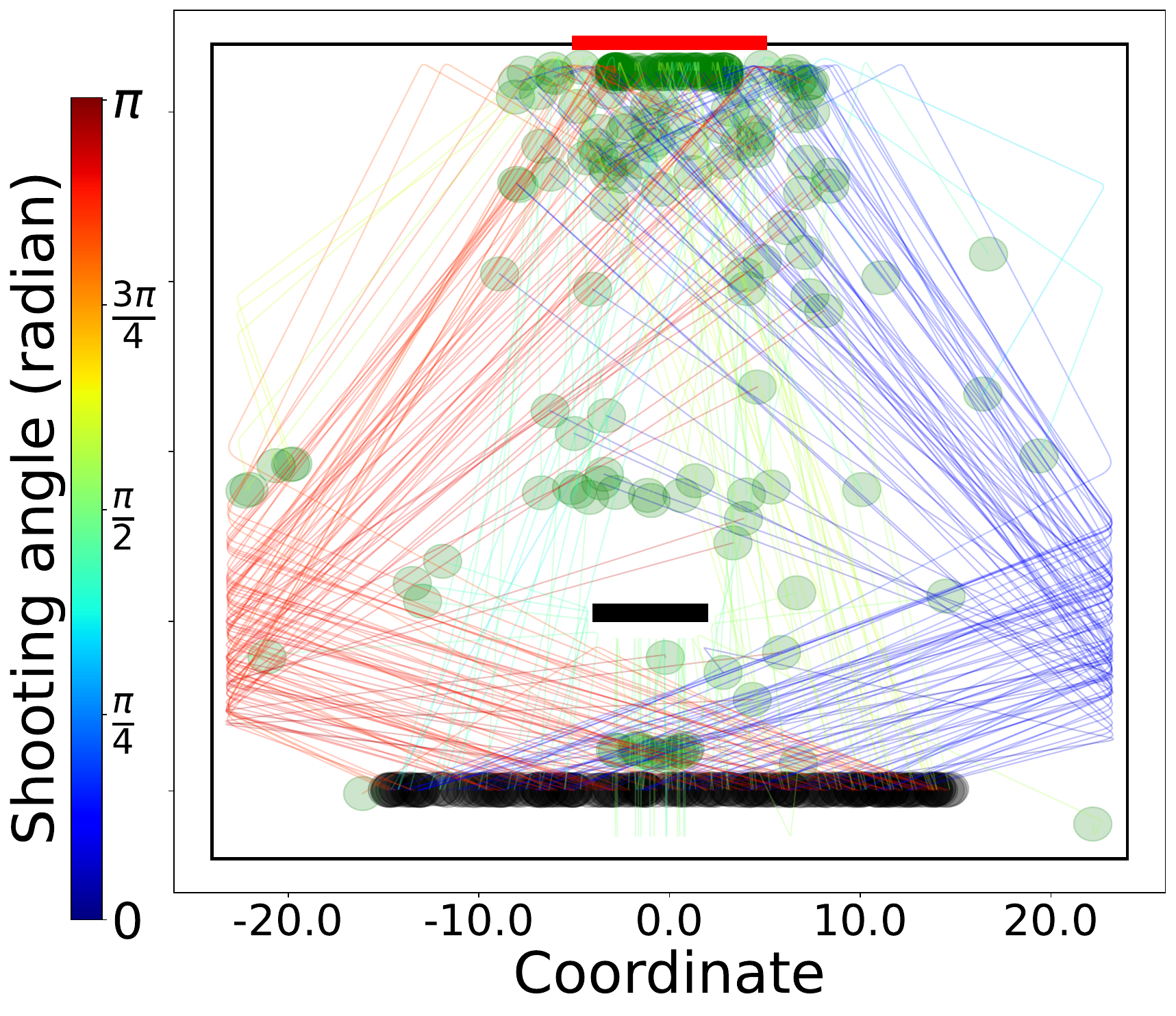}
        \subcaption{OMGP policy with $M=3$}
        \label{fig:shooting:plot:c}
    \end{minipage}\\
    \caption{Visualization of learned policy by each policy: Left figures show the prediction of learned policy. Red markers are sampled actions from learned policy. Black line is expert policy for task. Blue markers are training data. Right figures show ball paths generated from predicted actions. Bottom, top, and middle lines indicate initial line, target area, and obstacle. Line color indicates value of the action. (a), (b), and (c) results CGP-Flow policy with $M=2$, NGGP policy, and OMGP policy with $M=3$.}
    \label{fig:shooting:plot}
\end{figure}

\subsection{Ball-shooting Task}
\subsubsection{Simulation Settings}
The ball-shooting task seeks to shoot a ball from an initial line to a target area.
We created a 25 cm $\times$ 25 cm, simulation (Fig.~\ref{fig:shooting:environment}) with a Mujoco simulation engine \cite{6386109}.
A 0.1 kg, 1.5 cm radius ball is shot from the initial line at coordinate $y=-20$ where the x-coordinates range from -15 to 15.
The ball-shooting's target area is located at $x=0$, $y=23.5$; it is sized 6 cm by 3 cm.
An obstacle is placed at coordinates $x=0$, $y=-10$.

The ball-shooting task begins by randomly placing the ball on the initial line.
State $s$ and action $a$ of the task are defined as the first position on the initial line and the ball's launch angle.
The ball's shooting velocity is set at 1.5 m/s.
One state-action pair is obtained by executing this task once.
As illustrated in Fig.~\ref{fig:shooting:environment:b}, the number of ball paths to the target area depends on the state.
If the ball is near the center of the environment, there are two possible paths to the target area; otherwise, there are three.
The number of paths the ball can take and the changes in this number appear as multimodality and discontinuity in policy learning.

\subsubsection{BC Settings}
To learn a control policy for the ball-shooting task, the task were executed 100 times to collect learning data $\mathcal D = \{s_n, a_n\}_{n=1}^{100}$.
The collected states were determined by a random position on the initial line based on the environmental settings, and the collected actions are the ball's launch angles that generate one of the paths to the target position.

\subsubsection{Comparison Methods}
We compared the performance with CGP-Flow by employ two stochastic models as policy models: OMGP, and NGGP.
We set the modality to the base distribution of the CGP-Flow as $M=2$ and the OMGP as $M=3$.
The NGGP and CGP-Flow models utilize the same cCNF structure, consisting of a single block with three layers, each containing 256 units and employing SiLu~\cite{elfwing2018sigmoid} as the activation function.
Both the CGP-Flow and NGGP models apply \textit{dopri5} as an ODE solver; the tolerances are set to $10^{-5}$.
The initial hyperparameters of OMGP $\theta^{(m)}$ and $\sigma$ are set to $\theta^{(m)}=0.5$, $\sigma=0.05$.
NGGP's hyperparameter is set at $\theta=0.5$.
Initial hyperparameters $\theta^{(m)}$ and $\sigma$ of the CGP-Flow are set at $\theta^{(m)}=0.5$ and $\sigma=0.1$.
Since the state input just one-dimensional, we used the state as the condition for the CGP-Flow and NGGP, and thus the condition extractor is defined as $h(s_n)=s_n$.

The Adam optimizer was employed in the training of the NGGP model, as well as during the M-step of both the OMGP and CGP-Flow models.
We trained the OMGP with $M=3$ and the CGP-Flow with $M=2$ models with an EM-like algorithm consisting of 10 E-steps and 10 M-steps.
During each M-step, the hyperparameters and parameters were iteratively updated for 500 steps.
For both the OMGP and CGP-Flow models, the initial learning rate was set at 2e-3 and reduced by $10\%$ following each M-step to facilitate better convergence.
The NGGP model's hyperparameters and parameters were optimized with 5000 steps.
NGGP's initial learning rate, which was also set at 2e-3, was reduced by $10\%$ every 500 steps.
The performance of each method was evaluated in terms of the success rate and the computation time required for training and prediction.
The success rate is calculated by executing the task 200 times with the trained policy.
The specifications of the computer used in the experiment are shown in Appendix~\ref{appendix:computational:settings}.

\subsubsection{Result}
Table~\ref{tab:shooting:success_rate} shows the success rate and the computation time in the training and the prediction of each policy model.
The CGP-Flow model outperformed both the NGGP and OMGP models in terms of success rate and its computation time was similar to NGGP's.

Learned policies are shown in  Fig.~\ref{fig:shooting:plot}.
According to the sampled actions, a shot ball is heading towards the target area.
The CGP-Flow learned the policy with high accuracy, and actions sampled from the learned policy are nearly plotted on the expert policy's line.
The learned policy captures the multimodality of the expert policy and the discontinuity of its middle line.
The policy learned with NGGP captures the expert policy similarly to the policies learned with the CGP-Flow.
However, the variance of the sampled actions exceeds that of the CGP-Flow policy.
The larger variance in the predictions of the NGGP policy lowers the success rate.
The OMGP policy has a larger variance in the sampled actions than the other policies and learns a continuous relationship to the discontinuous middle lines.
The results of additional validation using the NGGP with original settings described in \cite{sendera2021non} are shown in Appendix~\ref{appendix:baseline:nggp}.

We compared the success frequency of CGP-Flows with NGGPs and OMGPs using chi-square tests~\cite{franke2012chi}.
The results show that the difference in success frequency between CGP-Flows and NGGPs is significant, with $\chi^2 = 14.59$ and $p = 1.3 \times 10^{-4} \leq 1.0 \times 10^{-3}$.
Similarly, the difference between CGP-Flows and OMGPs is significant, with $\chi^2 = 35.02$ and $p = 3.3 \times 10^{-9}\leq 1.0 \times 10^{-3}$.
These results demonstrate that CGP-Flows outperform both NGGPs and OMGPs with significant differences in success frequency, confirming the effectiveness of our proposed method in capturing complex, multimodal, and discontinuous policies.

\begin{table}[!t]
    \centering
    \caption{The performance of learned policy by each method on ball-shooting task. Each policy is evaluated based on the success rate of ball-shooting and the computation time in training and prediction.}
    \label{tab:shooting:success_rate}
    \begin{tabular}{|c||c|c|c|} \hline
                      & \multicolumn{3}{c|}{Models}        \\ \hline
                      & OMGP       & NGGP     & CGP-Flow  \\ 
                      & with $M=3$ &          & with $M=2$ \\ \hline \hline
        Success rate  & 38.2$\%$   & 56.3$\%$ & 81.0$\%$   \\ \hline
        Computation time during training (s)
                      & 17         & 1450     & 1540     \\ \hline
        Computation time during prediction (ms) 
                      & 1.00       & 74.0     & 49.0       \\ \hline
    \end{tabular}
\end{table}

\subsection{Object-swiping Task}

\begin{figure}[t]
    \centering
    \begin{minipage}[b]{0.3\linewidth}
        \centering
        \includegraphics[width=0.9\linewidth]{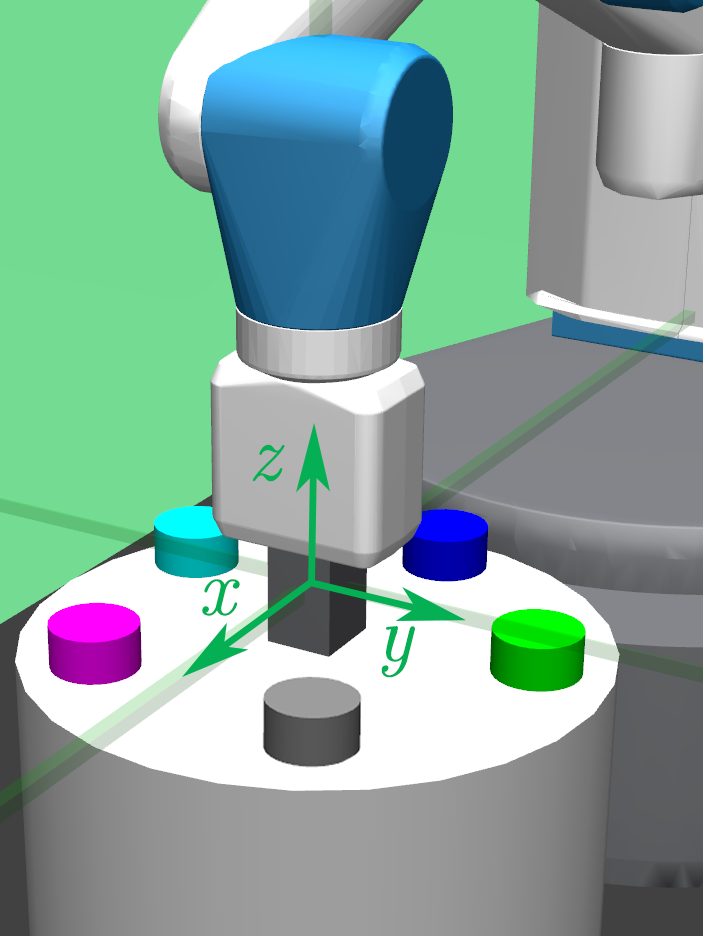}
        \subcaption{}
        \label{fig:sim:init}
    \end{minipage}
    \begin{minipage}[b]{0.3\linewidth}
        \centering
        \includegraphics[width=0.9\linewidth]{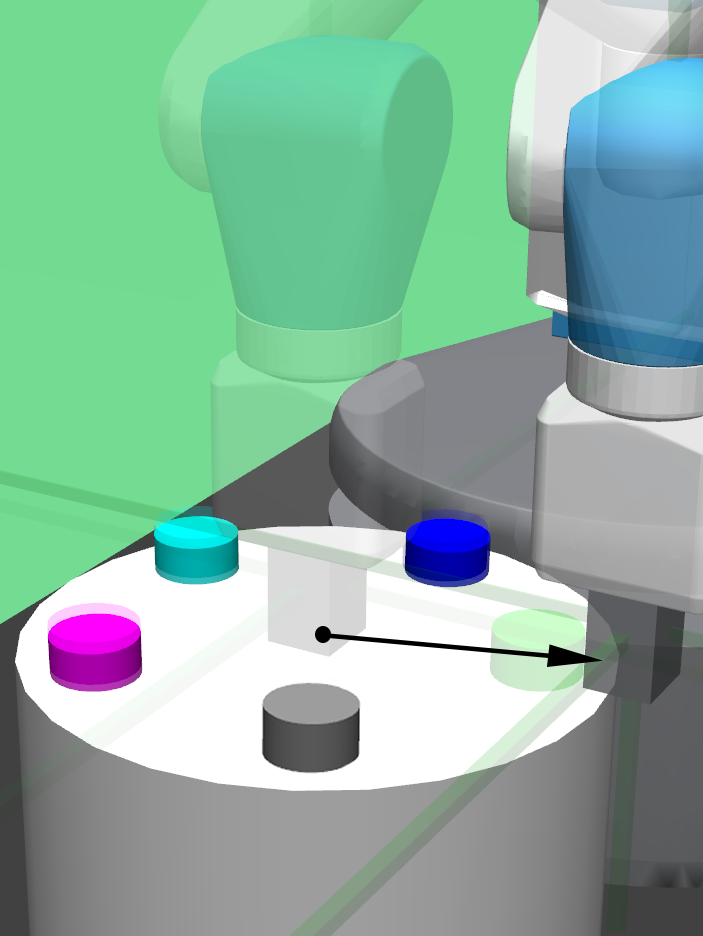}
        \subcaption{}
        \label{fig:sim:swipe}
    \end{minipage}
    \begin{minipage}[b]{0.3\linewidth}
        \centering
        \includegraphics[width=0.9\linewidth]{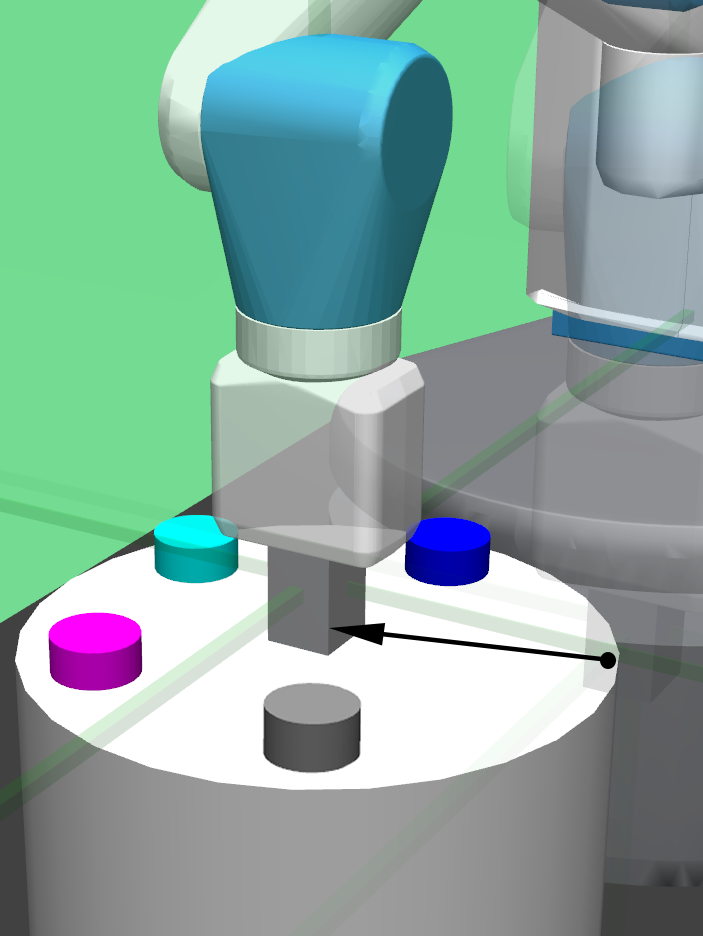}
        \subcaption{}
        \label{fig:sim:return}
    \end{minipage}
    \caption{Object-swiping task environment: (a) Initial position of robot arm and five objects.
    Robot's motion to swipe an object: Robot arm moves to target position specified by action (b) and returns to its initial position (c).}
    \label{fig:swiping:sim}
\end{figure}

\subsubsection{Simulation Settings}
The object-swiping task is implemented in a simulation environment on the MuJoCo physics engine \cite{6386109} (Fig.~\ref{fig:swiping:sim}).
The task's aim is to swipe all the objects on the table.
The table is a cylinder with a radius of 20 cm and a height of 19 cm.
Five identical objects have a radius of 3 cm and a height of 1.5 cm.
They are placed 15 cm from the center of the table and arranged at every 72 degrees.
Their initial positions are fixed (Fig.~\ref{fig:sim:init}).

A state is represented by vector ${\mathbf{s}}_n=\{x^{o}_n, y^{o}_n\}_{o=1}^{5}$, where $x^{o}_n$ and $y^{o}_n$ denote the position on X and Y axes of the $o$-th object.
When it is swiped down from the table, the elements of the state are set to $x^{o}_n = y^{o}_n = 0$.
The action is ${\mathbf{a}}_{n}=\{a_{n}^{x}, a_{n}^{y}\}$, where $a_{n}^{x}$ and $a_{n}^{y}$ indicate a target position on the X and Y axes.
Therefore, the robot arm's motion is limited to the plane on the table.
The values of $a^x_n$ and $a^y_n$ are limited from $-10$ to $10$.
The robot arm motion is executed according to an action in the following three steps:
i) The robot arm is placed in its initial position, which is the center of the table (Fig.~\ref{fig:sim:init}).
ii) It moves to a target position dictated by action ${\mathbf a}_{n}$ (Fig.~\ref{fig:sim:swipe}).
iii) It returns to its initial position (Fig.~\ref{fig:sim:return}).
We define the above three moves as one step of the task whose aim is to swipe down all five objects by five steps.
The object-swiping task is terminated after the robot has taken action in five steps regardless whether five objects were swiped down from the table or not.

The complexity of this task emerges from its multiple optimal actions, which correspond to the remaining number of objects on the table.
Since this number changes with each successful swiping action, the modality of the swiping policy also varies accordingly.

\subsubsection{BC Settings}
We designed an expert policy to collect expert demonstration data for BC.
The expert policy computes action ${\mathbf{a}}_n=\{a_n^x, a_n^y\}$ using a randomly selected unswiped object position $\{x_n^o, y_n^o\}$:
\begin{align}
    a_n^x &= d(x_n^o-C_x)/r+\varepsilon_x, \nonumber \\
    a_n^y &= d(y_n^o-C_y)/r+\varepsilon_y, \nonumber
\end{align}
where $\varepsilon_x, \varepsilon_y \sim {\mathcal N}(0, \sigma)$, $C_x$ and $C_y$ are the initial positions of the robot arm in the X and Y axes, and $r = \sqrt{(x^{o}_n-C_{x})^{2}+(y^{o}_n-C_{y})^2}$, $o$ is an index of unswiped objects,  $\sigma=0.2$.
$d$ is the length moved by the robot arm moves, set to $d=8$.
The expert policy executes the simulation 50 times to collect the training data.
Since the task terminated in five steps, training data ${\mathcal D}$ have 250 state-action pairs ${\mathcal D}=\{{\mathbf{s}}_n, {\mathbf{a}}_n\}_{n=1}^{250}$.

\subsubsection{Comparison Methods}
We confirmed the effect of the CGP-Flow settings on the performance of the learned policy.
To compare the modalities of the base distribution, we varied the modalities of CGP-Flows from $M=2$ to 5, and compared them with NGGP.
For each CGP-Flow and NGGP model, we employed $\textit{dopri5}$ as an ODE solver, and its parameters \textit{atol} and \textit{rtol} are set from 1e-6 to 1e-2.
The structures of CNFs are set to a single block three-layer network and a single block five-layer network, each with eight units per layer.
The activation function of each network is Swish~\cite{ramachandran2017searching}.
The initial hyperparameter of NGGP $\theta$ is set to $\theta=1.0$.
All the CGP-Flows share the same initial hyperparameters, where $\theta^{(m)}=1.0$ and $\sigma=0.1$.
We created condition extractor $h(\cdot)$ that converts a ten-dimensional state to an integer value from 1 to 31 since the number of cases with at least one object on the table is 31.
The specifications of the computer used in the experiment are shown in Appendix~\ref{appendix:computational:settings}.

The leand policy's performance is evaluated using the score, which is the number of swiped objects.
In each simulation, one successfully swiped object awards $1.0$ point to the policy; the maximum score is $5.0$ points.
We compared the score, the NFEs, and the computation time on the training of the NGGPs and the CGP-Flows under different settings of cCNF structures, mixture numbers of GPs, and tolerance.

\subsubsection{Result}
\begin{figure}[t]
    \centering
    \begin{minipage}[b]{0.49\hsize}
        \centering
        \includegraphics[width=0.7\hsize]{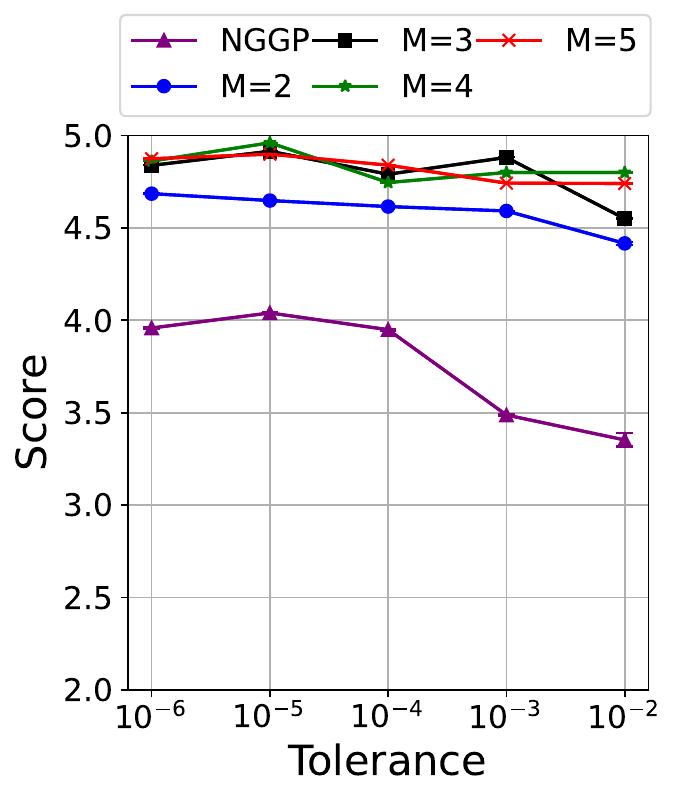}
        \subcaption{Five-layer network}
        \label{fig:swiping:sucess rate:a}
    \end{minipage}
    \begin{minipage}[b]{0.49\hsize}
        \centering
        \includegraphics[width=0.7\hsize]{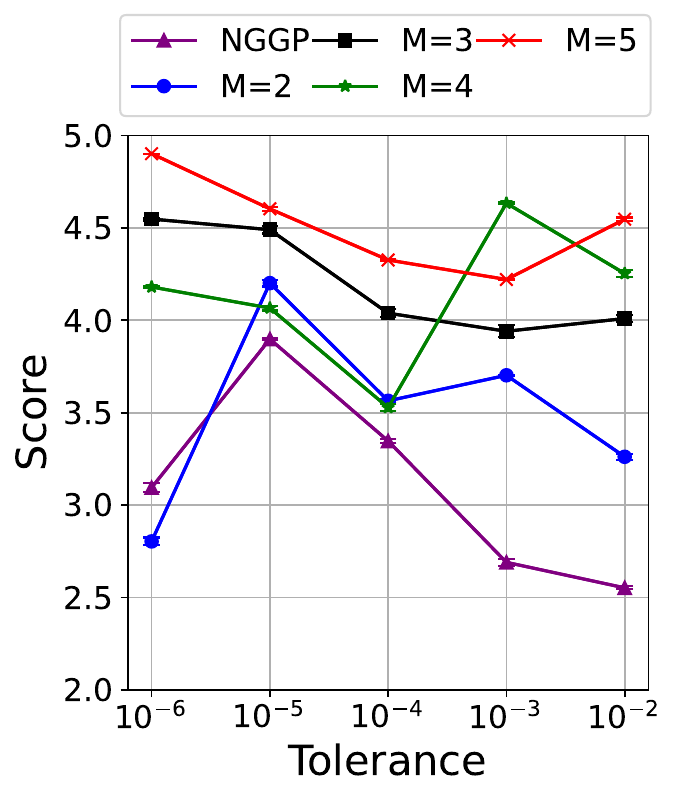}
        \subcaption{Three-layer network}
        \label{fig:swiping:sucess rate:b}
    \end{minipage}
    \vspace{-3mm}
    \caption{Success rates of task achieved by CGP-Flow policies with different settings. For models with identical cCNF structures, larger $M$ value positively impacts success rate. Results represent average of 10 independent training sessions.}
    \label{fig:swiping:sucess rate}
\end{figure}
\begin{figure}[t]
    \centering
    \begin{minipage}[b]{0.49\hsize}
        \centering
        \includegraphics[width=0.7\hsize]{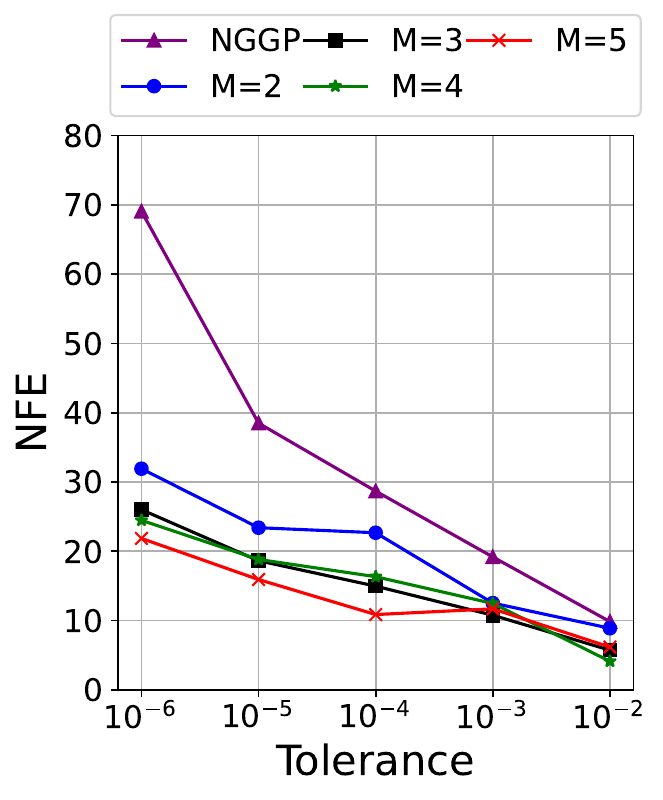}
        \subcaption{Five-layer network}
        \label{fig:swipping:nfe:a}
    \end{minipage}
    \begin{minipage}[b]{0.49\hsize}
        \centering
        \includegraphics[width=0.7\hsize]{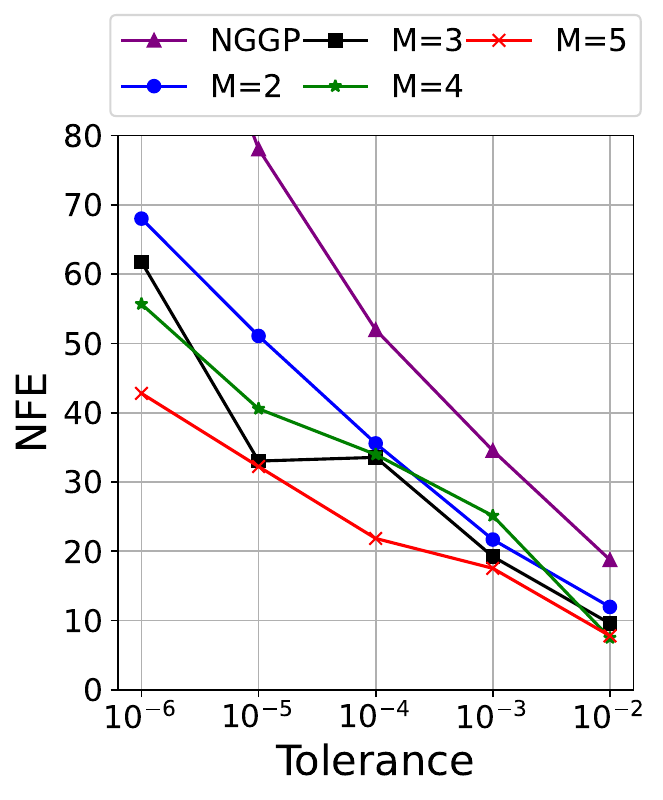}
        \subcaption{Three-layer network}
        \label{fig:swipping:nfe:b}
    \end{minipage}
    \vspace{-3mm}
    \caption{NFE counts of ODE solvers when CGP-Flow policies are trained under different settings. For models with identical cCNF structures, larger $M$ value reduces NFE count. Results represent average of 10 independent training sessions.}
    \label{fig:swipping:nfe}
\end{figure}

We show the object-swiping scores by learned CGP-Flow and NGGP policies in Fig.~\ref{fig:swiping:sucess rate}.
The CGP-Flow models outperformed the unimodal base distributions even with mixture number $M=2$. The five-layer networks generally had higher scores.
However, the results of the three-layer networks with $M\geq3$ show that base distributions with higher modality outperform the unimodal base distributions (NGGP) when the neural network's structure is overly simplistic for the given task.

Fig.~\ref{fig:swipping:nfe} shows the training NFE of the NGGP and CGP-Flows policy models.
Owing to the fact that a Neural-ODE with more parameters can learn an ODE that is easier for the solver, models with five-layer networks generally require fewer NFE. However, when comparing models with the same cCNF structure, the NFE are inversely related to the $M$ value of the CGP-Flow models.

Table \ref{tab:time:cost} shows the training time cost of the CGP-Flow policies.
Although the number of modalities of the base distributions affects the computational complexity, a smaller NFE count significantly reduces the training time of the CGP-Flow models.

\begin{figure}[t]
    \centering
    \begin{minipage}[b]{0.49\linewidth}
        \centering
        \includegraphics[width=0.605\linewidth]{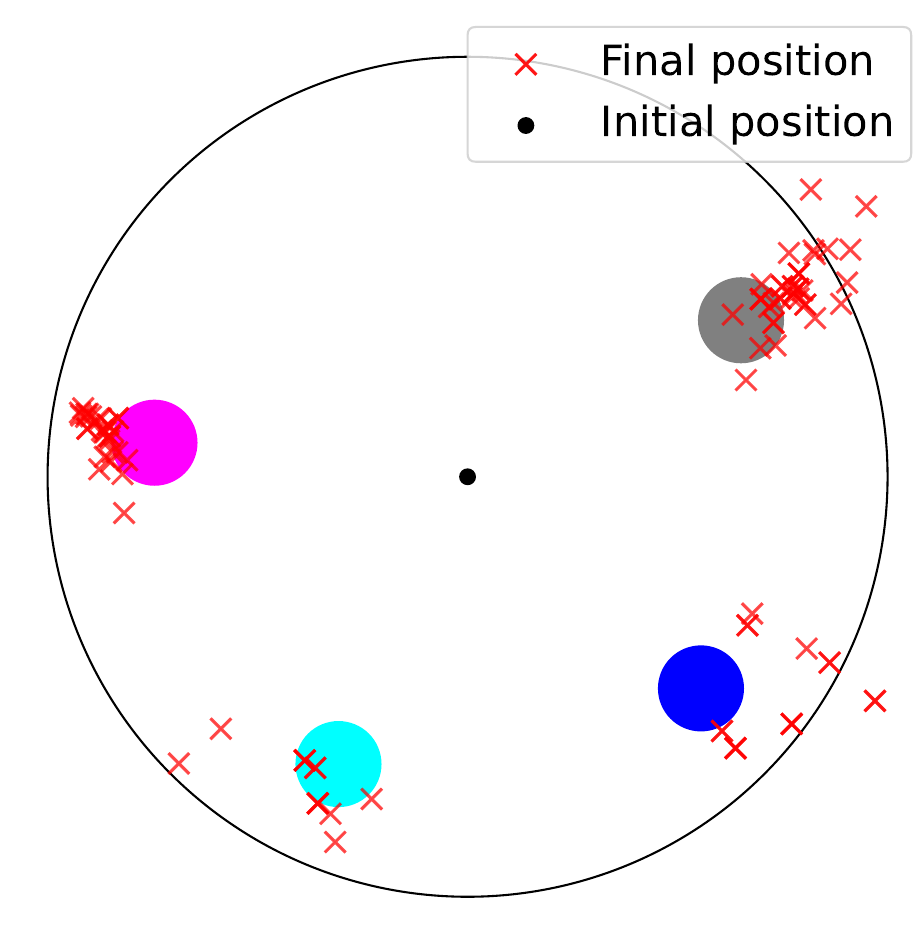}
        \subcaption{CGP-Flow ($4$)}
        \label{fig:swiping:action:a}
    \end{minipage}
    \begin{minipage}[b]{0.49\linewidth}
        \centering
        \includegraphics[width=0.6\linewidth]{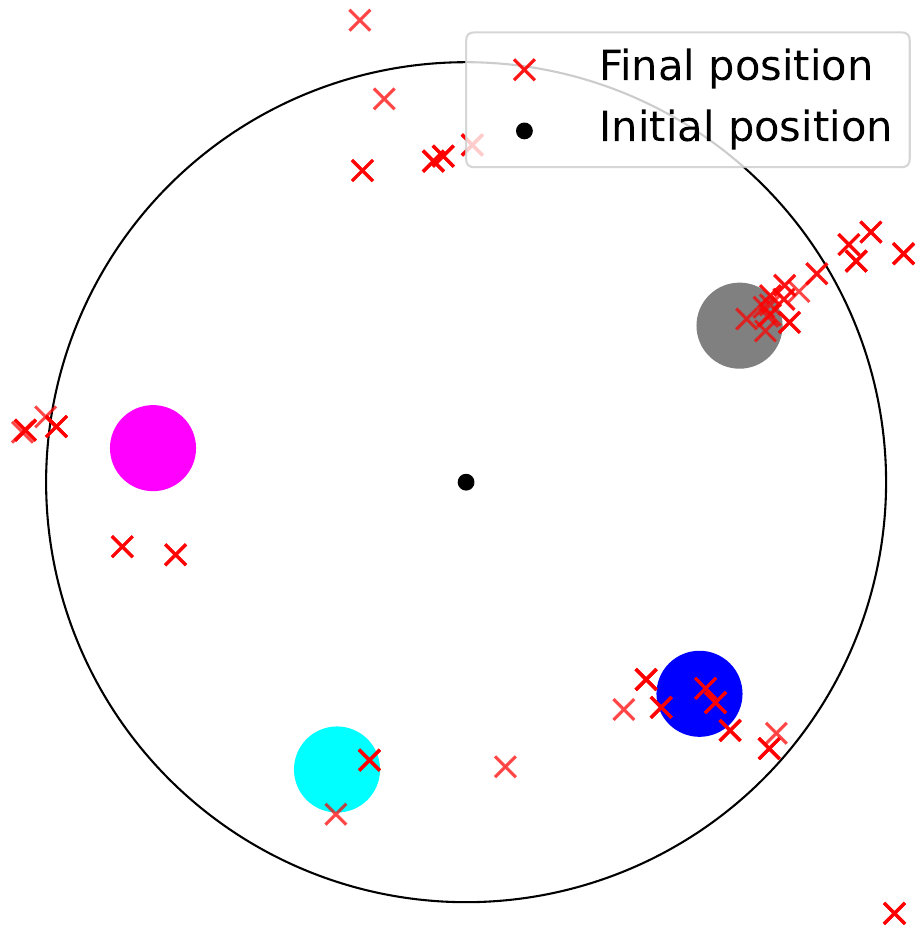}
        \subcaption{NGGP}
        \label{fig:swiping:action:b}
    \end{minipage}
    \caption{Sampled actions of object-swiping simulation: With four objects are on the table, actions sampled by three policy models are illustrated. Four colored circles represent objects, red marks indicate robot's final position, and large black circle represents the table. With greater modality in base distributions, policy model achieves better predictions.}
    \label{fig:swiping:action}
\end{figure}

To visualize the predictive output of the policy models, we plotted 100 actions of three policy models at a state where one object is swiped down from the table and four objects remain (Fig.~\ref{fig:swiping:action}). All models have five-layer networks; their tolerances are set to 1e-5.
Under this situation, the final position of the policy models should be located at the objects remaining on the table.
However, the changing modality can affect the accuracy of the predictive output of the policy models.
Fig.~\ref{fig:swiping:action:a} shows that the final positions of the CGP-Flow with a $M=4$ model are mostly correct.
Fig.~\ref{fig:swiping:action:b} shows that the NGGP have incorrect final positions. 
This comparison confirms that the CGP-Flow with $M=4$ is less affected by the changing modality than the NGGP.

\begin{table}[t]
    \centering
    \renewcommand{\arraystretch}{1.1}
    \caption{Training time cost for the CGP-Flow and the NGGP models with different base distributions and tolerances, measured in seconds.}
    \label{tab:time:cost}
    \begin{tabular}{|c||c|c|c|c|c|} \hline
                      & NGGPs & \multicolumn{4}{c|}{CGP-Flows with modality} \\ \hline
        \textit{tol}s &       & $M=2$ & $M=3$ & $M=4$ & $M=5$ \\ \hline\hline
        1e-2 & 236   & 148   & 99    & 80    & 10  \\ \hline
        1e-3 & 341   & 195   & 162   & 177   & 167 \\ \hline
        1e-4 & 496   & 376   & 218   & 207   & 178 \\ \hline
        1e-5 & 672   & 403   & 278   & 300   & 254 \\ \hline
        1e-6 & 1230  & 540   & 423   & 412   & 355 \\ \hline
    \end{tabular}
\end{table}

\section{Real-world Robot Experiment}
In this section, we applied CGP-Flow to an object-grasping task with an actual robot arm and confirmed the applicability of CGP-Flows to real-world robotic tasks.

\begin{figure}[t]
    \centering
    \includegraphics[width=0.6\linewidth, trim=150 0 0 30, clip]{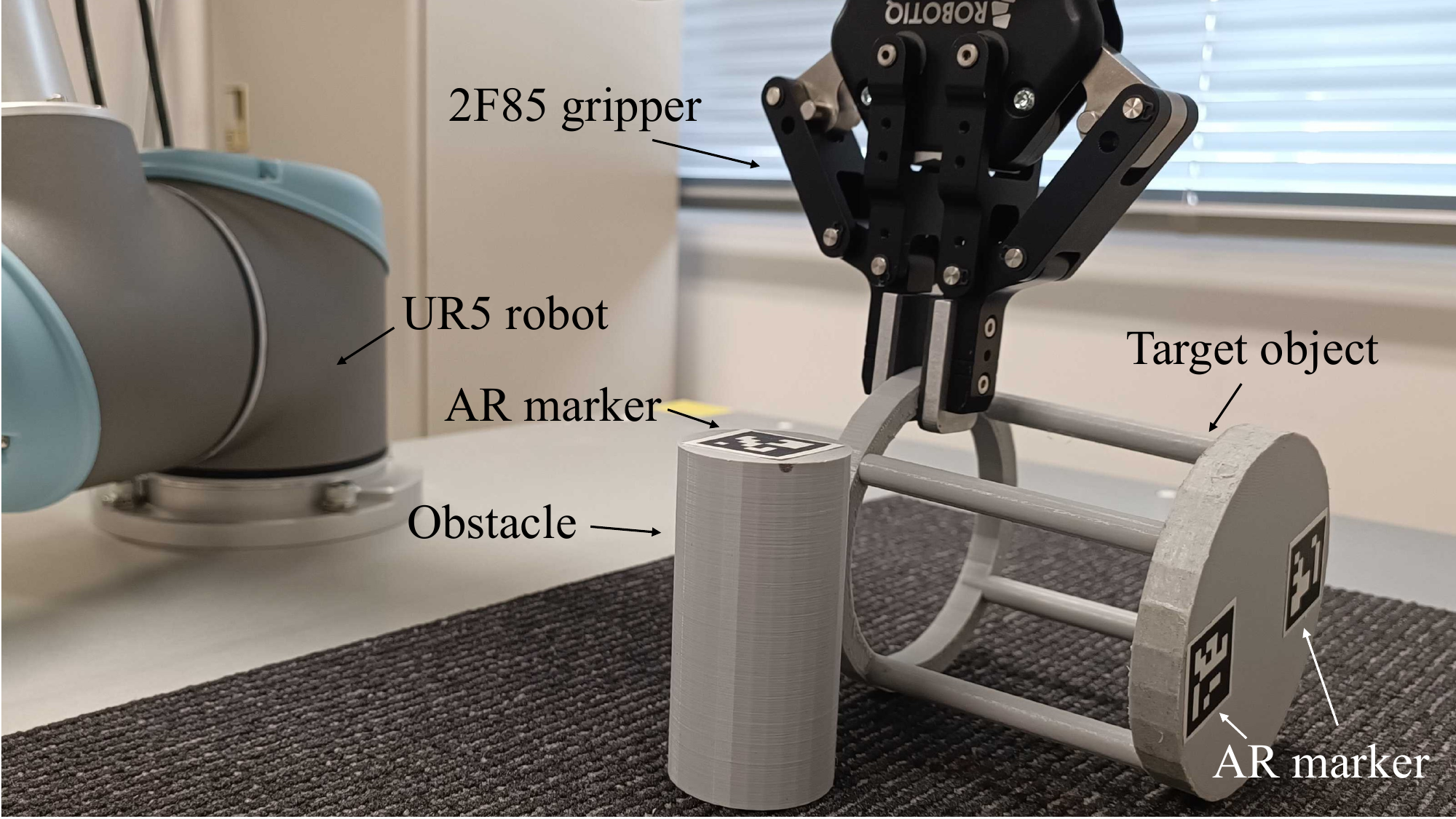}
    \caption{Environment of object-grasping task with actual robotic arm. The task's aim is to grasp a chair-like object while avoiding a cylindrical obstacle.}
    \label{fig:grasping:environment}
\end{figure}

\begin{figure}[t]
    \begin{minipage}[b]{0.32\linewidth}
        \centering
        \includegraphics[width=0.8\linewidth]{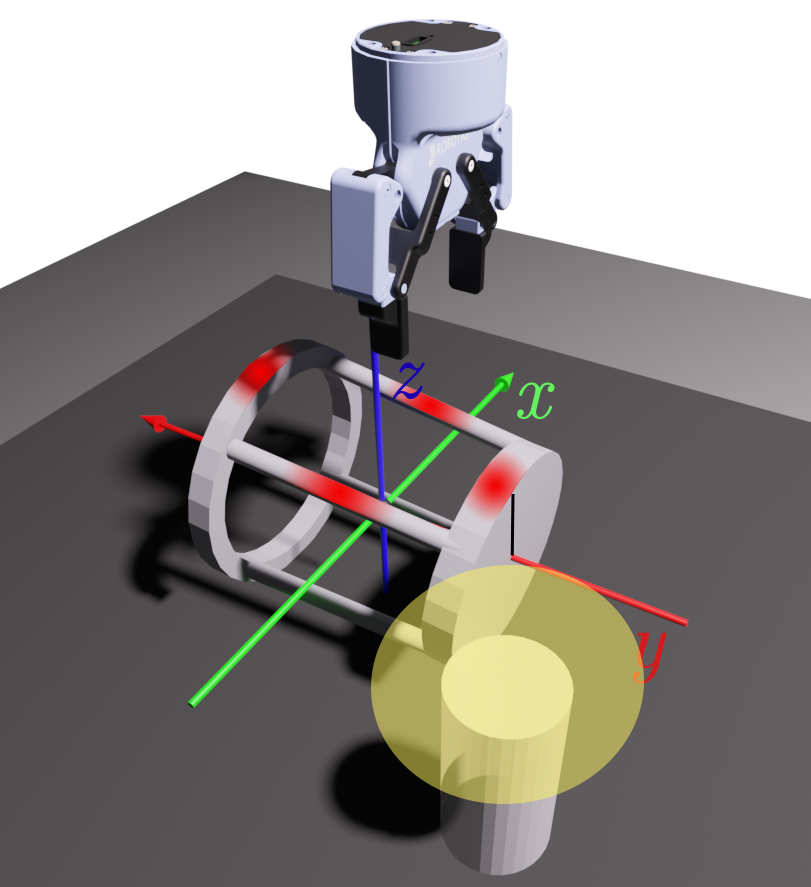}
        \subcaption{}
        \label{fig:grasping:3d_model:a}
    \end{minipage}
    \begin{minipage}[b]{0.32\linewidth}
        \centering
        \includegraphics[width=0.8\linewidth]{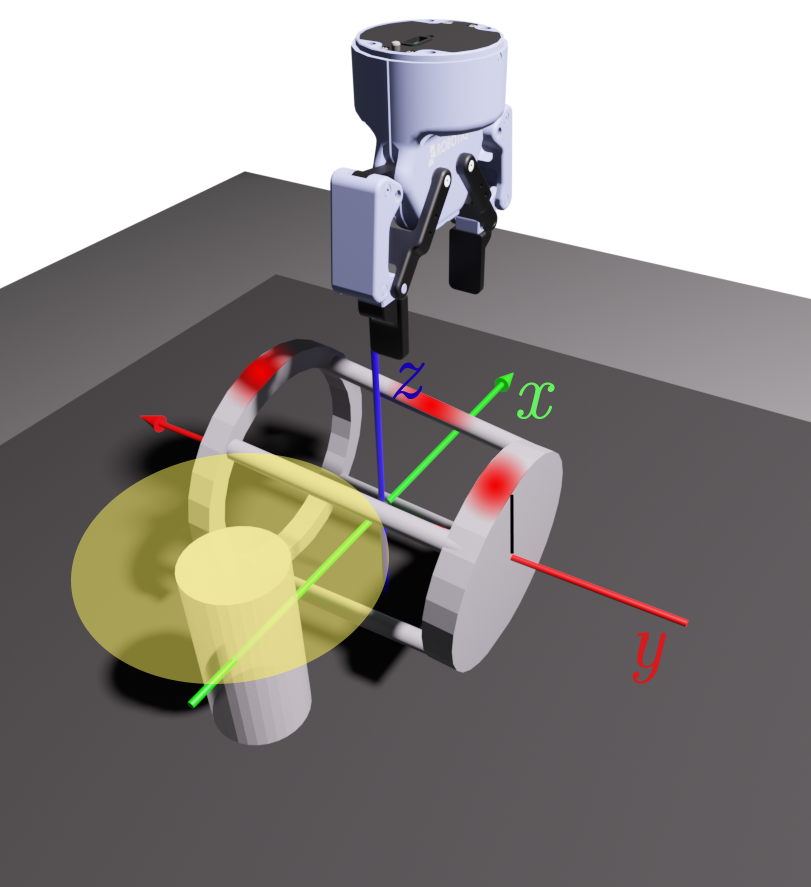}
        \subcaption{}
        \label{fig:grasping:3d_model:b}
    \end{minipage}
    \begin{minipage}[b]{0.32\linewidth}
        \centering
        \includegraphics[width=0.8\linewidth]{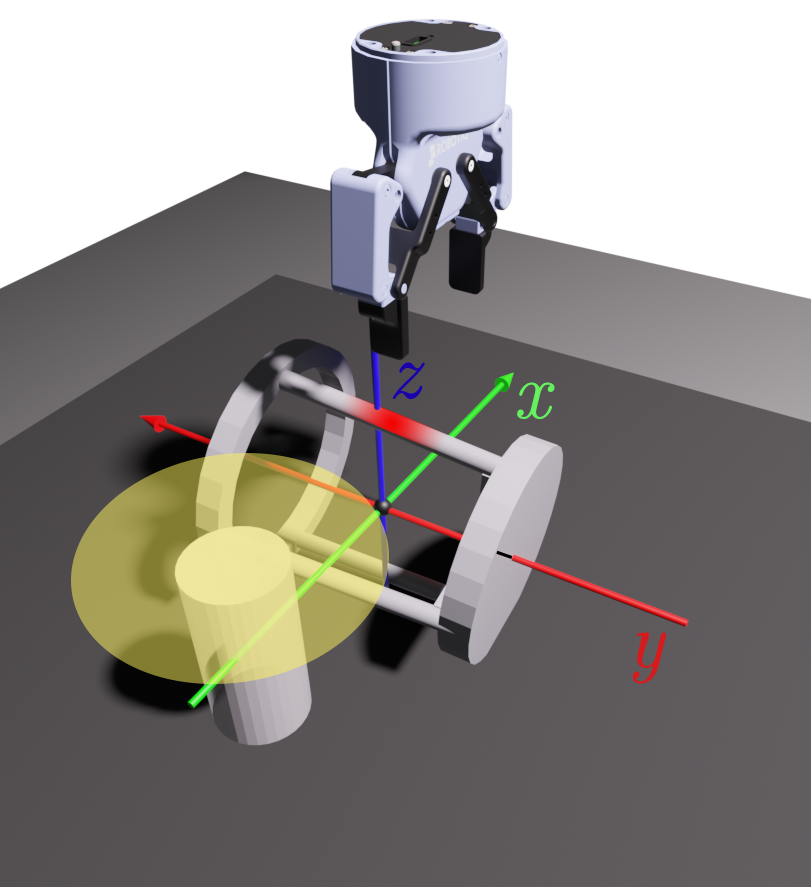}
        \subcaption{}
        \label{fig:grasping:3d_model:c}
    \end{minipage}
    \vspace{-0mm}
    \caption{3D model of grasping experiment: Red areas denote grasping points, and 10 cm yellow circles represent area that might collide with gripper.
    (a) Location of obstacle allows four grasping points, resulting in a four-modal grasping policy.
    (b) One grasping points falls within this circle, leading to its elimination. Consequently, what was originally a four-modal grasping policy becomes trimodal.
    (c) Rotation of target object influences grasping points; in this case, the grasping policy is unimodal.}
    \label{fig:grasping:3d_model}
\end{figure}

\subsection{Settings of Object-grasping Task}
Fig.~\ref{fig:grasping:environment} shows the environment of the object-grasping task.
Its aim is to grasp a chair-like object while avoiding an obstacle.
The chair-like object has a height of 12 cm, a upper radius of 4 cm, a lower radius of 5 cm, and four legs.
The obstacle, a cylinder with a height of 10 cm and a radius of 2 cm placed, was placed near the chair-like object.
The object's graspable positions change due to the chair's rotation and the number of positions where the robot can grasp the object varies depending of the obstacle's presence.
Fig.~\ref{fig:grasping:3d_model} illustrates how the graspable positions are affected by the pose of the chair and the obstacle.

\begin{figure}[t]
    \centering
    \begin{minipage}[b]{0.32\linewidth}
        \centering
        \includegraphics[width=0.8\linewidth]{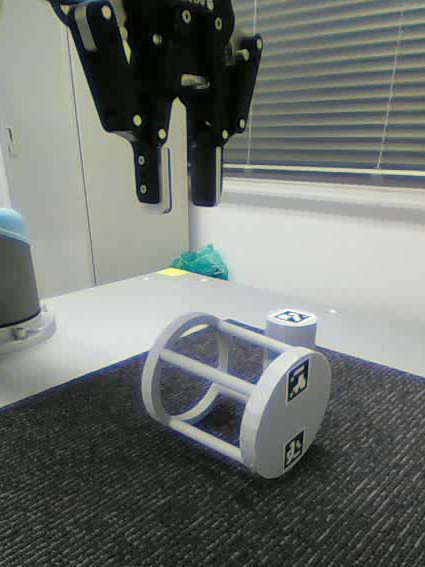}
        \subcaption{Initialize}
        \label{fig:grasping:grasping:a}
    \end{minipage}
    \begin{minipage}[b]{0.32\linewidth}
        \centering
        \includegraphics[width=0.8\linewidth]{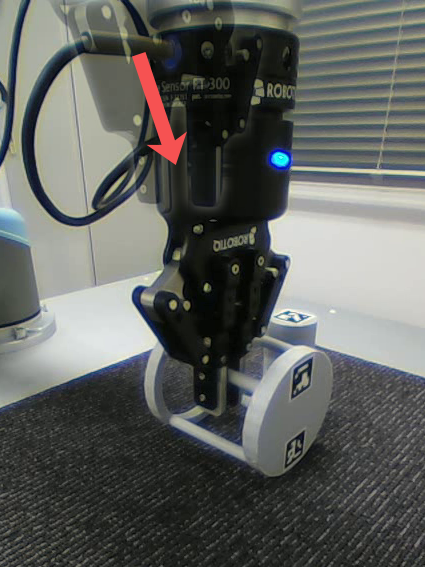}
        \subcaption{Grip}
        \label{fig:grasping:grasping:b}
    \end{minipage}
    \begin{minipage}[b]{0.32\linewidth}
        \centering
        \includegraphics[width=0.8\linewidth]{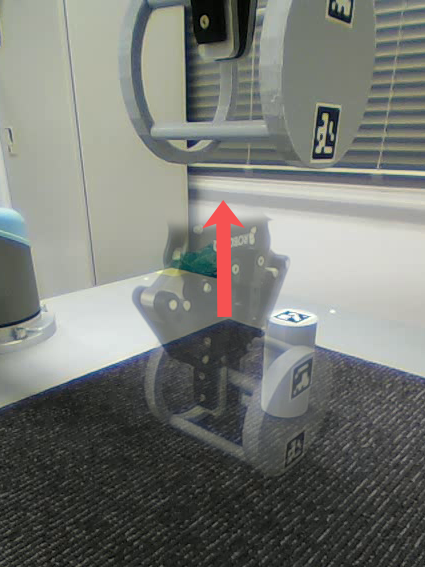}
        \subcaption{Lift}
        \label{fig:grasping:grasping:c}
    \end{minipage}
    \vspace{-2mm}
    \caption{Successful grasping action: Robot movement follows from (a) to (c). (a) Gripper is positioned at its initial location. (b) Gripper moves to a grasping position as determined by the given action. (c) It closes and lifts chair-like object.}
    \label{fig:grasping:grasping}
\end{figure}

We defined a two-dimensional state as $\mathbf{s}_{n}=[r^\mathrm{chair}_n, r^\mathrm{obs}_n]$.
In this setup, $r^{\mathrm{chair}}_n$ refers to the chair's rotation about the y-axis, and $r^\mathrm{obs}_n$ is the obstacle's rotation around the z axis. Both rotations are measured from the center of the chair, and $n$ is the number of experiments.
We assume that the obstacle is placed near the chair, and the distance is fixed between the chair and the obstacle.
We attached AR markers to the chair-like object and the obstacle to measure these rotations.
The AR markers are red using RealSense D435.
The action required for the task is represented by vector $\mathbf{a}_n=[a^x_n, a^y_n, a^z_n, a^{rz}_n]$.
Elements $a^x_n$, $a^y_n$, and $a^z_n$ represent the position for executinge the grasping motion from the object's center along the x, y, and z axes.
Term $a^{rz}_n$ specifies the gripper's rotation around the z axis.
The poses of both the chair-like object and the obstacle are randomly initialized.
The robot's grasping motion is shown in Fig.~\ref{fig:grasping:grasping}.

\subsection{BC Settings}
We developed a complex expert policy for grasping the object and collecting learning data.
Initially, the expert policy computes the potential grasping points based on the chair-like object's rotation $r^\mathrm{chair}$.
Then we consider the obstacle's relative rotation $s_{rz}$ to the chair's center.
The obstacle eliminates any grasping points within a 10 cm radius circle (Fig. \ref{fig:grasping:grasping}). Finally, the policy randomly selects from the remaining viable options.
The expert policy executes the task 2000 times for 2000 random states. The collected learning data consists of state-action pairs ${\mathcal D}=\{{\mathbf{s}}_n, {\mathbf{a}}_n\}_{n=1}^{2000}$ to train the policy models. 

\begin{table}[t]
    \centering
    \renewcommand{\arraystretch}{1.1}
    \caption{Success rate, training time cost and prediction time cost for one step of each grasping task model.}
    \label{tab:grasping:sucess_rate}
    \begin{tabular}{|c||c|c|c|c|c|} \hline
                                            & \multicolumn{5}{c|}{Models} \\ \hline
                                            & OMGP      & OMGP      & NGGP  & CGP-Flow  & CGP-Flow  \\
                                            & with $M=3$ & with $M=4$ &        & with $M=2$ & with $M=3$ \\ \hline\hline
        Success rate                        & 55$\%$     & 67.5$\%$   & 50$\%$ & 85$\%$     & 90$\%$     \\ \hline
        Computation time      & 180        & 223        & 1302   & 1283       & 1072       \\ 
        during training (s)  & & & & & \\ \hline
       Computation time     & 12         & 15         & 74     & 62         & 70         \\ 
       during prediction (ms) & & & & & \\ \hline
    \end{tabular}
\end{table}

\begin{figure}[t]
    \centering
    \begin{minipage}[t]{0.49\linewidth}
        \centering
        \includegraphics[width=0.8\linewidth]{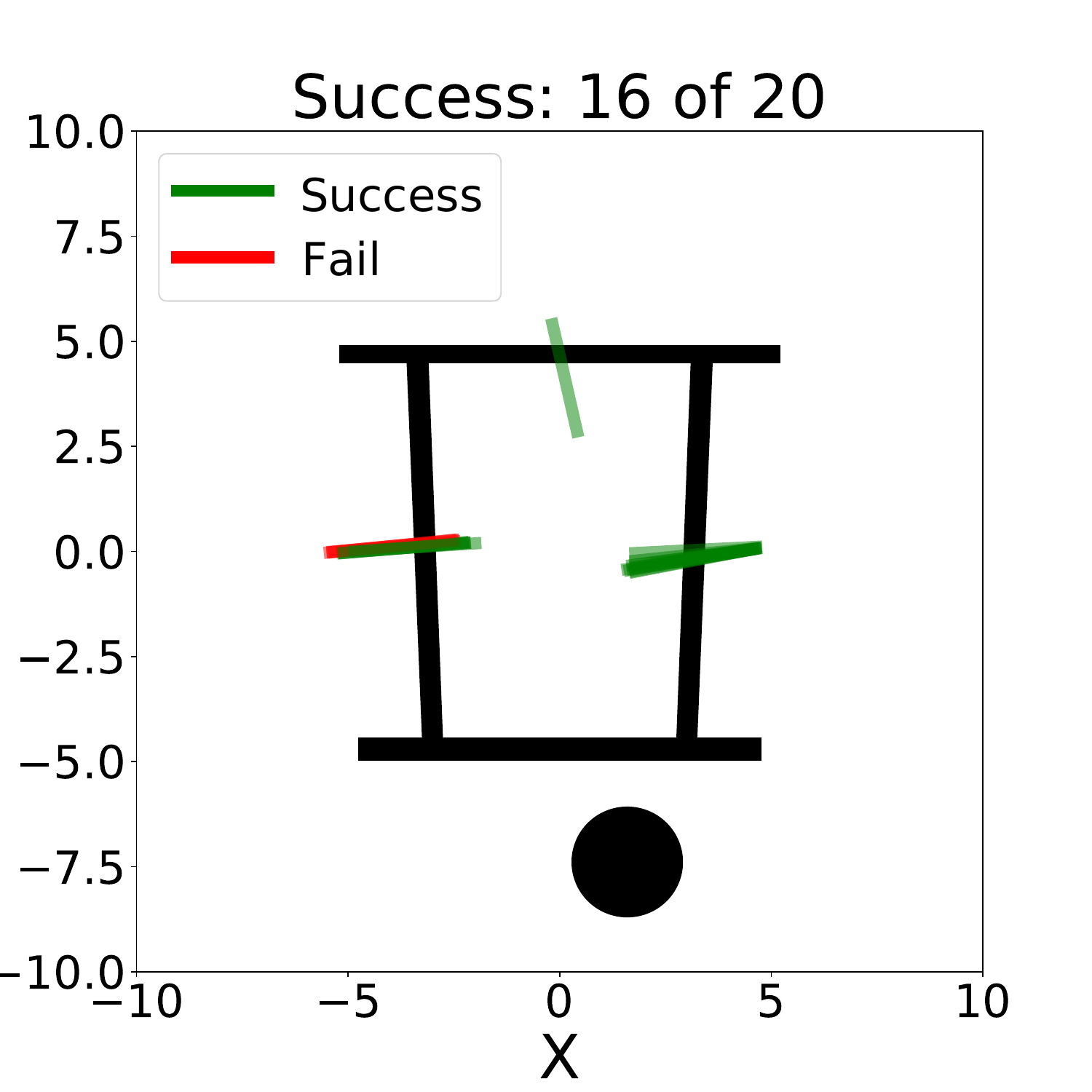}
        \vspace{-2mm}
        \subcaption{CGP-Flow policy with $M=2$}
        \label{fig:grasping:distribution:a}
    \end{minipage}%
    \begin{minipage}[t]{0.49\linewidth}
        \centering
        \includegraphics[width=0.8\linewidth]{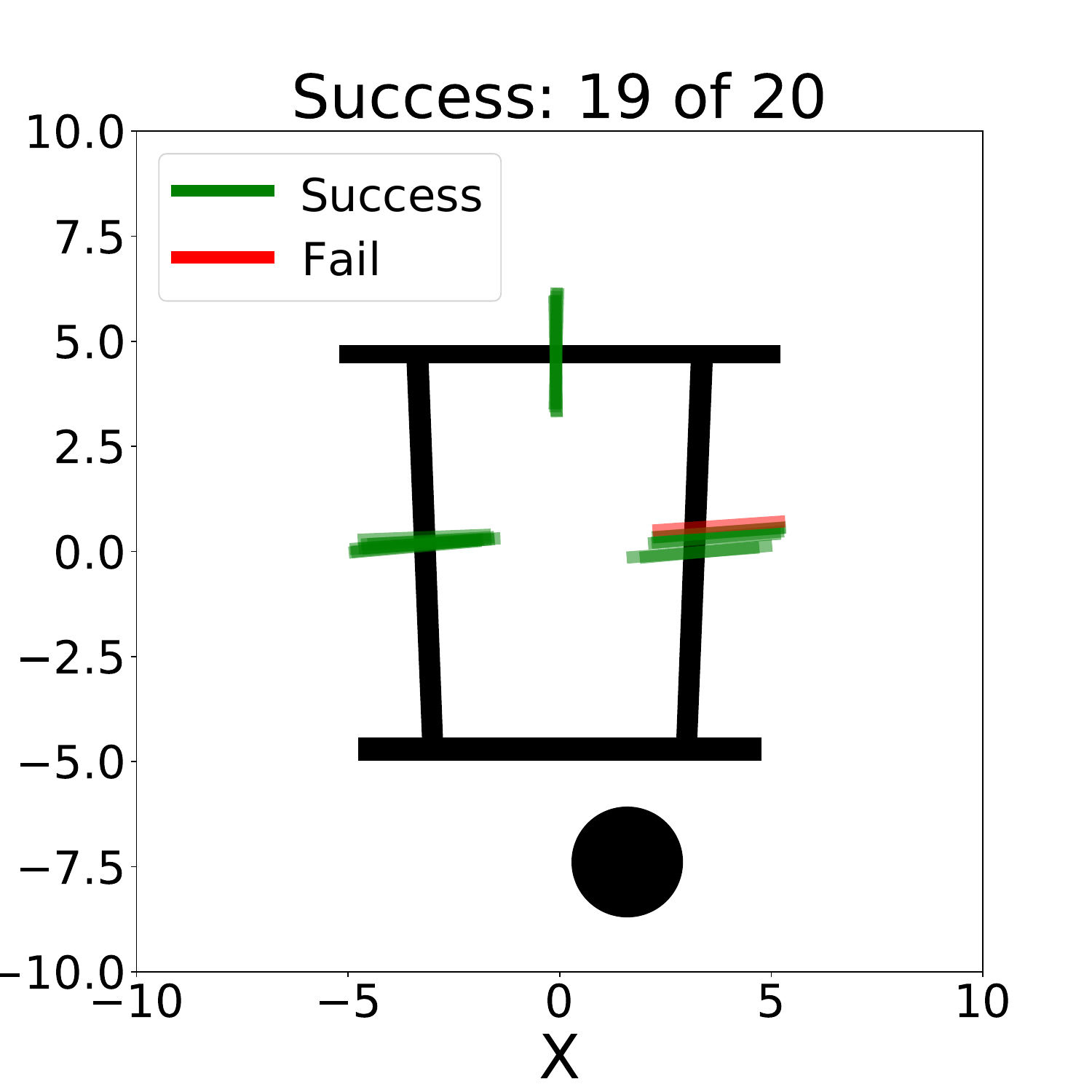}
        \vspace{-2mm}
        \subcaption{CGP-Flow policy with $M=3$}
       \label{fig:grasping:distribution:b}
    \end{minipage}
    \begin{minipage}[t]{0.49\linewidth}
        \centering
        \includegraphics[width=0.8\linewidth]{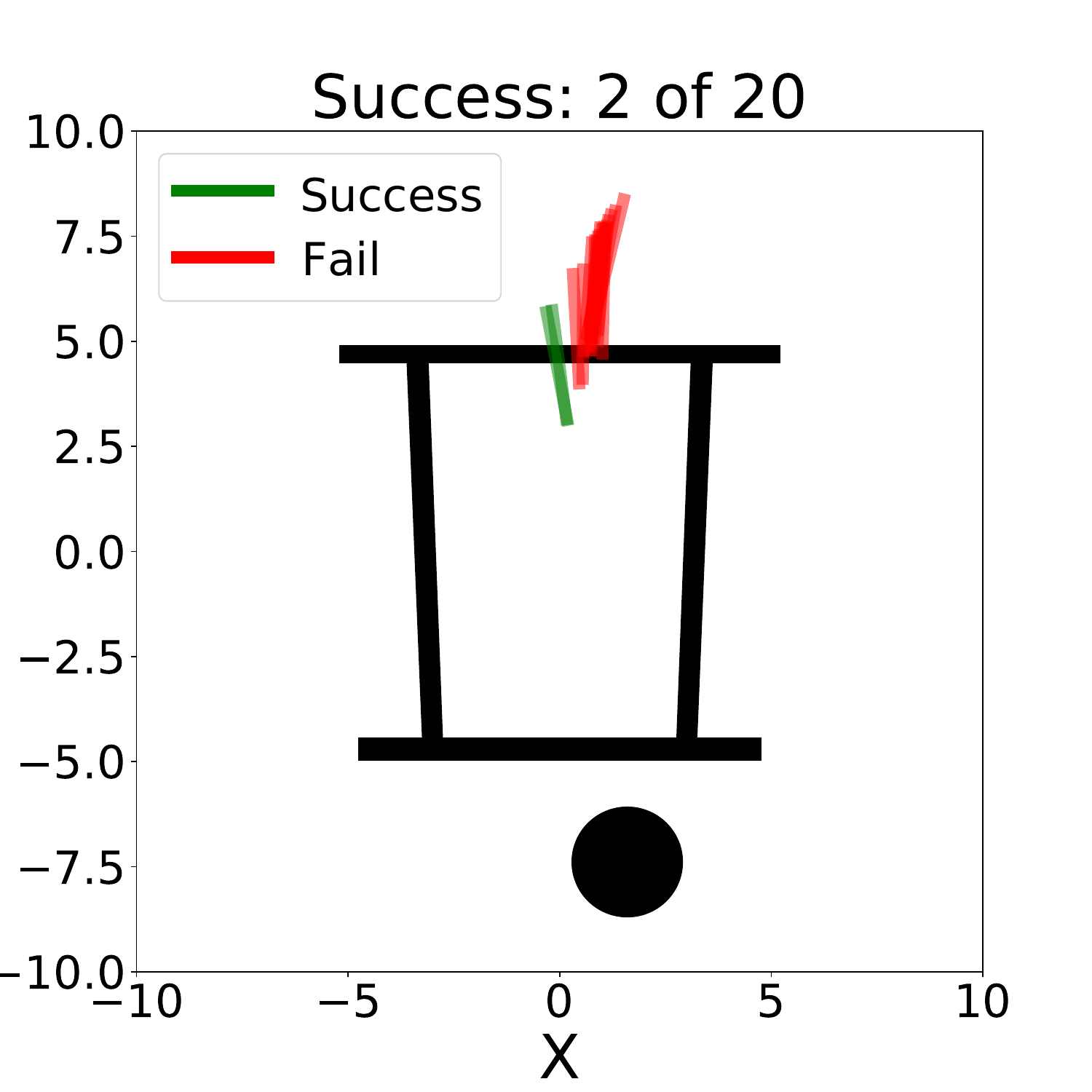}
        \vspace{-2mm}
        \subcaption{NGGP policy}
        \label{fig:grasping:distribution:c}
    \end{minipage}
    \begin{minipage}[t]{0.49\linewidth}
        \centering
        \includegraphics[width=0.8\linewidth]{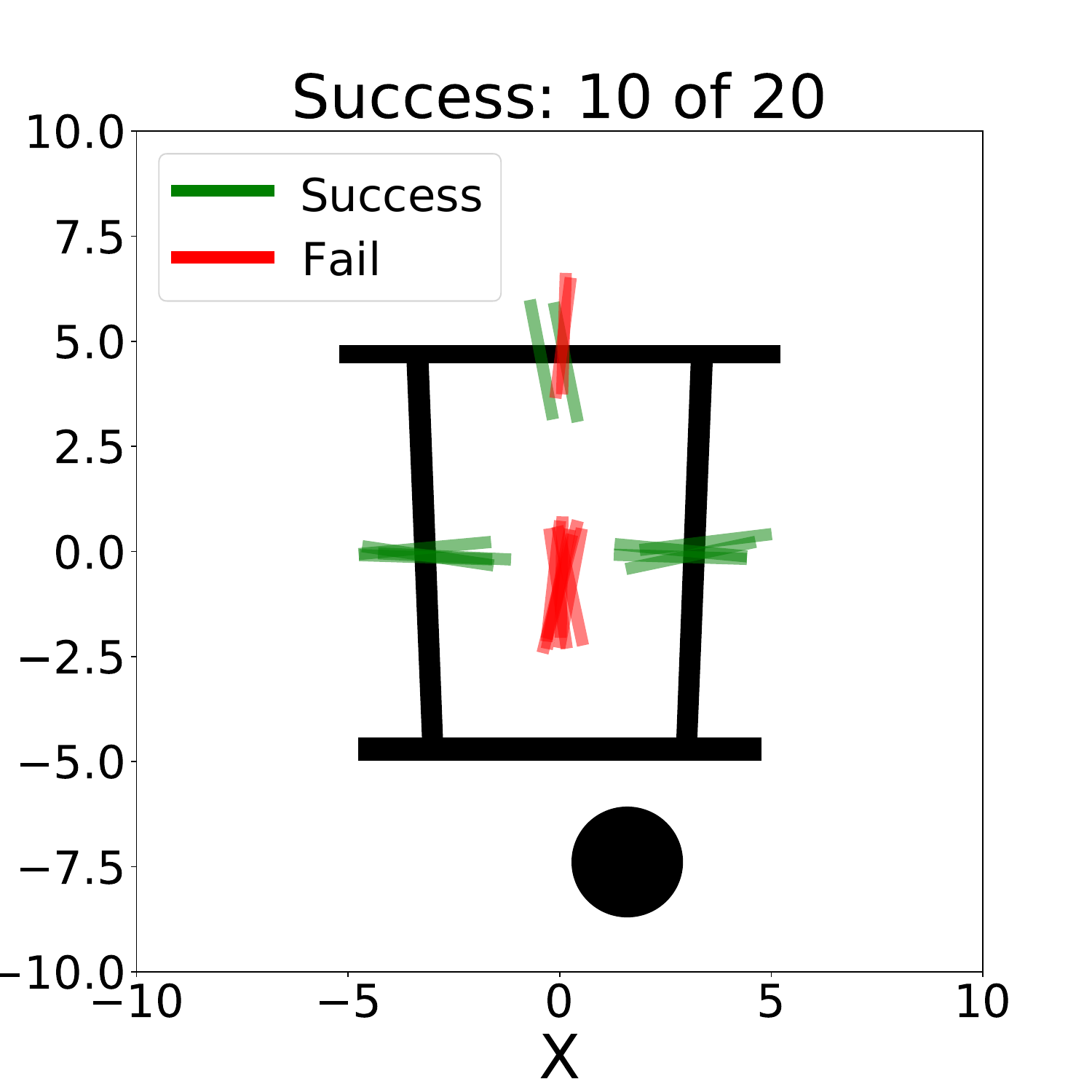}
        \vspace{-2mm}
        \subcaption{OMGP policy with $M=4$}
        \label{fig:grasping:distribution:d}
    \end{minipage}
    \caption{
    Grasping pose plots of four policies: Black lines constitute shape of chair-like target object, and black circle denotes obstacle. Colored short lines denote grasping pose of policies, where green color lines are successful grasps and red color lines are failed grasps.
    (a) CGP-Flow policy with $M=2$ learning trimodal target distributions.
    (b) CGP-Flow policy with $M=3$ learning trimodal target distributions.
    (c) NGGP policy.
    (d) OMGP policy with $M=4$ learning distributions with an obstacle present.
    }
    \label{fig:grasping:distribution}
\end{figure}

\subsubsection{Comparison Methods}
In the real-world grasping task, we employed NGGP, two OMGPs and two CGP-Flows.
For the CGP-Flows, we set the modalities of the base distribution to two and three to evaluate the impact of the multimodality of the base distributions.
The NGGP model has the identical cCNF structure and tolerances as CGP-Flows.
We also utilized the OMGPs with modalities three and four.
CGP-Flows and the NGGP share an identical cCNF, where a single block has two layers, 180 units for each layer.
We chose \textit{dopri5} as the ODE solver of the CGP-Flows and NGGP and set the tolerances to $10^{-5}$.
The initial hyperparameters of OMGPs $\theta^{(m)}$ and $\sigma$ were set to $\theta^{(m)}=0.5$ and $\sigma=0.02$.
The initial hyperparameter of NGGP $\theta$ was set to $\theta=0.2$.
The CGP-Flows share the same initial hyperparameters, where $\theta^{(m)}=0.2$ and $\sigma=0.1$.
Since the state input is only one-dimensional, we used the state as the condition for the CGP-Flows and the NGGP, and thus the condition extractor is defined as $h(s_n)=s_n$.

The optimizer of all the models is still Adam. 
The OMGPs and the CGP-Flows repeat the E-step and the M-step 10 times.
The E-step is terminated when the variational distributions converge, and in the M-step, the parameters are updated for 500 steps.
The learning rate of the OMGPs and CGP-Flows were set at 2e-3 and reduced by $10\%$ following each M-step.
In contrast, the NGGP was still optimized using MLE for 5000 steps.
It also has the same initial learning rate as 2e-3, which is reduced by $10\%$ every 500 steps.
The specifications of the computer used in the experiment are shown in Appendix~\ref{appendix:computational:settings}.

\subsection{Result}
This testing procedure was repeated 40 times, and the success rates and time costs of the policies are presented in Table.~\ref{tab:grasping:sucess_rate}.
The success rates of two CGP-Flow policies are notably higher compared to those of the OMGPs and NGGP policies.

For a visual comparison, the grasping poses of each model are visualized in Fig.~\ref{fig:grasping:distribution}.
Under this state, since the modality number of the task is three, we generated 20 predictive outputs for four policies to compare the performance.
Fig.~\ref{fig:grasping:distribution:a} and Fig.~\ref{fig:grasping:distribution:b} show that the CGP-Flow ($2$) and CGP-Flow ($3$) models successfully learned all three modalities and produced accurate predictive outputs. Notably, all 20 predictive outputs correctly located the correct grasping positions.
The NGGP model grapples with the task's modality complexity, correctly predicting only 2 of the outputs and learned just one of the three required modalities.
The OMGPs ($4$) model learned an incorrect distribution, primarily due to the obstacle disrupting the continuity of the multimodal target distributions.

\section{Discussion}
The implementation of the multimodal base distributions in CGP-Flows is a simple yet highly effective method that significantly enhances both the computational efficiency and the accuracy of the cCNF model.
The computational complexity of multimodal distributions is greater than that of unimodal distributions. However, their expressiveness further reduces the computation time of Neural-ODEs.
In contrast to the above advantages, CGP-flows still have two limitations.

The first limitation lies in the difficulty of updating parameters during the M-step.
CGP-Flows simultaneously optimize the parameters of the Neural-ODE and the sparse OMGP in the M-step of an EM-like algorithm.
However, these parameters exhibit different sensitivities to the model.
Specifically, the sparse OMGP is much more sensitive to parameter updates than the Neural-ODE, and training them similarly can lead to sparse OMGP overfitting.
To address this issue, we adjusted the learning rate of the sparse OMGP parameters to 10$\%$ of that used for the Neural-ODE parameters, effectively aligning the learning progress of the two models.
Additionally, we introduced a learning rate schedule that decreases the rate as iterations of the EM-like algorithm progress.
This approach was found to stabilize the learning process.

The second limitation involves improving learning efficiency when the base distribution exhibits excessive modalities.
CGP-Flows reduce the computational cost of Neural-ODE by using a multimodal base distribution to capture the multimodal nature of the data distribution.
However, when the number of modes in the base distribution exceeds that of the data distribution, the computational cost of Neural-ODE increases.
We consider integrating a Dirichlet Process into the OMGP to mitigate this issue~\cite{ross13a, OH202342}.
This approach is expected to adaptively determine the appropriate number of modalities, thereby enhancing computational efficiency.

In this work, we proposed leveraging enriched base distributions to enhance the performance and flexibility of cCNF-based models, demonstrating how thoughtful design of the non-parametric components can significantly improve policy learning. 
Following this idea, we aim to further explore the potential of employing rich base distributions to extend the capabilities of cCNF-based models.
For instance, integrating an outlier-robust base distribution such as that proposed in~\cite{jylanki11a} into cCNF could improve robustness in noisy environments.
Another promising direction lies in improving the condition extractor $h(\cdot)$, which in our framework is currently implemented as a pre-designed function for simplicity and efficiency.
To enhance performance, $h(\cdot)$ could adopt advanced techniques such as CNN-based feature extraction~\cite{li2020learning, zhang2021bioinspired}, enabling the model to process blurred or obscured visual states more effectively.
Additionally, employing discriminative models~\cite{jin2019multi, CHEN2020150} as $h(\cdot)$ offers greater flexibility in handling unknown and complex scenarios, though this might increase computational demands depending on the task.
These directions highlight the potential for further improving cCNF-based models to tackle increasingly sophisticated robotic policy learning challenges.

\section{Conclusions}
In this paper, we introduced CGP-Flows, an innovative approach designed for learning complex robotic control policies that require modeling smoothness, multimodality, and local discontinuities. Our method combines OMGPs and CNFs, targeting the increased expressivity in cCNF base distributions to reduce the complexity of the transformation and improve modeling precision. This strategic integration not only enhances computational efficiency but also significantly improves the accuracy of capturing multimodal and discontinuous behaviors in robotic tasks. By verifying our hypotheses through simulations and real-world applications, CGP-Flows have demonstrated remarkable improvements in adaptability and performance, confirming the potential of expressive base distributions to refine the accuracy and efficiency of policy modeling in complex robotic environments.

Future work on CGP-Flows includes several key directions for improvement.
These involve developing more efficient learning methods to address the sensitivity of parameters in both non-parametric and parametric models, extending the capabilities of non-parametric models, and expanding feature extractors to enable application to a broader range of robotic tasks.
These improvements are expected to apply CGP-Flows to more sophisticated and diverse robotic tasks.

\section*{Acknowledgment}
This work was supported by JSPS KAKENHI Grant Number JP24K03018.

\appendix
\section*{Appendix}
\section{ODE Solvers and NFE}\label{section:dopri5_nfe}
ODE solvers are a crucial component of both the cCNF and our proposed method. In this section, we explain both their basic methodology and an evaluation method based on NFE.
\subsection{Dormand-Prince 5 ODE Solver}
Given that the core component of the cCNF network is a parametric Neural-ODE, which cannot be solved analytically, the deployment of an ODE solver becomes essential~\cite{chen2018neural}. An ODE solver is a computational algorithm designed to approximate solutions for ODEs without analytical solutions, by iteratively estimating values~\cite{dallas2017comparison}. In this work, our primary ODE solver is Dormand-Prince 5 (\textit{dopri5})~\cite{wanner1996solving}, known for its efficient learning of both simple and complex ODEs.
For any ODE, such as $\frac{\mathrm{d}g}{\mathrm{d}t} = f_{ODE}\left(t, g(t)\right)$
 \textit{dopri5} computes $g_N = g(t)$ in $N$ steps.
At each step, $g_{n}$ is computed~\cite{hu2003numerical}:
\begin{align}\label{dopri5}
    g_{n}&=g_{n-1}+\sum^{I}_{i=1}b_i  k_i, \nonumber \\
    k_i&=h f_{ODE}\left(t_{n-1}+c_i h, g_{n-1}+\sum_{j=1}^I a_{i j} k_j\right) \nonumber \\
    t_{n-1} &= t_{0} + nh. 
\end{align}
In Eq. \ref{dopri5}, $i$ represents the stage number, $h$ is the step length, $t_n$ is the current time and $\{a_{i,j}, b_{i}, c_{i} \}$ are the parameters
$g$ is solved iteratively from the current solution $g_n$.

\subsection{Tolerances and NFE}\label{section:nfe}
As a self-adaptive ODE solver, step length $h$ of \textit{dopri5} is adjusted with respect to absolute tolerances \textit{atol} and relative tolerances \textit{rtol}.
The tolerances are treated as hyperparameters of the \textit{dopri5} ODE solver.
For simplicity, in this work we set \textit{atol} and \textit{rtol} to the same value.
The tolerances affect step length by $sc$~\cite{hairer1993}:
\begin{align}
    sc=\textit{atol}+\max \left(\left|g_{n-1}\right|,\left|g_{n}\right|\right) \cdot \textit {rtol},
\end{align}
where step length $h$ is positive proportional to $sc$~\cite{hairer1993}. 
Consequently, larger tolerances generally lead \textit{dopri5} to opt for larger step length $h$, resulting in faster computations.
Conversely, smaller tolerances cause \textit{dopri5} to use smaller steps, expending more time to solve the ODE while maintaining accuracy.

For a fixed tolerance, \textit{dopri5} requires fewer steps for straightforward ODEs, thus enabling faster computations. In contrast, for more complex ODEs, where adhering to the tolerance level is challenging, additional steps and extended computation time are often necessary to ensure accuracy~\cite{grathwohl2019scalable}.

Since each step in the solver equals one evaluation of the target ODE, the total number of steps is also referred to as the NFE\cite{grathwohl2018ffjord}. 
An NFE, which refers to how many times Neural-ODEs are computed by the ODE solver is a measure of the computational effort required by the ODE solver.
Conversely, simpler ODEs require fewer function evaluations for the given \textit{rtol} and \textit{atol}.
Thus, we utilize NFE as a metric to assess the computational efficiency of Neural-ODE based policy models.

\section{Analytical Solutions of OMGP base distributions}\label{section:solutions}
This section provides analytical solutions of several variables for reference.

\subsection{Expectation-maximization Like Method}
In this expectation step, the analytical solution of variational distribution $q(\bar{\mathbf{f}}^{(m)})$ can be computed:
\begin{align}\label{eq:analytical:qf}
    q(\bar{\mathbf{f}}^{(m)}) & =\mathcal{N}(\bar{\mathbf{f}}^{(m)} \mid \boldsymbol{\mu}^{(m)}, \boldsymbol{\Sigma}^{(m)}), \nonumber \\
    \boldsymbol{\mu}^{(m)} & =\mathbf{K}_{\bar{\mathbf{X}}}^{(m)} \mathbf{Q}_L^{(m)^{-1}} \mathbf{K}_{\bar{\mathbf{X}} \mathbf{X}}^{(m)} \mathbf{B}^{(m)} \mathbf{L}, \nonumber \\
    \boldsymbol{\Sigma}^{(m)} & =\mathbf{K}_{\bar{\mathbf{X}}}^{(m)} \mathbf{Q}_L^{(m)^{-1}} \mathbf{K}_{\bar{\mathbf{X}}}^{(m)}, \nonumber \\
    \mathbf{Q}_L^{(m)} & =\mathbf{K}_{\bar{\mathbf{X}}}^{(m)}+\mathbf{K}_{\bar{\mathbf{X}} \mathbf{X}}^{(m)} \mathbf{B}^{(m)} \mathbf{K}_{\mathbf{X} \bar{\mathbf{X}}}^{(m)}, \nonumber \\
    \mathbf{B}^{(m)} & =\operatorname{diag}\left\{\frac{\hat{\mathbf{\Pi}}_{n m}}{\lambda_n^{(m)}+\sigma^2}\right\} ,
\end{align}
where $\lambda^{(m)}\!=\!\mathbf{K}^{(m)}_{\boldsymbol{X}}-\mathbf{K}^{(m)}_{\mathbf{X} \bar{\mathbf{X}}} \mathbf{K}_{\bar{\mathbf{X}}}^{(m)^{-1}} \mathbf{K}^{(m)}_{\bar{\mathbf{X}} \mathbf{X}}$. 

Next the analytical solution of variational distribution $q\left(\mathbf{Z}\right)$ can be obtained:
\begin{align}\label{eq:analytical:qz}
    q(\mathbf{Z}) & =\prod_{n=1, m=1}^{N, M} [\hat{\boldsymbol{\Pi}}]_{n m}^{[\mathbf{Z}]_{n m}}, \nonumber \\
    [\hat{\boldsymbol{\Pi}}]_{n m} &= [\boldsymbol{\Pi}]_{n m} \exp \left(b_{n m}\right), \nonumber \\
    b_{n m} & =-\frac{1}{2} \log 2 \pi(\lambda_n^{(m)}+\sigma^2) 
     -\frac{\left(\mathbf{l}_n- \mathbf{K}_{\mathbf{X}_n \bar{\mathbf{X}}}^{(m)} \mathbf{K}_{\bar{\mathbf{X}}}^{(m)^{-1}} \boldsymbol{\mu}^{(m)}\right)^2}{2(\lambda_n^{(m)}+\sigma^2)} \nonumber \\
    &~~ -\frac{\mathbf{K}_{\mathbf{X}_n \bar{\mathbf{X}}}^{(m)} \mathbf{Q}_L^{(m)^{-1}} \mathbf{K}_{\bar{\mathbf{X}} \mathbf{X}_n}^{(m)}}{2(\lambda_n^{(m)}+\sigma^2)}. 
\end{align}
 
In the maximization step, the analytical solution of the OMGP's lower bound of log-likelihood can be computed as follows:
\begin{align}\label{eq:analytical:OMGP_ll}
    J_\mathrm{base}({\mathbf L}, q({\mathbf{Z}}), q(\{\bar{\mathbf{f}}^{(m)}\}), \boldsymbol{\theta})
    & = \sum_{m=1}^M-\frac{1}{2} \mathbf{L}^T \left(\mathbf{B}^{(m)^{-1}}+ \mathbf{K}_{\mathbf{X} \bar{\mathbf{X}}}^{(m)} \mathbf{K}_{\bar{\mathbf{X}}}^{(m)^{-1}} \mathbf{K}_{\bar{\mathbf{X}} \mathbf{X}}^{(m)} \right)^{-1}  \mathbf{L} \nonumber \\
    & +\sum_{n=1, m=1}^{N, M} \log [\mathbf{R}^{(m)}]_{nn}-\operatorname{KL}(q(\mathbf{Z}) \mid\mid p(\mathbf{Z})) \nonumber \\
    & -\frac{1}{2} \sum_{n=1, m=1}^{N, M}[\hat{\mathbf{\Pi}}]_{n m} \log 2 \pi(\lambda_n^{(m)}+\sigma^2), \nonumber \\
    \mathbf{R}^{(m)}&=\operatorname{chol}\left(\mathbf{I}+\mathbf{B}^{(m)^{-1 / 2}} \mathbf{K}_{\mathbf{X} \bar{\mathbf{X}}}^{(m)} \mathbf{K}_{\bar{\mathbf{X}}}^{(m)^{-1}} \mathbf{K}_{\bar{\mathbf{X}} \mathbf{X}}^{(m)}  \mathbf{B}^{(m)^{-1 / 2}}\right) .
\end{align}

\subsection{OMGP Predictions}
In the prediction step, the predictive distribution
of the base distributions of OMGP is computed:
\begin{align}\label{eq:analytical:prediction}
    p(\mathbf{l}_*\mid\mathbf{x}_*,\bar{\mathbf{X}}, \mathbf{L})
    & = \prod^{M}_{m=1} [\boldsymbol{\Pi}]_{*m} \int p(\mathbf{l}_* \mid \bar{\mathbf{f}}^{(m)}, \mathbf{x}_* , \bar{\mathbf{X}}) p(\bar{\mathbf{f}}^{(m)}\mid \bar{\mathbf{X}}, \mathbf{L}) \mathrm{d} \bar{\mathbf{f}}^{(m)}  \nonumber \\
    & \approx \prod^{M}_{m=1} [\boldsymbol{\Pi}]_{*m} \int p(\mathbf{l}_* \mid \bar{\mathbf{f}}^{(m)}, \mathbf{x}_* , \bar{\mathbf{X}}) q^*(\bar{\mathbf{f}}^{(m)}) \mathrm{d} \bar{\mathbf{f}}^{(m)}  \nonumber \\
    & =\prod^{M}_{m=1} [\boldsymbol{\Pi}]_{*m} \mathcal{N}(\mathbf{l}_* \mid \boldsymbol{\mu}_*^{(m)}, \boldsymbol{\sigma}_*^{(m)}), \nonumber\\
    \boldsymbol{\mu}_*^{(m)} &=\mathbf{K}_{\mathbf{x}_* \bar{\mathbf{X}}}^{(m)} \mathbf{Q}_L^{(m)^{-1}} \mathbf{K}_{\bar{\mathbf{X}} \mathbf{X}}^{(m)} \mathbf{B}^{(m)} \mathbf{L}, \nonumber\\
    \boldsymbol{\sigma}_*^{(m)} &=\mathbf{K}_{\mathbf{x}_*}^{(m)}+\sigma^2 -\mathbf{K}^{(m)}_{\mathbf{x}_* \bar{\mathbf{X}}}\left(\mathbf{K}_{\bar{\mathbf{X}}}^{(m)^{-1}}-\mathbf{Q}_L^{(m)^{-1}}\right) \mathbf{K}_{\bar{\mathbf{X}} \mathbf{x}_*}^{(m)},
\end{align}
where mixing factors $\boldsymbol{\Pi}$ are the prior probabilities of each GP, and typically constant for all the inputs~\cite{lazaro2012overlapping}.

\section{Experimental Environment}\label{appendix:computational:settings}
The experiments in this paper utilized two distinct hardware setups.
The detailed specifications of the computational environments are as follows:

\begin{table}[h]
    \centering
    \caption{Computing environment for the ball-shooting simulations and real-world robot experiments}
    \label{tab:time:cost}
    \begin{tabular}{|c||l|} \hline
        Components & Spesification \\ \hline \hline
        CPU        & 13th Gen Intel(R) Core(TM) i9-13900KF \\
        GPU        & GeForce RTX 4080 with 16GB of VRAM  \\
        Memory     & 32GB $\times$ 4 (128GB RAM)  \\
        OS         & Ubuntu 22.04.5 LTS  \\
        Python     & Python 3.8.17  \\
        Pytorch    & 2.0.1+cu117  \\
        CUDA       & 11.7 \\
        \hline
    \end{tabular}
\end{table}

\begin{table}[h]
    \centering
    \caption{Computing environment for the object-swiping simulation}
    \label{tab:time:cost}
    \begin{tabular}{|c||l|} \hline
        Components & Spesification \\ \hline \hline
        CPU        & 11th Gen Intel(R) Core(TM) i9-11900K  \\
        GPU        & GeForce RTX 3080 with 10GB of VRAM  \\
        Memory     & 16GB $\times$ 2 (32GB RAM) \\
        OS         & Ubuntu 20.04.5 LTS  \\
        Python     & Python 3.8.17  \\
        Pytorch    & 1.12.1+cu116  \\
        CUDA       & 11.6  \\
        \hline
    \end{tabular}
\end{table}

These setups ensure computational efficiency and demonstrate that the proposed methods are feasible for both simulation and real-world applications.

\section{Baseline Analysis: NGGP Performance in Our Simulation}\label{appendix:baseline:nggp}
To ensure fairness in evaluation, we conducted simulations in the ball-shooting environment using both our optimized NGGP configuration and the original settings proposed in the NGGP paper~\cite{sendera2021non}.
The original configuration was implemented to replicate the experimental setup as closely as possible, providing a direct comparison with the baseline.
For original NGGP settings, the model is trained for 10,000 iterations using the Adam optimizer with learning rate = 0.001, $\beta_1 = 0.9$, $\beta_2 = 0.999$.
The feature extractor is a 2-layer MLP with ReLU activations and a hidden dimension of 40, following the settings in~\cite{sendera2021non}.
The NGGP utilizes two stacked cCNF blocks, each comprising two hidden layers (64 units each) with Tanh activations. These layers are conditioned on the 40-dimensional output of the feature extractor for the cCNF variant. 

\begin{figure}[t]
    \centering
    \begin{minipage}[t]{0.9\linewidth}
        \centering
        \includegraphics[width=0.46\linewidth]{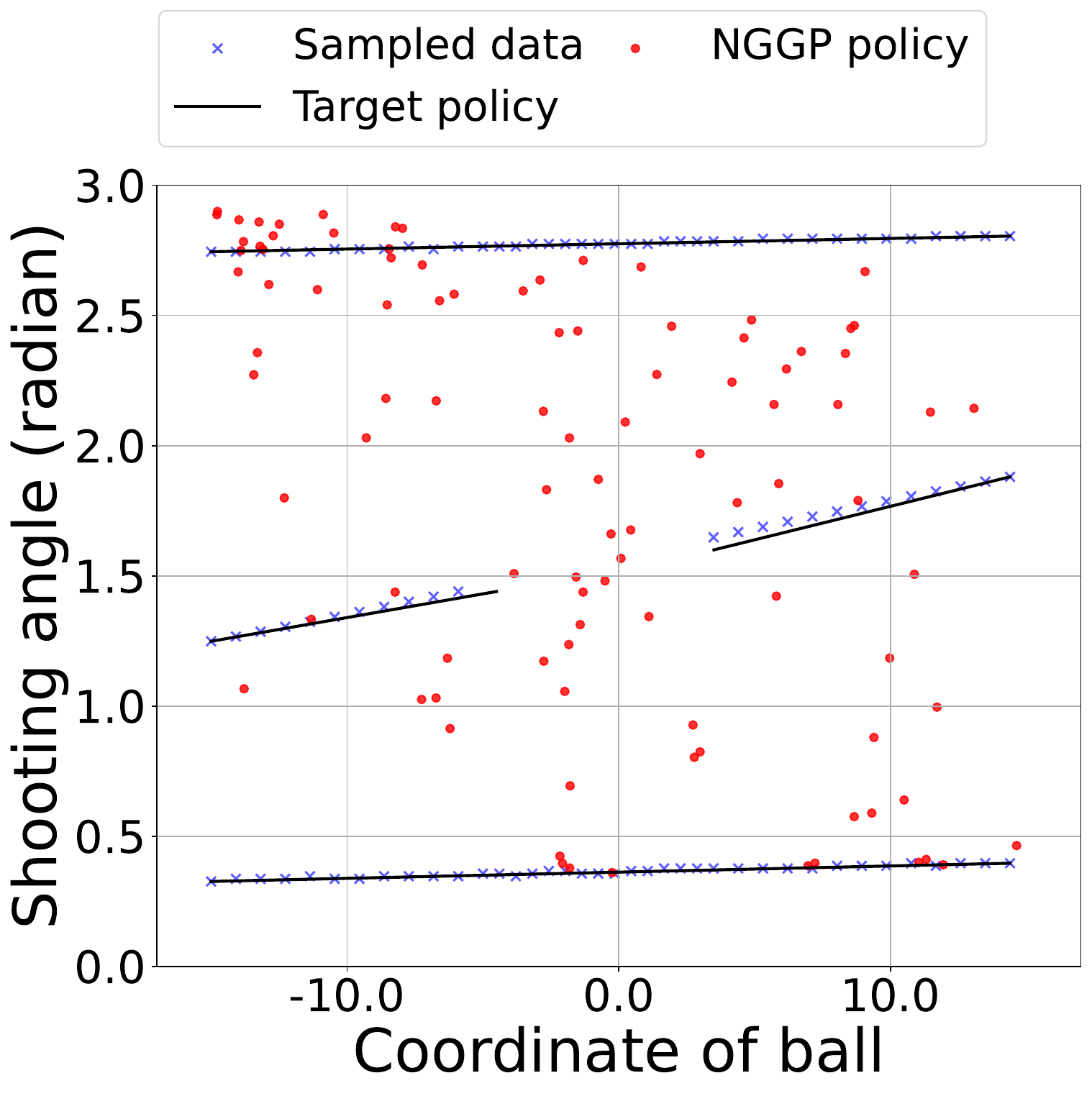}
        \includegraphics[width=0.52\linewidth]{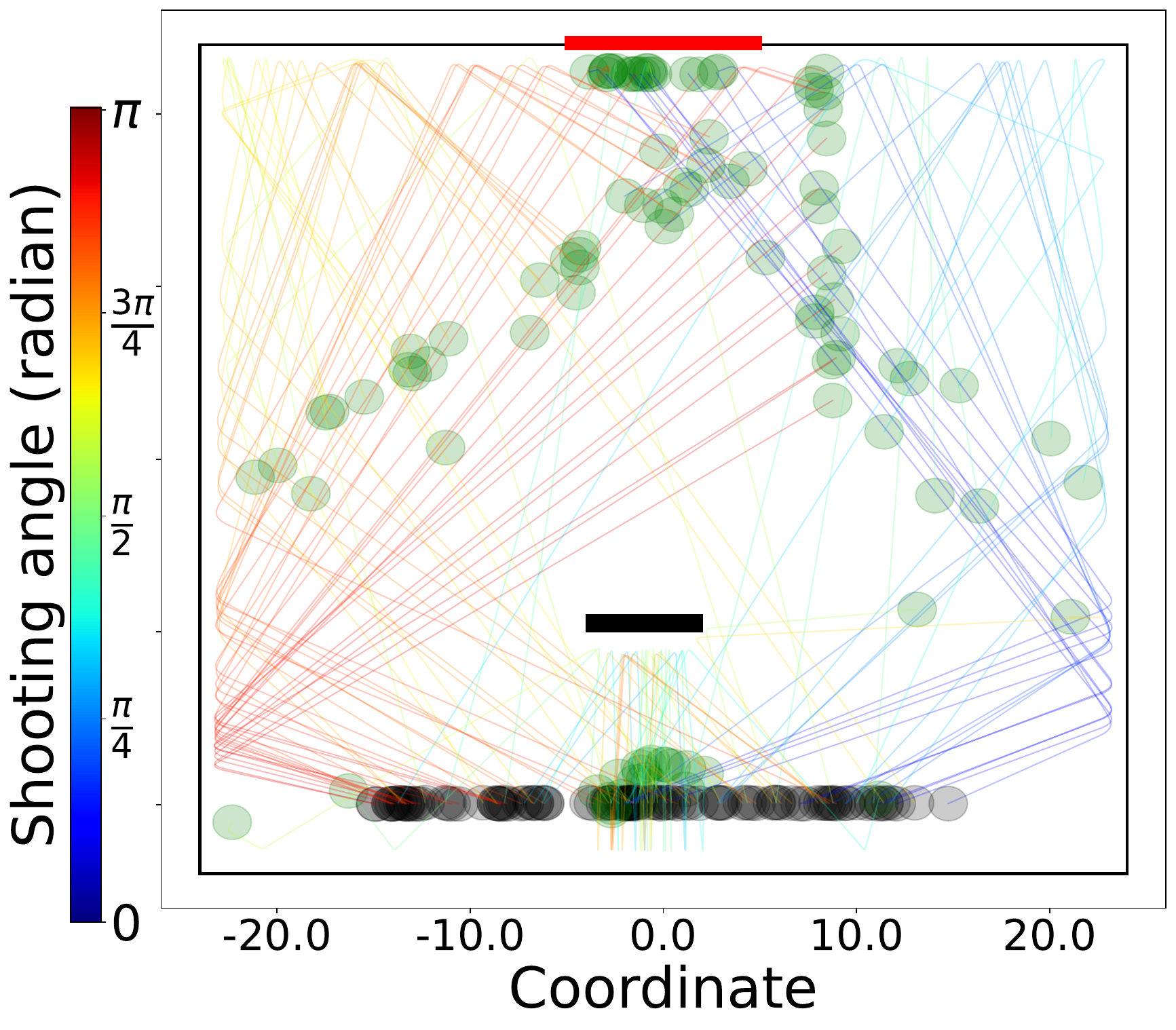}
    \end{minipage}
    \caption{Visualization of the NGGP policy applied to the ball-shooting task. (Left) Action distribution learned by NGGP. (Right) Trajectory outcomes under the default NGGP configuration proposed in~\cite{sendera2021non}.}
    \label{fig:shooting:nggp_compare}
\end{figure}

Figure~\ref{fig:shooting:nggp_compare} presents the results of this experiment.
Under the original NGGP settings, the success rate in our ball-shooting task was 18.1\%, with a training time of 1242.33 seconds and a prediction time of 55.0 milliseconds per query.
This performance highlights the challenges posed by our task, which involves modeling multimodal and discontinuous policies.
The original NGGP configuration was not well-suited to this complexity, resulting in a relatively low success rate. Consequently, in the simulations discussed in the main text, we used optimized configurations for both the CGP-Flow model and the cCNF component of NGGP, ensuring better adaptation to the unique demands of our task.

\bibliographystyle{sn-mathphys-num} 
\bibliography{ref}


\begin{thebibliography}{37}
\ifx \bisbn   \undefined \def \bisbn  #1{ISBN #1}\fi
\ifx \binits  \undefined \def \binits#1{#1}\fi
\ifx \bauthor  \undefined \def \bauthor#1{#1}\fi
\ifx \batitle  \undefined \def \batitle#1{#1}\fi
\ifx \bjtitle  \undefined \def \bjtitle#1{#1}\fi
\ifx \bvolume  \undefined \def \bvolume#1{\textbf{#1}}\fi
\ifx \byear  \undefined \def \byear#1{#1}\fi
\ifx \bissue  \undefined \def \bissue#1{#1}\fi
\ifx \bfpage  \undefined \def \bfpage#1{#1}\fi
\ifx \blpage  \undefined \def \blpage #1{#1}\fi
\ifx \burl  \undefined \def \burl#1{\textsf{#1}}\fi
\ifx \doiurl  \undefined \def \doiurl#1{\url{https://doi.org/#1}}\fi
\ifx \betal  \undefined \def \betal{\textit{et al.}}\fi
\ifx \binstitute  \undefined \def \binstitute#1{#1}\fi
\ifx \binstitutionaled  \undefined \def \binstitutionaled#1{#1}\fi
\ifx \bctitle  \undefined \def \bctitle#1{#1}\fi
\ifx \beditor  \undefined \def \beditor#1{#1}\fi
\ifx \bpublisher  \undefined \def \bpublisher#1{#1}\fi
\ifx \bbtitle  \undefined \def \bbtitle#1{#1}\fi
\ifx \bedition  \undefined \def \bedition#1{#1}\fi
\ifx \bseriesno  \undefined \def \bseriesno#1{#1}\fi
\ifx \blocation  \undefined \def \blocation#1{#1}\fi
\ifx \bsertitle  \undefined \def \bsertitle#1{#1}\fi
\ifx \bsnm \undefined \def \bsnm#1{#1}\fi
\ifx \bsuffix \undefined \def \bsuffix#1{#1}\fi
\ifx \bparticle \undefined \def \bparticle#1{#1}\fi
\ifx \barticle \undefined \def \barticle#1{#1}\fi
\bibcommenthead
\ifx \bconfdate \undefined \def \bconfdate #1{#1}\fi
\ifx \botherref \undefined \def \botherref #1{#1}\fi
\ifx \url \undefined \def \url#1{\textsf{#1}}\fi
\ifx \bchapter \undefined \def \bchapter#1{#1}\fi
\ifx \bbook \undefined \def \bbook#1{#1}\fi
\ifx \bcomment \undefined \def \bcomment#1{#1}\fi
\ifx \oauthor \undefined \def \oauthor#1{#1}\fi
\ifx \citeauthoryear \undefined \def \citeauthoryear#1{#1}\fi
\ifx \endbibitem  \undefined \def \endbibitem {}\fi
\ifx \bconflocation  \undefined \def \bconflocation#1{#1}\fi
\ifx \arxivurl  \undefined \def \arxivurl#1{\textsf{#1}}\fi
\csname PreBibitemsHook\endcsname

\bibitem[\protect\citeauthoryear{Deisenroth
  et~al.}{2013}]{deisenroth2013gaussian}
\begin{barticle}
\bauthor{\bsnm{Deisenroth}, \binits{M.P.}},
\bauthor{\bsnm{Fox}, \binits{D.}},
\bauthor{\bsnm{Rasmussen}, \binits{C.E.}}:
\batitle{Gaussian processes for data-efficient learning in robotics and
  control}.
\bjtitle{IEEE Transactions on Pattern Analysis and Machine Intelligence}
\bvolume{37}(\bissue{2}),
\bfpage{408}--\blpage{423}
(\byear{2013})
\end{barticle}
\endbibitem

\bibitem[\protect\citeauthoryear{Schreiter et~al.}{2015}]{schreiter2015sparse}
\begin{bchapter}
\bauthor{\bsnm{Schreiter}, \binits{J.}},
\bauthor{\bsnm{Englert}, \binits{P.}},
\bauthor{\bsnm{Nguyen-Tuong}, \binits{D.}},
\bauthor{\bsnm{Toussaint}, \binits{M.}}:
\bctitle{Sparse gaussian process regression for compliant, real-time robot
  control}.
In: \bbtitle{IEEE International Conference on Robotics and Automation (ICRA)},
pp. \bfpage{2586}--\blpage{2591}
(\byear{2015})
\end{bchapter}
\endbibitem

\bibitem[\protect\citeauthoryear{Sasaki et~al.}{2022}]{sasaki2022gaussian}
\begin{bchapter}
\bauthor{\bsnm{Sasaki}, \binits{H.}},
\bauthor{\bsnm{Hirabayashi}, \binits{T.}},
\bauthor{\bsnm{Kawabata}, \binits{K.}},
\bauthor{\bsnm{Matsubara}, \binits{T.}}:
\bctitle{Gaussian process self-triggered policy search in weakly observable
  environments}.
In: \bbtitle{International Conference on Robotics and Automation (ICRA)},
pp. \bfpage{5946}--\blpage{5952}
(\byear{2022})
\end{bchapter}
\endbibitem

\bibitem[\protect\citeauthoryear{Tsurumine et~al.}{2019}]{tsurumine2019deep}
\begin{barticle}
\bauthor{\bsnm{Tsurumine}, \binits{Y.}},
\bauthor{\bsnm{Cui}, \binits{Y.}},
\bauthor{\bsnm{Uchibe}, \binits{E.}},
\bauthor{\bsnm{Matsubara}, \binits{T.}}:
\batitle{Deep reinforcement learning with smooth policy update: Application to
  robotic cloth manipulation}.
\bjtitle{Robotics and Autonomous Systems}
\bvolume{112},
\bfpage{72}--\blpage{83}
(\byear{2019})
\end{barticle}
\endbibitem

\bibitem[\protect\citeauthoryear{Shen et~al.}{2020}]{shen2020deep}
\begin{bchapter}
\bauthor{\bsnm{Shen}, \binits{Q.}},
\bauthor{\bsnm{Li}, \binits{Y.}},
\bauthor{\bsnm{Jiang}, \binits{H.}},
\bauthor{\bsnm{Wang}, \binits{Z.}},
\bauthor{\bsnm{Zhao}, \binits{T.}}:
\bctitle{Deep reinforcement learning with robust and smooth policy}.
In: \bbtitle{International Conference on Machine Learning (ICML)},
pp. \bfpage{8707}--\blpage{8718}
(\byear{2020})
\end{bchapter}
\endbibitem

\bibitem[\protect\citeauthoryear{Schulz et~al.}{2018}]{schulz2018tutorial}
\begin{barticle}
\bauthor{\bsnm{Schulz}, \binits{E.}},
\bauthor{\bsnm{Speekenbrink}, \binits{M.}},
\bauthor{\bsnm{Krause}, \binits{A.}}:
\batitle{A tutorial on gaussian process regression: Modelling, exploring, and
  exploiting functions}.
\bjtitle{Journal of Mathematical Psychology}
\bvolume{85},
\bfpage{1}--\blpage{16}
(\byear{2018})
\end{barticle}
\endbibitem

\bibitem[\protect\citeauthoryear{L{\'a}zaro-Gredilla
  et~al.}{2012}]{lazaro2012overlapping}
\begin{barticle}
\bauthor{\bsnm{L{\'a}zaro-Gredilla}, \binits{M.}},
\bauthor{\bsnm{Van~Vaerenbergh}, \binits{S.}},
\bauthor{\bsnm{Lawrence}, \binits{N.D.}}:
\batitle{Overlapping mixtures of gaussian processes for the data association
  problem}.
\bjtitle{Pattern Recognition}
\bvolume{45}(\bissue{4}),
\bfpage{1386}--\blpage{1395}
(\byear{2012})
\end{barticle}
\endbibitem

\bibitem[\protect\citeauthoryear{Sendera et~al.}{2021}]{sendera2021non}
\begin{bchapter}
\bauthor{\bsnm{Sendera}, \binits{M.}},
\bauthor{\bsnm{Tabor}, \binits{J.}},
\bauthor{\bsnm{Nowak}, \binits{A.}},
\bauthor{\bsnm{Bedychaj}, \binits{A.}},
\bauthor{\bsnm{Patacchiola}, \binits{M.}},
\bauthor{\bsnm{Trzcinski}, \binits{T.}},
\bauthor{\bsnm{Spurek}, \binits{P.}},
\bauthor{\bsnm{Zieba}, \binits{M.}}:
\bctitle{Non-gaussian gaussian processes for few-shot regression}.
In: \bbtitle{Neural Information Processing Systems (NeurIPS)},
vol. \bseriesno{34},
pp. \bfpage{10285}--\blpage{10298}
(\byear{2021})
\end{bchapter}
\endbibitem

\bibitem[\protect\citeauthoryear{Papamakarios
  et~al.}{2021}]{papamakarios2021normalizing}
\begin{barticle}
\bauthor{\bsnm{Papamakarios}, \binits{G.}},
\bauthor{\bsnm{Nalisnick}, \binits{E.}},
\bauthor{\bsnm{Rezende}, \binits{D.J.}},
\bauthor{\bsnm{Mohamed}, \binits{S.}},
\bauthor{\bsnm{Lakshminarayanan}, \binits{B.}}:
\batitle{Normalizing flows for probabilistic modeling and inference}.
\bjtitle{Journal of Machine Learning Research}
\bvolume{22}(\bissue{57}),
\bfpage{1}--\blpage{64}
(\byear{2021})
\end{barticle}
\endbibitem

\bibitem[\protect\citeauthoryear{Chen et~al.}{2018}]{chen2018neural}
\begin{bchapter}
\bauthor{\bsnm{Chen}, \binits{R.T.}},
\bauthor{\bsnm{Rubanova}, \binits{Y.}},
\bauthor{\bsnm{Bettencourt}, \binits{J.}},
\bauthor{\bsnm{Duvenaud}, \binits{D.K.}}:
\bctitle{Neural ordinary differential equations}.
In: \bbtitle{Neural Information Processing Systems (NeurIPS)},
vol. \bseriesno{31},
pp. \bfpage{6572}--\blpage{6583}
(\byear{2018})
\end{bchapter}
\endbibitem

\bibitem[\protect\citeauthoryear{Grathwohl et~al.}{2019}]{grathwohl2018ffjord}
\begin{bchapter}
\bauthor{\bsnm{Grathwohl}, \binits{W.}},
\bauthor{\bsnm{Chen}, \binits{R.T.}},
\bauthor{\bsnm{Bettencourt}, \binits{J.}},
\bauthor{\bsnm{Sutskever}, \binits{I.}},
\bauthor{\bsnm{Duvenaud}, \binits{D.}}:
\bctitle{{FFJORD}: Free-form continuous dynamics for scalable reversible
  generative models}.
In: \bbtitle{International Conference on Learning Representations (ICLR)}
(\byear{2019})
\end{bchapter}
\endbibitem

\bibitem[\protect\citeauthoryear{Osa et~al.}{2018}]{osa2018}
\begin{barticle}
\bauthor{\bsnm{Osa}, \binits{T.}},
\bauthor{\bsnm{Pajarinen}, \binits{J.}},
\bauthor{\bsnm{Neumann}, \binits{G.}},
\bauthor{\bsnm{Bagnell}, \binits{J.A.}},
\bauthor{\bsnm{Abbeel}, \binits{P.}},
\bauthor{\bsnm{Peters}, \binits{J.}}:
\batitle{An algorithmic perspective on imitation learning}.
\bjtitle{Foundations and Trends in Robotics}
\bvolume{7}(\bissue{1-2}),
\bfpage{1}--\blpage{179}
(\byear{2018})
\end{barticle}
\endbibitem

\bibitem[\protect\citeauthoryear{Chang et~al.}{2021}]{chang2021ilflow}
\begin{bchapter}
\bauthor{\bsnm{Chang}, \binits{W.}},
\bauthor{\bsnm{Higuera}, \binits{J.C.G.}},
\bauthor{\bsnm{Fujimoto}, \binits{S.}},
\bauthor{\bsnm{Meger}, \binits{D.}},
\bauthor{\bsnm{Dudek}, \binits{G.}}:
\bctitle{{IL-flOw: Imitation Learning from Observation using Normalizing
  Flows}}.
In: \bbtitle{Proceedings of the 4th Robot Learning Workshop at NeurIPS 2021}
(\byear{2021})
\end{bchapter}
\endbibitem

\bibitem[\protect\citeauthoryear{Chisari et~al.}{2024}]{chisari2024learning}
\begin{botherref}
\oauthor{\bsnm{Chisari}, \binits{E.}},
\oauthor{\bsnm{Heppert}, \binits{N.}},
\oauthor{\bsnm{Argus}, \binits{M.}},
\oauthor{\bsnm{Welschehold}, \binits{T.}},
\oauthor{\bsnm{Brox}, \binits{T.}},
\oauthor{\bsnm{Valada}, \binits{A.}}:
Learning robotic manipulation policies from point clouds with conditional flow
  matching.
Conference on Robot Learning (CoRL)
(2024)
\end{botherref}
\endbibitem

\bibitem[\protect\citeauthoryear{Fadel et~al.}{2023}]{fadel2023contextual}
\begin{barticle}
\bauthor{\bsnm{Fadel}, \binits{S.G.}},
\bauthor{\bsnm{Mair}, \binits{S.}},
\bauthor{\bsnm{Silva~Torres}, \binits{R.}},
\bauthor{\bsnm{Brefeld}, \binits{U.}}:
\batitle{Contextual movement models based on normalizing flows}.
\bjtitle{AStA Advances in Statistical Analysis}
\bvolume{107}(\bissue{1}),
\bfpage{51}--\blpage{72}
(\byear{2023})
\end{barticle}
\endbibitem

\bibitem[\protect\citeauthoryear{Sasaki and
  Matsubara}{2021}]{sasaki2021variational}
\begin{barticle}
\bauthor{\bsnm{Sasaki}, \binits{H.}},
\bauthor{\bsnm{Matsubara}, \binits{T.}}:
\batitle{Variational policy search using sparse gaussian process priors for
  learning multimodal optimal actions}.
\bjtitle{Neural Networks}
\bvolume{143},
\bfpage{291}--\blpage{302}
(\byear{2021})
\end{barticle}
\endbibitem

\bibitem[\protect\citeauthoryear{Hausman et~al.}{2017}]{hausman2017multi}
\begin{bchapter}
\bauthor{\bsnm{Hausman}, \binits{K.}},
\bauthor{\bsnm{Chebotar}, \binits{Y.}},
\bauthor{\bsnm{Schaal}, \binits{S.}},
\bauthor{\bsnm{Sukhatme}, \binits{G.}},
\bauthor{\bsnm{Lim}, \binits{J.J.}}:
\bctitle{Multi-modal imitation learning from unstructured demonstrations using
  generative adversarial nets}.
In: \bbtitle{Neural Information Processing Systems (NeurIPS)},
pp. \bfpage{1235}--\blpage{1245}
(\byear{2017})
\end{bchapter}
\endbibitem

\bibitem[\protect\citeauthoryear{Creswell
  et~al.}{2018}]{creswell2018generative}
\begin{barticle}
\bauthor{\bsnm{Creswell}, \binits{A.}},
\bauthor{\bsnm{White}, \binits{T.}},
\bauthor{\bsnm{Dumoulin}, \binits{V.}},
\bauthor{\bsnm{Arulkumaran}, \binits{K.}},
\bauthor{\bsnm{Sengupta}, \binits{B.}},
\bauthor{\bsnm{Bharath}, \binits{A.A.}}:
\batitle{Generative adversarial networks: An overview}.
\bjtitle{IEEE Signal Processing Magazine}
\bvolume{35}(\bissue{1}),
\bfpage{53}--\blpage{65}
(\byear{2018})
\end{barticle}
\endbibitem

\bibitem[\protect\citeauthoryear{Florence et~al.}{2022}]{florence2022implicit}
\begin{bchapter}
\bauthor{\bsnm{Florence}, \binits{P.}},
\bauthor{\bsnm{Lynch}, \binits{C.}},
\bauthor{\bsnm{Zeng}, \binits{A.}},
\bauthor{\bsnm{Ramirez}, \binits{O.A.}},
\bauthor{\bsnm{Wahid}, \binits{A.}},
\bauthor{\bsnm{Downs}, \binits{L.}},
\bauthor{\bsnm{Wong}, \binits{A.}},
\bauthor{\bsnm{Lee}, \binits{J.}},
\bauthor{\bsnm{Mordatch}, \binits{I.}},
\bauthor{\bsnm{Tompson}, \binits{J.}}:
\bctitle{Implicit behavioral cloning}.
In: \bbtitle{Conference on Robot Learning (CoRL)},
pp. \bfpage{158}--\blpage{168}
(\byear{2022})
\end{bchapter}
\endbibitem

\bibitem[\protect\citeauthoryear{Chi et~al.}{2024}]{chi2023diffusion}
\begin{botherref}
\oauthor{\bsnm{Chi}, \binits{C.}},
\oauthor{\bsnm{Xu}, \binits{Z.}},
\oauthor{\bsnm{Feng}, \binits{S.}},
\oauthor{\bsnm{Cousineau}, \binits{E.}},
\oauthor{\bsnm{Du}, \binits{Y.}},
\oauthor{\bsnm{Burchfiel}, \binits{B.}},
\oauthor{\bsnm{Tedrake}, \binits{R.}},
\oauthor{\bsnm{Song}, \binits{S.}}:
Diffusion policy: Visuomotor policy learning via action diffusion.
The International Journal of Robotics Research
(2024)
\end{botherref}
\endbibitem

\bibitem[\protect\citeauthoryear{Rasmussen and Williams}{2006}]{RasmussenW06}
\begin{bbook}
\bauthor{\bsnm{Rasmussen}, \binits{C.E.}},
\bauthor{\bsnm{Williams}, \binits{C.K.I.}}:
\bbtitle{Gaussian Processes for Machine Learning}.
\bsertitle{Adaptive Computation and Machine Learning}.
\bpublisher{MIT Press},
\blocation{Cambridge, MA}
(\byear{2006})
\end{bbook}
\endbibitem

\bibitem[\protect\citeauthoryear{Todorov et~al.}{2012}]{6386109}
\begin{bchapter}
\bauthor{\bsnm{Todorov}, \binits{E.}},
\bauthor{\bsnm{Erez}, \binits{T.}},
\bauthor{\bsnm{Tassa}, \binits{Y.}}:
\bctitle{Mujoco: A physics engine for model-based control}.
In: \bbtitle{2012 IEEE/RSJ International Conference on Intelligent Robots and
  Systems (IROS)},
pp. \bfpage{5026}--\blpage{5033}
(\byear{2012})
\end{bchapter}
\endbibitem

\bibitem[\protect\citeauthoryear{Elfwing et~al.}{2018}]{elfwing2018sigmoid}
\begin{barticle}
\bauthor{\bsnm{Elfwing}, \binits{S.}},
\bauthor{\bsnm{Uchibe}, \binits{E.}},
\bauthor{\bsnm{Doya}, \binits{K.}}:
\batitle{Sigmoid-weighted linear units for neural network function
  approximation in reinforcement learning}.
\bjtitle{Neural Networks}
\bvolume{107},
\bfpage{3}--\blpage{11}
(\byear{2018})
\end{barticle}
\endbibitem

\bibitem[\protect\citeauthoryear{Franke et~al.}{2012}]{franke2012chi}
\begin{barticle}
\bauthor{\bsnm{Franke}, \binits{T.M.}},
\bauthor{\bsnm{Ho}, \binits{T.}},
\bauthor{\bsnm{Christie}, \binits{C.A.}}:
\batitle{The chi-square test: Often used and more often misinterpreted}.
\bjtitle{American journal of evaluation}
\bvolume{33}(\bissue{3}),
\bfpage{448}--\blpage{458}
(\byear{2012})
\end{barticle}
\endbibitem

\bibitem[\protect\citeauthoryear{Ramachandran
  et~al.}{2018}]{ramachandran2017searching}
\begin{bchapter}
\bauthor{\bsnm{Ramachandran}, \binits{P.}},
\bauthor{\bsnm{Zoph}, \binits{B.}},
\bauthor{\bsnm{Le}, \binits{Q.V.}}:
\bctitle{Searching for activation functions}.
In: \bbtitle{International Conference on Learning Representations (ICLR)}
(\byear{2018})
\end{bchapter}
\endbibitem

\bibitem[\protect\citeauthoryear{Ross and Dy}{2013}]{ross13a}
\begin{bchapter}
\bauthor{\bsnm{Ross}, \binits{J.}},
\bauthor{\bsnm{Dy}, \binits{J.}}:
\bctitle{Nonparametric mixture of gaussian processes with constraints}.
In: \beditor{\bsnm{Dasgupta}, \binits{S.}},
\beditor{\bsnm{McAllester}, \binits{D.}} (eds.)
\bbtitle{International Conference on Machine Learning (ICML)},
vol. \bseriesno{28},
pp. \bfpage{1346}--\blpage{1354}
(\byear{2013})
\end{bchapter}
\endbibitem

\bibitem[\protect\citeauthoryear{Oh et~al.}{2023}]{OH202342}
\begin{barticle}
\bauthor{\bsnm{Oh}, \binits{H.}},
\bauthor{\bsnm{Sasaki}, \binits{H.}},
\bauthor{\bsnm{Michael}, \binits{B.}},
\bauthor{\bsnm{Matsubara}, \binits{T.}}:
\batitle{Bayesian disturbance injection: Robust imitation learning of flexible
  policies for robot manipulation}.
\bjtitle{Neural Networks}
\bvolume{158},
\bfpage{42}--\blpage{58}
(\byear{2023})
\end{barticle}
\endbibitem

\bibitem[\protect\citeauthoryear{Jyl{{\"a}}nki et~al.}{2011}]{jylanki11a}
\begin{barticle}
\bauthor{\bsnm{Jyl{{\"a}}nki}, \binits{P.}},
\bauthor{\bsnm{Vanhatalo}, \binits{J.}},
\bauthor{\bsnm{Vehtari}, \binits{A.}}:
\batitle{Robust gaussian process regression with a student-t likelihood}.
\bjtitle{Journal of Machine Learning Research}
\bvolume{12}(\bissue{99}),
\bfpage{3227}--\blpage{3257}
(\byear{2011})
\end{barticle}
\endbibitem

\bibitem[\protect\citeauthoryear{Li et~al.}{2020}]{li2020learning}
\begin{barticle}
\bauthor{\bsnm{Li}, \binits{X.}},
\bauthor{\bsnm{Huang}, \binits{H.}},
\bauthor{\bsnm{Zhao}, \binits{H.}},
\bauthor{\bsnm{Wang}, \binits{Y.}},
\bauthor{\bsnm{Hu}, \binits{M.}}:
\batitle{Learning a convolutional neural network for propagation-based stereo
  image segmentation}.
\bjtitle{The Visual Computer}
\bvolume{36},
\bfpage{39}--\blpage{52}
(\byear{2020})
\end{barticle}
\endbibitem

\bibitem[\protect\citeauthoryear{Zhang et~al.}{2021}]{zhang2021bioinspired}
\begin{barticle}
\bauthor{\bsnm{Zhang}, \binits{L.}},
\bauthor{\bsnm{Su}, \binits{G.}},
\bauthor{\bsnm{Yin}, \binits{J.}},
\bauthor{\bsnm{Li}, \binits{Y.}},
\bauthor{\bsnm{Lin}, \binits{Q.}},
\bauthor{\bsnm{Zhang}, \binits{X.}},
\bauthor{\bsnm{Shao}, \binits{L.}}:
\batitle{Bioinspired scene classification by deep active learning with remote
  sensing applications}.
\bjtitle{IEEE Transactions on Cybernetics}
\bvolume{52}(\bissue{7}),
\bfpage{5682}--\blpage{5694}
(\byear{2021})
\end{barticle}
\endbibitem

\bibitem[\protect\citeauthoryear{Jin et~al.}{2019}]{jin2019multi}
\begin{barticle}
\bauthor{\bsnm{Jin}, \binits{X.}},
\bauthor{\bsnm{He}, \binits{T.}},
\bauthor{\bsnm{Lin}, \binits{Y.}}:
\batitle{Multi-objective model selection algorithm for online sequential
  ultimate learning machine}.
\bjtitle{EURASIP Journal on Wireless Communications and Networking}
\bvolume{2019},
\bfpage{1}--\blpage{7}
(\byear{2019})
\end{barticle}
\endbibitem

\bibitem[\protect\citeauthoryear{Chen et~al.}{2020}]{CHEN2020150}
\begin{barticle}
\bauthor{\bsnm{Chen}, \binits{L.}},
\bauthor{\bsnm{Su}, \binits{W.}},
\bauthor{\bsnm{Feng}, \binits{Y.}},
\bauthor{\bsnm{Wu}, \binits{M.}},
\bauthor{\bsnm{She}, \binits{J.}},
\bauthor{\bsnm{Hirota}, \binits{K.}}:
\batitle{Two-layer fuzzy multiple random forest for speech emotion recognition
  in human-robot interaction}.
\bjtitle{Information Sciences}
\bvolume{509},
\bfpage{150}--\blpage{163}
(\byear{2020})
\end{barticle}
\endbibitem

\bibitem[\protect\citeauthoryear{Dallas et~al.}{2017}]{dallas2017comparison}
\begin{barticle}
\bauthor{\bsnm{Dallas}, \binits{S.}},
\bauthor{\bsnm{Machairas}, \binits{K.}},
\bauthor{\bsnm{Papadopoulos}, \binits{E.}}:
\batitle{A comparison of ordinary differential equation solvers for dynamical
  systems with impacts}.
\bjtitle{Journal of Computational and Nonlinear Dynamics}
\bvolume{12}(\bissue{6}),
\bfpage{061016}
(\byear{2017})
\end{barticle}
\endbibitem

\bibitem[\protect\citeauthoryear{Wanner and Hairer}{1996}]{wanner1996solving}
\begin{bbook}
\bauthor{\bsnm{Wanner}, \binits{G.}},
\bauthor{\bsnm{Hairer}, \binits{E.}}:
\bbtitle{Solving Ordinary Differential Equations II}
vol. \bseriesno{375}.
\bpublisher{Springer},
\blocation{Heidelberg}
(\byear{1996})
\end{bbook}
\endbibitem

\bibitem[\protect\citeauthoryear{Hu and Tang}{2003}]{hu2003numerical}
\begin{bbook}
\bauthor{\bsnm{Hu}, \binits{J.}},
\bauthor{\bsnm{Tang}, \binits{H.}}:
\bbtitle{Numerical Methods for Differential Equations},
pp. \bfpage{6}--\blpage{9}.
\bpublisher{City University},
\blocation{Hong Kong}
(\byear{2003})
\end{bbook}
\endbibitem

\bibitem[\protect\citeauthoryear{Ernst~Hairer}{1993}]{hairer1993}
\begin{bbook}
\bauthor{\bsnm{Ernst~Hairer}, \binits{S.P.N.} \bsuffix{Gerhard~Wanner}}:
\bbtitle{Solving Ordinary Differential Equations I: Nonstiff Problems},
pp. \bfpage{129}--\blpage{353}.
\bpublisher{Springer},
\blocation{Heidelberg}
(\byear{1993})
\end{bbook}
\endbibitem

\bibitem[\protect\citeauthoryear{Grathwohl
  et~al.}{2018}]{grathwohl2019scalable}
\begin{bchapter}
\bauthor{\bsnm{Grathwohl}, \binits{W.}},
\bauthor{\bsnm{Chen}, \binits{R.T.}},
\bauthor{\bsnm{Bettencourt}, \binits{J.}},
\bauthor{\bsnm{Duvenaud}, \binits{D.}}:
\bctitle{Scalable reversible generative models with free-form continuous
  dynamics}.
In: \bbtitle{1st Symposium on Advances in Approximate Bayesian Inference},
pp. \bfpage{1}--\blpage{14}
(\byear{2018})
\end{bchapter}
\endbibitem

\end{thebibliography}

\end{document}